\RequirePackage{fix-cm}
\documentclass{article}       
\usepackage{graphicx}
\usepackage{lineno,hyperref}
\usepackage{amsmath,amssymb,amsfonts}
\usepackage{algorithmic}
\usepackage{textcomp}
\usepackage[english]{babel}
\usepackage[utf8]{inputenc}
\usepackage[T1]{fontenc}
\usepackage{graphics,booktabs,colortbl,multirow}
\usepackage{color,soul}
\usepackage{lscape} 
\usepackage[table]{xcolor}
\usepackage{tikz}
\usepackage{array}
\newcolumntype{M}[1]{>{\centering\arraybackslash}m{#1}}
\def\BibTeX{{\rm B\kern-.05em{\sc i\kern-.025em b}\kern-.08em
    T\kern-.1667em\lower.7ex\hbox{E}\kern-.125emX}}
\modulolinenumbers[5]
\begin{document}

\title{Neural Style Transfer and Unpaired Image-to-Image Translation to deal with the Domain Shift Problem on Spheroid Segmentation
}


\author{M. García-Domínguez \and C.Dom\'inguez \and J.Heras \and E.Mata  \and V.Pascual} 




\maketitle

\begin{abstract}

Domain shift is a generalisation problem of machine learning models that occurs when the data distribution of the training set is different to the data distribution encountered by the model when it is deployed. In this work, we address this common problem in the context of the segmentation of tumour spheroids by studying both neural style transfer algorithms and unpaired image-to-image translation methods. We have illustrated the domain shift problem with 4 deep learning segmentation models that decreased their performance when applied in different distributions. In order to deal with this problem, we have explored 3 style transfer algorithms, and 6 unpaired image-to-image translations algorithms. These algorithms have been integrated into a high-level API that facilitates their application to other contexts where the domain-shift problem occurs. We have considerably improved the performance of the 4 segmentation models when applied to images other distributions by using the high level API. In particular, there are 2 style transfer algorithms (NST and deep image analogy) and 1 unpaired image-to-image translations algorithm (CycleGAN) that improve the IoU of the models in a range from 0.24 to 76.07.

\end{abstract}

\section{Introduction}
Deep convolutional neural networks have become the state-of-the-art approach to tackle segmentation problems in medicine~\cite{Unet,Sirinukunwattana17}. However, there are several challenges that hinder the training and deployment of deep learning models in this context. First of all, a considerable amount of annotated images is needed to train a deep model, and annotating datasets for image segmentation is a tedious and time-consuming task that requires expert knowledge~\cite{Litjens17}. Moreover, there is an important generalisation challenge when using trained models that is known as \emph{domain shift} (also known as distribution shift)~\cite{Arvidsson18,Choudary20}. This problem arises when the data distribution of the dataset used for training a model is not the same than the data that the model encounters when deployed. This is common in biomedical datasets since images greatly vary due to experimental conditions, the equipment (for instance, microscopes) and settings (for instance, focus and magnification) employed for capturing those images.

This generalisation problem can be tackled by combining datasets from multiple sources~\cite{CheXpert} or using techniques like data augmentation~\cite{Simard03}; nevertheless, it is not possible to foresee every new and unknown distribution. A different approach consists in applying transfer learning~\cite{Razavian14}, a technique that, instead of training a model from scratch, reuses a model pre-trained in a source dataset to train a new model in a target dataset. However, this requires the annotation of the target dataset, a time-consuming task that should be carried out for every new dataset. A different approach to handle the domain shift problem is image-to-image translation~\cite{pix2pix}, a set of techniques that aim to learn the mapping between an input image and an output image using a training set of aligned image pairs; however, this requires paired data from the source and target domains, a challenge that can be faced by using unpaired image-to-image translation~\cite{cyclegan}.

Unpaired image-to-image translation methods translate an image from a domain $A$ to a domain $B$, and vice versa, in the absence of paired examples. This approach has been already employed in several medical segmentation tasks; for instance, the segmentation of the left ventricle in magnetic resonance images~\cite{Yan19}, the segmentation of digitally reconstructed radiographs~\cite{Zhang18}, and the segmentation of magnetic resonance imaging (MRI), abdominal CT and MRI, and mammography X-rays~\cite{CAI2019174}. All these works are based on variants of CycleGAN~\cite{cyclegan}, an unpaired image-to-image translation method based on Generative Adversarial Networks (GANs) that requires two datasets: one of them contains images from the distribution employed for training the segmentation model, and the other contains images acquired in a different setting. This approach poses two challenges. First, both datasets must be available, and this might be an issue due to privacy concerns~\cite{JAMA12}; and, secondly, CycleGAN variants must be trained, a process that demands the usage of GPUs and might be challenging for several users due to the difficulties of training GAN models~\cite{GANDifficult}. The approach proposed in this paper to tackle these drawbacks consists in using style transfer methods~\cite{NeuralStyleTransfer}; that is, techniques that render the content of an image using the style of another. Those techniques do not require a training process, and it is enough with releasing one image of the dataset employed for training the model that suffers the domain shift problem.  

In this work, we have studied both unpaired image-to-image translation methods and style transfer techniques to deal with the domain shift problem. As a running example, we have focused on the problem of segmenting tumour spheroids~\cite{Nath16}. In this context, we have observed, see Section~\ref{sec:materials}, that models that achieve a mean IoU over 97\% when evaluating with data following the same distribution as the training set, fail when they are employed with data following a different distribution (the IoU is, in some cases, under 15\%). We have faced this domain shift problem by using both unpaired image-to-image translation methods and style transfer techniques. Namely, the contributions of our work are:

\begin{itemize}
    \item We explore several state-of-the-art style transfer and unpaired image-to-image translation methods to tackle the domain shift problem in the context of tumour spheroid segmentation. 
    \item We demonstrate the effectiveness of using both style transfer and unpaired image-to-image translation methods to improve the performance of a variety of advanced deep segmentation networks. 
    \item We show that style transfer methods achieve similar performance than unpaired image-to-image translation methods with the advantage of skipping the training step. 
    \item We provide an API to apply the studied methods not only in the context of spheroid segmentation but in general for medical imaging tasks. The API is available at \url{https://github.com/ManuGar/ImageStyleTransfer}.
\end{itemize}

\section{Materials}\label{sec:materials}

Spheroids are the most widely used 3D models to study cancer since they can be used for studying the effects of different micro-environmental characteristics on tumour behaviour and for testing different preclinical and clinical treatments~\cite{Nath16}. The images from tumour spheroids greatly vary depending on the  experimental conditions, and also on the equipment (microscopes) and conditions (focus and magnification) employed to capture the images~\cite{SpheroidJ}.

\begin{table}
\caption{Features of the 4 datasets employed in this work. The former three datasets were employed for training and the last dataset was used for testing}
    \label{tab:features}
    \centering
    \resizebox{\columnwidth}{!}{
    \begin{tabular}{c|cccccc}
    \toprule
         Dataset &  $\sharp$ Images & Image size & Microscope & Magnification & Format & Type  \\
         \midrule
         BL5S &  50 & 1296$\times$ 966 & Leica & 5x & TIFF & RGB \\
         BN2S &  154 & 1002$\times$ 1004 & Nikon & 2x & ND2 & Gray 16bits \\
         BN10S &  105 & 1002$\times$ 1004 & Nikon & 10x & ND2 & Gray 16bits \\
         \midrule 
         BO10S & 64 & 3136 $\times$ 2152 & Olympus & 10x & JPG & RGB \\
         \bottomrule
    \end{tabular}}
\end{table}

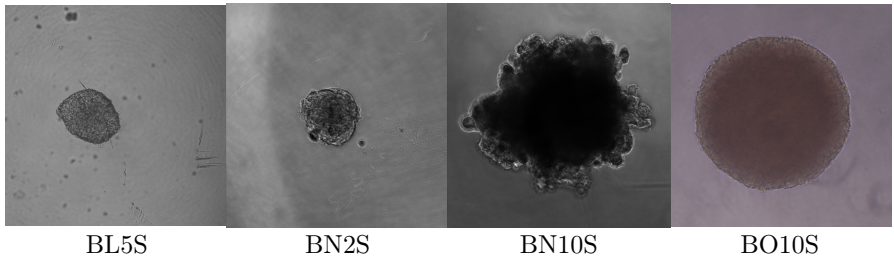
\begin{figure}
    \centering
    \resizebox{\columnwidth}{!}{
    \begin{tikzpicture}
     
     \draw (0,-1.75) node{BL5S};
     \draw (3,-1.75) node{BN2S};
     \draw (6,-1.75) node{BN10S};
     \draw (9,-1.75) node{BO10S};
     
     \draw (9,0) node{\includegraphics[width=0.25\columnwidth]{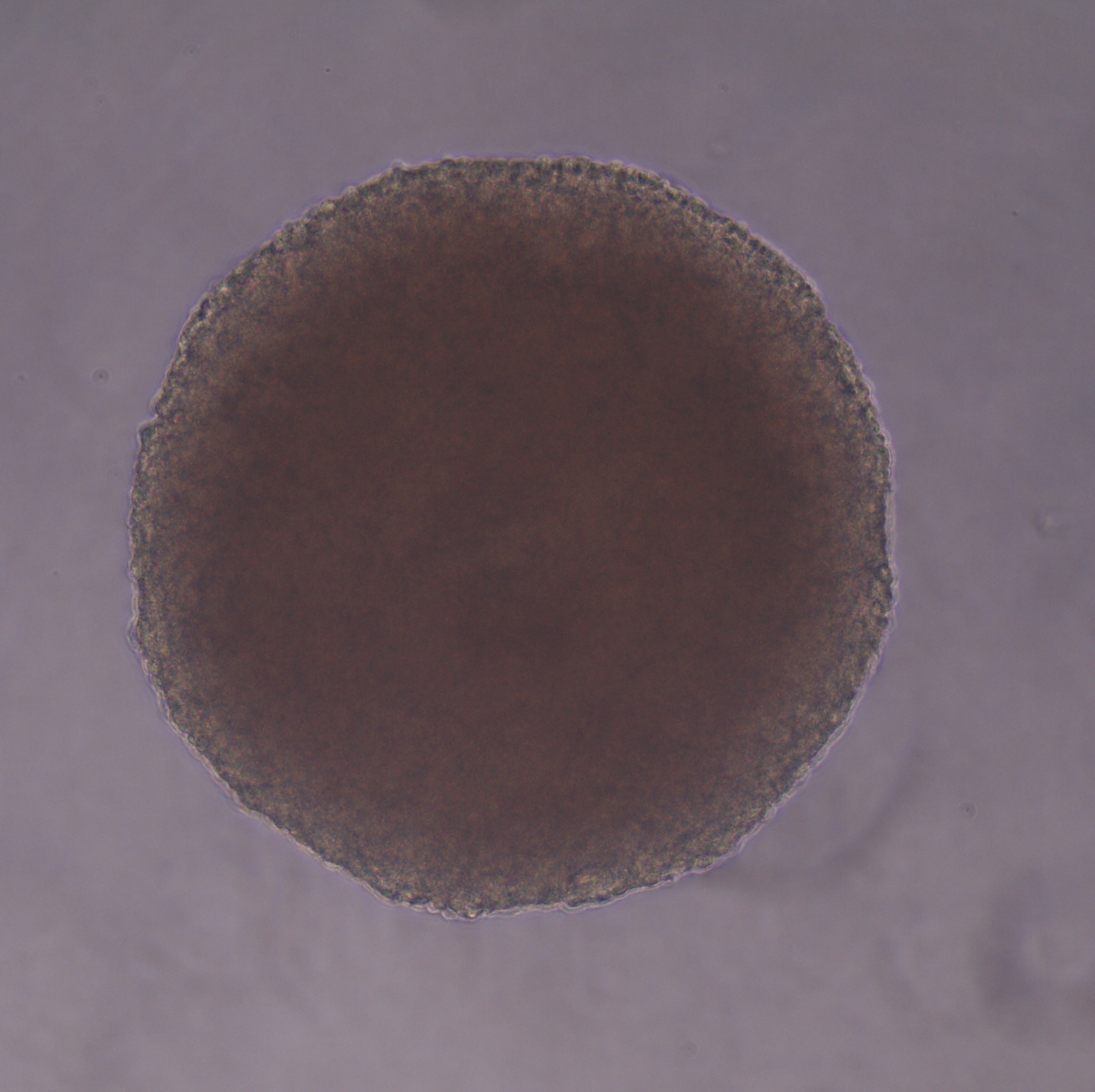}};
     \draw (0,0) node{\includegraphics[width=0.25\columnwidth]{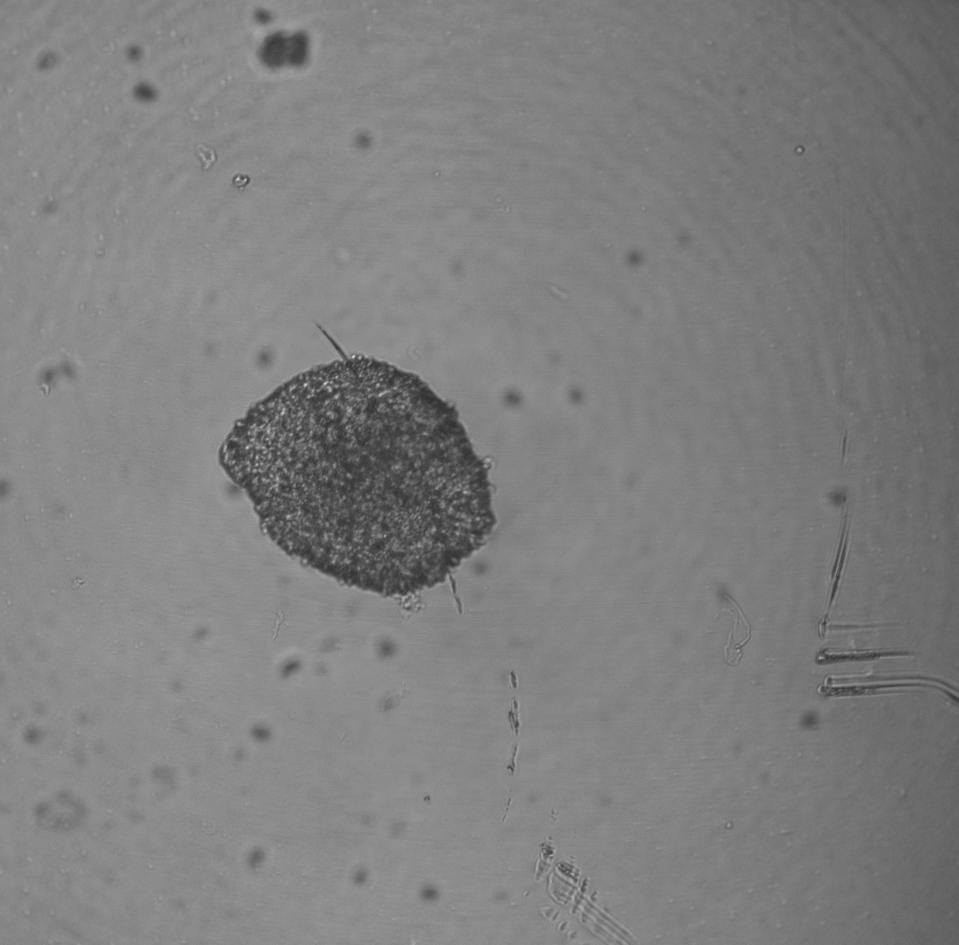}};
     \draw (3,0) node{\includegraphics[width=0.25\columnwidth]{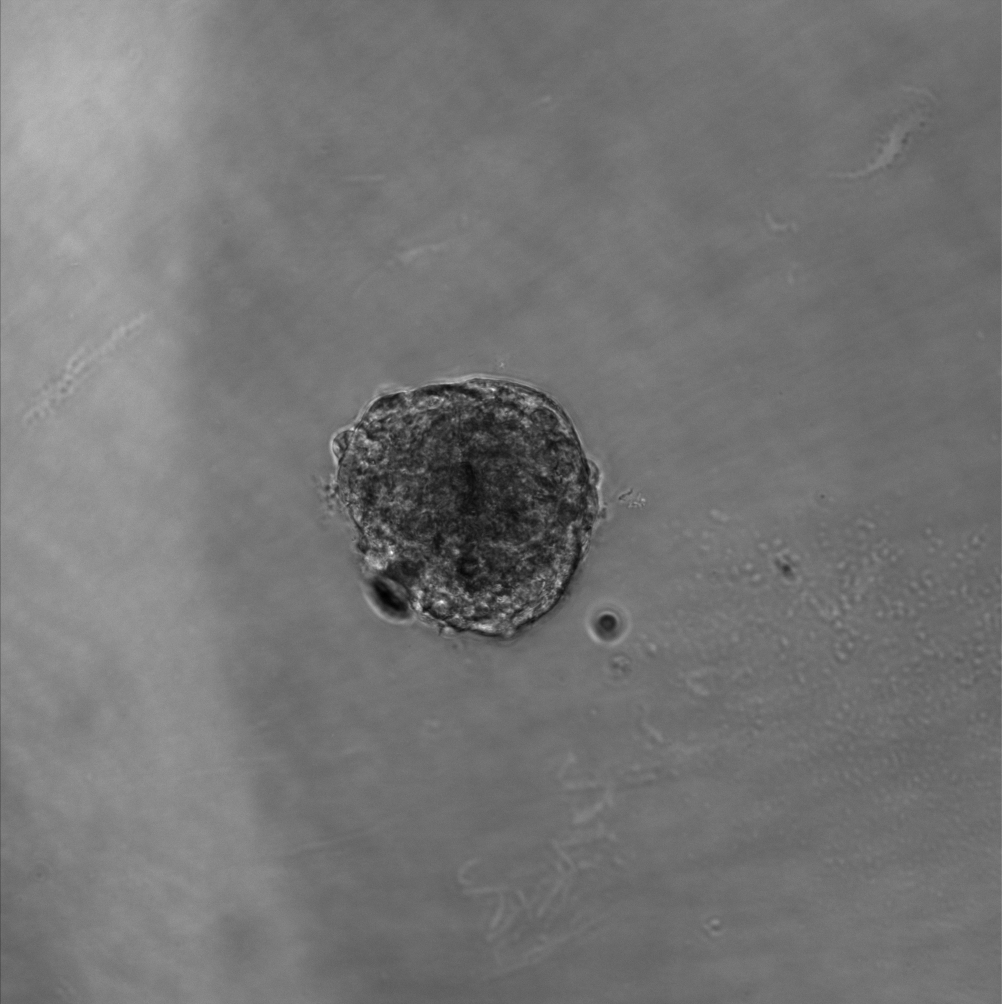}};
     \draw (6,0) node{\includegraphics[width=0.25\columnwidth]{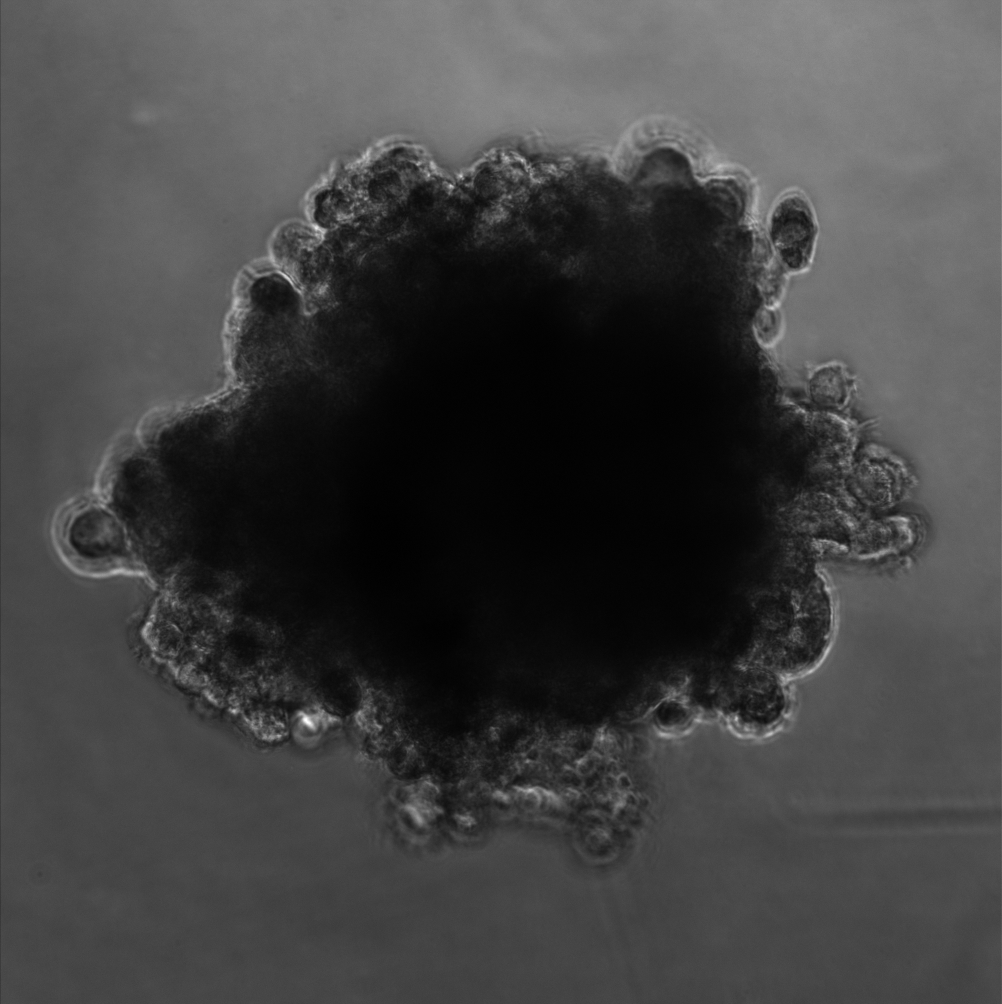}};
    \end{tikzpicture}}
    \caption{Samples from the 4 datasets employed in this work}
    \label{fig:samples}
\end{figure}

For our experiments, we have employed the 4 datasets presented in~\cite{SpheroidJ}; a description of those datasets is provided in Table~\ref{tab:features}, and an image of each dataset is shown in Figure~\ref{fig:samples}. As can be noticed from Table~\ref{tab:features} and Figure~\ref{fig:samples}, there are considerable differences among the images of each dataset. Three of those datasets (the BL5S, BN2S, and BN10S datasets) were employed for training 4 segmentation models (using the algorithms DeepLab v3~\cite{DeepLabv3}, HRNet Seg~\cite{HRNet}, U-net~\cite{Unet} and U$^2$-Net~\cite{U2Net}) and the last dataset (the BO10S dataset) was employed for testing. We have used this dataset split because the last dataset comes from a different laboratory so its style will not be the same as the others. The definition of those 4 architectures is available in the SemTorch package\footnote{The SemTorch package is available at \url{https://github.com/WaterKnight1998/SemTorch}}. All the architectures were trained with the libraries PyTorch~\cite{Pytorch} and FastAI~\cite{Fastai} and using a GPU Nvidia RTX 2080 Ti. In order to set the learning rate for the different architectures, we employed the procedure presented in \cite{Fastai}; and, we applied early stopping when training all the architectures to avoid overfitting. The metric employed to measure the accuracy of the different methods is the IoU, also known as Jaccard index --- this metric measures the area of intersection between the ground truth and the predicted region over the area of union between the ground truth and the predicted region.

When the models were evaluated using a test set formed by images following the same distribution than the training set, the 4 models achieved a performance over 97\%, see Table~\ref{tab:init-results}. On the contrary, when those models were employed with images captured under different conditions (namely, using the BO10S dataset), the performance of the models decrease by up to 84\%. In the next sections, we explore how style transfer methods, and unpaired image-to-image translation models can serve to deal with the domain shift problem in this context. 

\begin{table}[]
\caption{Performance of the 4 models when evaluating in a test set formed from images following the same distribution than the training set (BL5S-BN2S-BN10S), and when evaluated using a test set from a different distribution (BO10S)}\label{tab:init-results}
    \centering
    \resizebox{\columnwidth}{!}{
    \begin{tabular}{ccccc}
    \toprule
         &  DeepLab v3 & HRNet-Seg &  U-Net & U2-Net\\
         \midrule
         BL5S-BN2S-BN10S  & 97.00  &  97.32& 97.25  & 97.26\\
    BO10S   & 83.61  & 92.65 &  13.64 & 95.65 \\
    \bottomrule
    \end{tabular}}
\end{table}

\section{Style transfer}\label{sec:methods}

This section is devoted to present how style transfer methods can handle the domain shift problem. In addition, we introduce the API that we have developed to facilitate the use of those methods. Finally, we present the results obtained by the spheroid segmentation models when applied to images transformed using style transfer techniques.

\begin{figure}
    \centering
     \includegraphics[width=\columnwidth]{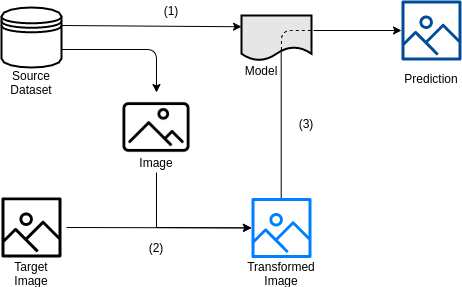}
    \caption{Workflow of the style transfer approach. (1) A model is trained using a source dataset. (2) The target image is transformed using the style from an image of the source dataset. (3) The transformed image is fed to the model.}
    \label{fig:samples1}
\end{figure}

We start by explaining the procedure to apply style transfer methods to deal with the domain shift problem of a model --- such a procedure is summarised in Figure~\ref{fig:samples1}. We assume that a model has been trained using a source dataset of images, and we are interested in applying such a model to obtain the prediction associated with an image from a different distribution than the source dataset; we call this image, the target image. Instead of feeding the target image directly to the model, we first take an image from the source dataset and transfer the style of that image to the target image but preserving its content producing a transformed image. Finally, the transformed image is fed to the model to obtain the associated prediction.

The key component of the aforementioned process is the algorithm that transfers the style from the source dataset but keeping the content of the target image. In the literature, there are several style transfer algorithms~\cite{Liu19}; but, for our experiments, we have focused on three of them: neural style transfer (NST)~\cite{NeuralStyleTransfer}, an optimisation technique that uses a Convolutional Neural Network (CNN) to decompose the content and style from images; deep image analogy~\cite{Liao17}, a method that finds semantically-meaningful correspondences between two input images by adapting the notion of image analogy with features extracted from a CNN; and STROTSS~\cite{Kolkin19}, a variant of the NST algorithm that changes the optimisation objective of NST.

It is worth noting that the style transfer approach presented here can be applied to deal with the domain shift problem not only for segmentation problems, as in our work, but also to other computer vision tasks. Hence, these methods can be helpful for a great variety of problems. However, it might be difficult to apply these techniques since they are implemented in different libraries and using different frameworks, and each of them has its own particularities. In this work, we have addressed this drawback by developing a high-level Python API that allows the integration of style transfer algorithms independently of their underlying library and framework. The API currently includes the aforementioned methods (the project webpage provides information about the library that implements each method) and can be easily extended with new techniques. In order to apply the previously introduced procedure using our API, users only have to provide the style image, the target image, and the name of the algorithm to apply; the rest of the transformation process is automatically conducted by the API.  

In our running example of segmenting tumour spheroids, and using our API, we randomly picked an image from the combination of the datasets BL5S, BN2S, and BN10S, and used it to transform the images from the BO10S dataset. Subsequently, we fed those images to the segmentation models presented in Section~\ref{sec:materials}, and evaluated their performance, see Table~\ref{tab:results-style-transfer}. From the three studied style transfer algorithms, both the NST and STROTSS algorithms handle the domain shift problem; whereas, the images transformed with the deep image analogy algorithm produce even worse results than the original images from the BO10S dataset. Using the NST algorithm, all the segmentation models improve their IoU (the U-Net model improves its performance from 13.64\% to 89.21\%, and the other models have an IoU close to 95\%). For the STROTSS algorithm, the results are also positive: two of the segmentation models improve (DeepLab and U-Net), and the other two achieve worse results, but still their IoU is over 92\%. In the next section, we extend this study with Unpaired image-to-image translation methods. 

\begin{table}[]
\caption{Performance for the BO10S dataset using the different style-transfer methods to deal with the domain shift problem. A \textcolor{blue}{$\uparrow$} indicates an improvement with respect to the base model, whereas a \textcolor{red}{$\downarrow$} indicates a declination in the performance.}
    \label{tab:results-style-transfer}
    \centering
    \resizebox{\columnwidth}{!}{
    \begin{tabular}{ccccc}
    \toprule
         &  DeepLab v3 & HRNet-Seg &  U-Net & U$^2$-Net\\
         \midrule
    Base   & 83.61  & 92.65 &  13.64 & 95.65 \\
    \midrule
    NST & 95.64\textcolor{blue}{$\uparrow$}   & 94.91\textcolor{blue}{$\uparrow$} &89.21\textcolor{blue}{$\uparrow$}   & 95.89\textcolor{blue}{$\uparrow$}  \\
     Deep Image Analogy & 0.00\textcolor{red}{$\downarrow$}&45.13 \textcolor{red}{$\downarrow$} & 0.66\textcolor{red}{$\downarrow$} & 0.84\textcolor{red}{$\downarrow$} \\
    STROTSS& 94.86\textcolor{blue}{$\uparrow$}&92.38\textcolor{red}{$\downarrow$} &78.08\textcolor{blue}{$\uparrow$} &94.14\textcolor{red}{$\downarrow$} \\
    \bottomrule
    \end{tabular}}
\end{table}




\section{Unpaired image-to-image translation}

\begin{figure}
    \centering
     \includegraphics[width=\columnwidth]{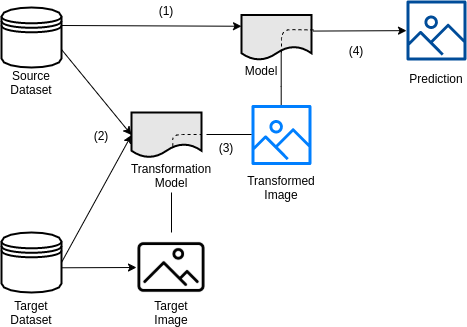}
    \caption{Workflow of the unpaired image-to-image translation models. (1) A model is trained using a source dataset. (2) A source dataset and a target dataset are employed to combine a GAN model. (3) Given an image from the distribution of the target dataset, the GAN model is employed to transform the image. (4) The transformed image is fed to the prediction model.}
    \label{fig:samples2}
\end{figure}

We focus now on the procedure to apply unpaired image-to-image translation methods to tackle the domain shift problem, see Figure~\ref{fig:samples2}. Analogously to the style transfer approach, we assume that a model has been trained using a source dataset of images, and we have a dataset of images with a different data distribution called the target dataset. From the source and target datasets, we build a model that transforms images following the data distribution of the source dataset to the data distribution of the target dataset, and vice versa. Now, when we are interested in obtaining the prediction associated with a target image, we first employ the transformation model to transform the image; and, subsequently, the transformed image is fed to the prediction model.

In this approach, the key component is the algorithm employed to construct the transformation model. Currently, the most successful approaches for this task are based on Generative Adversarial Networks (GANs)~\cite{GAN}; and, are variants of the CycleGAN algorithm~\cite{cyclegan}, which translates an image from a source domain $X$ to a target domain $Y$ by learning two mappings $G_X: X \rightarrow Y$ and $G_Y: Y \rightarrow X$ such that they satisfy the cycle-consistency properties; that is $G_Y (G_X (x)) \approx x$ and $G_X (G_Y (y)) \approx y$. For our experiments, we have studied 6 algorithms: CycleGAN~\cite{cyclegan}, DualGAN~\cite{dualgan},
ForkGAN~\cite{forkgan}, GANILLA~\cite{GANILLA}, CUT~\cite{park2020cut}, and FASTCUT~\cite{park2020cut}.

As we have previously mentioned for the style transfer methods, the aforementioned procedure for unpaired image-to-image translation can be applied to several kinds of bioimaging problems~\cite{unpairedTranlation}. Hence, we have extended the API presented in the previous section to provide access to unpaired image-to-image translation techniques. In this case, users of the API only have to provide the path to the source and target datasets, and the name of the translation algorithm to apply (the 6 algorithms previously mentioned are available in the API, and new methods can be easily included); after that, the transformation model is automatically trained and the images from the target dataset are transformed.  

In our running example, the source dataset was formed by the combination of the datasets BL5S, BN2S, and BN10S; and the target dataset was the BO10S dataset. For training the translation models, we employed a GPU Nvidia RTX 2080 Ti, and the results obtained by the segmentation models when they were fed with the transformed images are summarised in Table~\ref{tab:results-unpaired}. From the 6 studied algorithms, only the CycleGAN method solves the domain shift problem in our context. Using the transformation model produced by this algorithm, all the segmentation models improve their IoU (the U-Net model improves its performance from 13.64\% to 72.34\%, and the other models have an IoU over to 92\%). On the contrary, the images produced by the rest of the transformations models are more difficult to segment, and the performance of the segmentation models decreases --- the exception is the U-Net model that, in some cases, obtains better results with the transformed images, but still its IoU is under 40\%. 

\begin{table}[t]
\caption{Performance for the BO10S dataset using the different unpaired image-to-image translation methods to deal with the domain shift problem. A \textcolor{blue}{$\uparrow$} indicates an improvement with respect to the base model, whereas a \textcolor{red}{$\downarrow$} indicates a declination in the performance.}
    \label{tab:results-unpaired}
    \centering
\resizebox{\columnwidth}{!}{
    \begin{tabular}{ccccc}
    \toprule
         &  DeepLab v3 & HRNet-Seg &  U-Net & U2-Net\\
         \midrule
    Base   & 83.61  & 92.65 &  13.64 & 95.65 \\
    \midrule
    CycleGAN & 94.97\textcolor{blue}{$\uparrow$}  & 92.97\textcolor{blue}{$\uparrow$} &72.34\textcolor{blue}{$\uparrow$}  & 95.87\textcolor{blue}{$\uparrow$}\\
    DualGAN & 4.09\textcolor{red}{$\downarrow$}& 73.37\textcolor{red}{$\downarrow$}& 24.67\textcolor{blue}{$\uparrow$}&34.45\textcolor{red}{$\downarrow$}\\
    ForkGAN &  32.63\textcolor{red}{$\downarrow$}&46.10\textcolor{red}{$\downarrow$} &38.33\textcolor{blue}{$\uparrow$} &44.46\textcolor{red}{$\downarrow$} \\
    GANILLA & 24.27\textcolor{red}{$\downarrow$}& 76.24\textcolor{red}{$\downarrow$} & 3.26\textcolor{red}{$\downarrow$} &82.97 \textcolor{red}{$\downarrow$} \\
    CUT & 0.48\textcolor{red}{$\downarrow$}&38.01 \textcolor{red}{$\downarrow$}& 20.94\textcolor{blue}{$\uparrow$} & 52.20\textcolor{red}{$\downarrow$} \\
    FastCUT & 6.08\textcolor{red}{$\downarrow$} & 79.52 \textcolor{red}{$\downarrow$} & 1.12\textcolor{red}{$\downarrow$} & 2.98\textcolor{red}{$\downarrow$}\\
    \bottomrule
    \end{tabular}
    }
\end{table}

\section{Discussion}

In the previous sections, we have demonstrated that both style transfer and unpaired image-to-image translation algorithms can be applied to deal with the domain shift problem in the context of segmenting tumour spheroids. However, there are only three successful methods: NST, STROTSS, and CycleGAN. Two of those methods are style transfer algorithms (NST and STROTSS), this is a relevant result since these methods do not require the training step of image-to-image translation models, can be run in a computer without special purpose hardware like GPUs, and only require the availability of an image from the source dataset. The most likely reason for the failure of most unpaired image-to-image translation is the challenge of training their underlying GAN models~\cite{GANDifficult}; therefore, more research is necessary to facilitate the use of this kind of models.

\begin{figure*}
\begin{center}

    \centering\resizebox{0.85\textwidth}{!}{
    \begin{tabular}[t]{M{1.8cm}M{1.8cm}M{1.8cm}M{1.8cm}M{1.8cm}M{1.8cm}M{1.8cm}}
\toprule
                   & Image &Truth & DeepLab v3 & HRNet Seg & U-Net & U$^2$-Net                                                                                               \\ \midrule
\centering Base           & \includegraphics[width=0.15\textwidth]{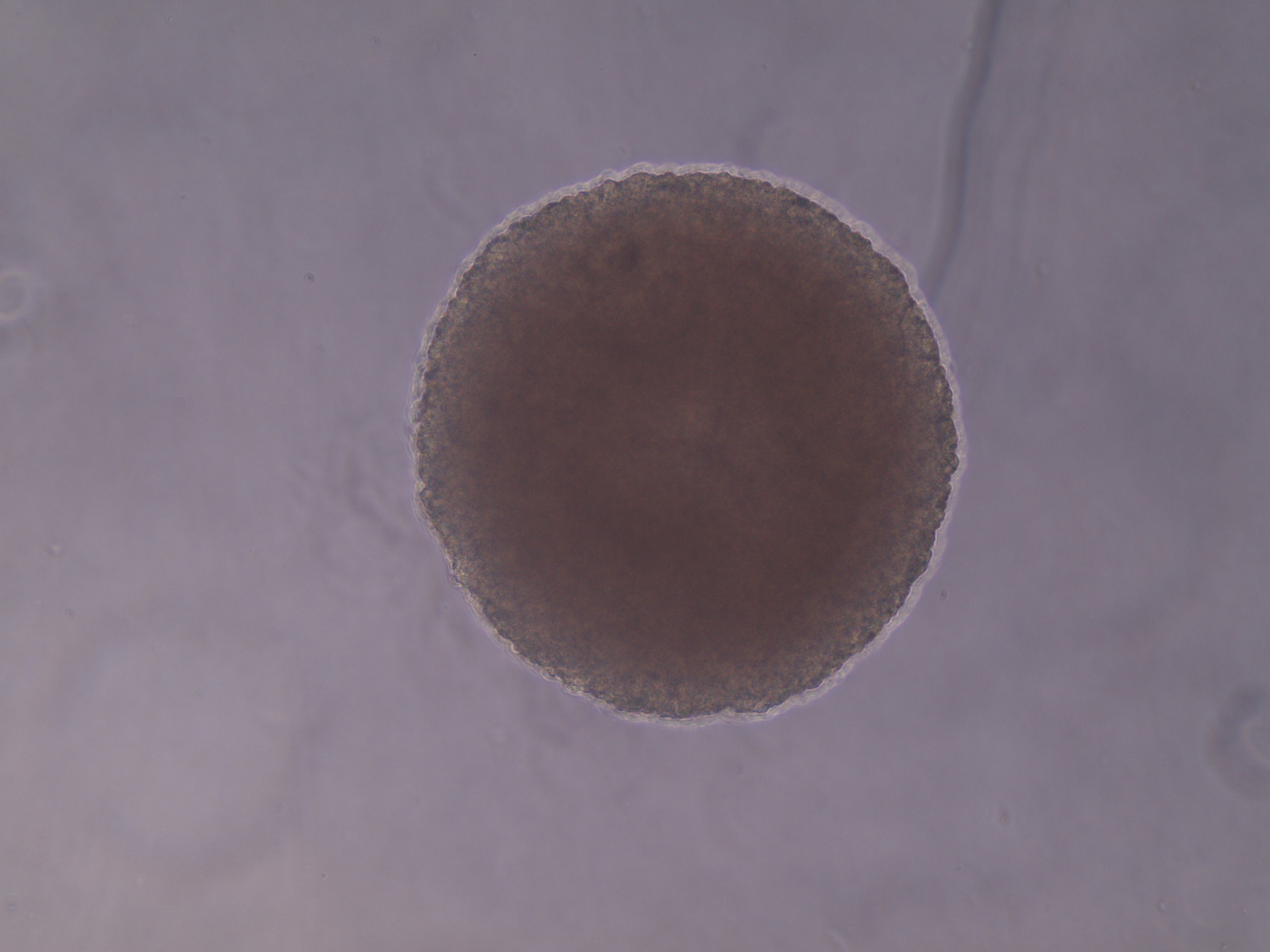} &
\includegraphics[width=0.15\textwidth]{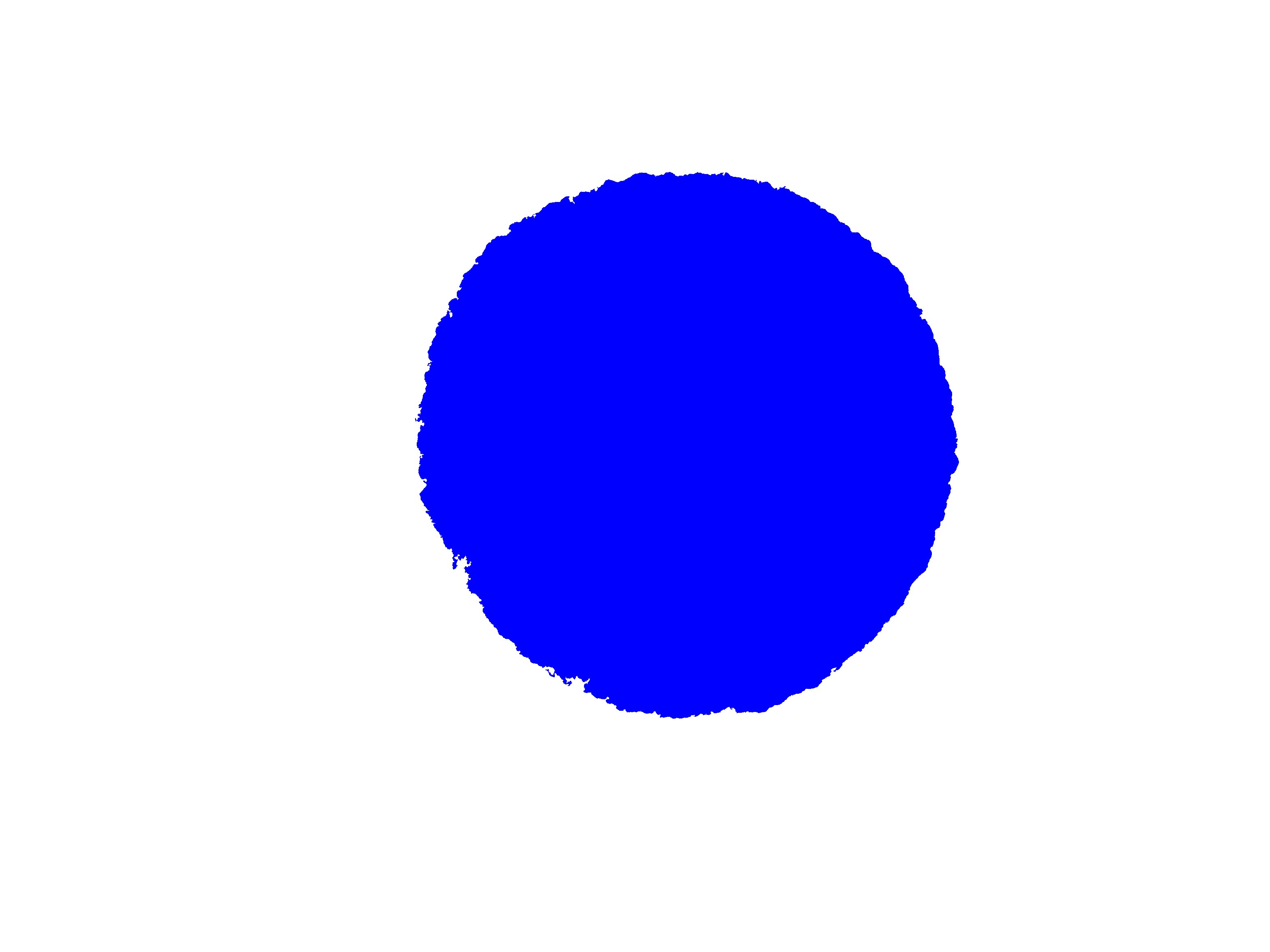}&
\includegraphics[width=0.15\textwidth]{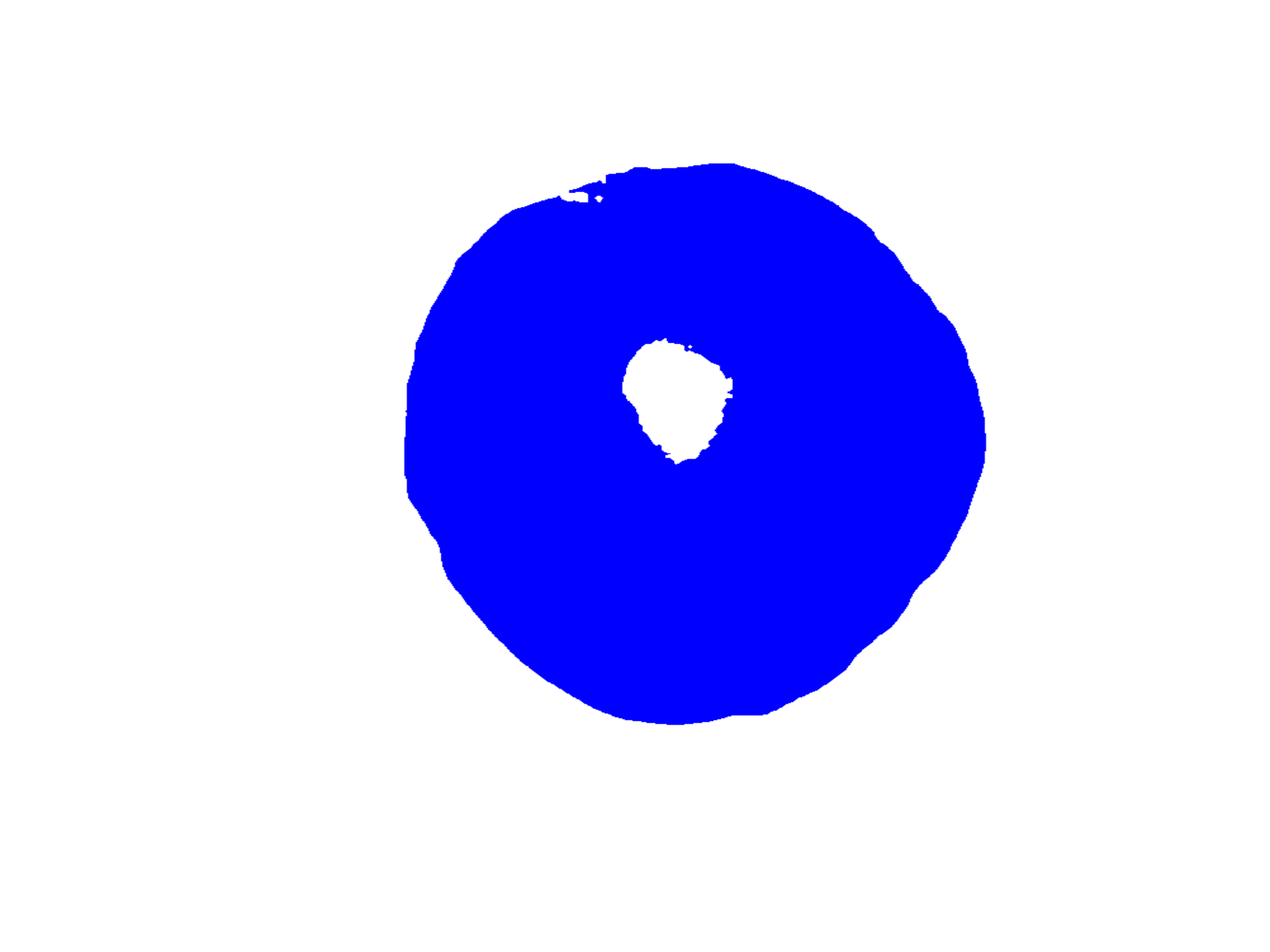} & \includegraphics[width=0.15\textwidth]{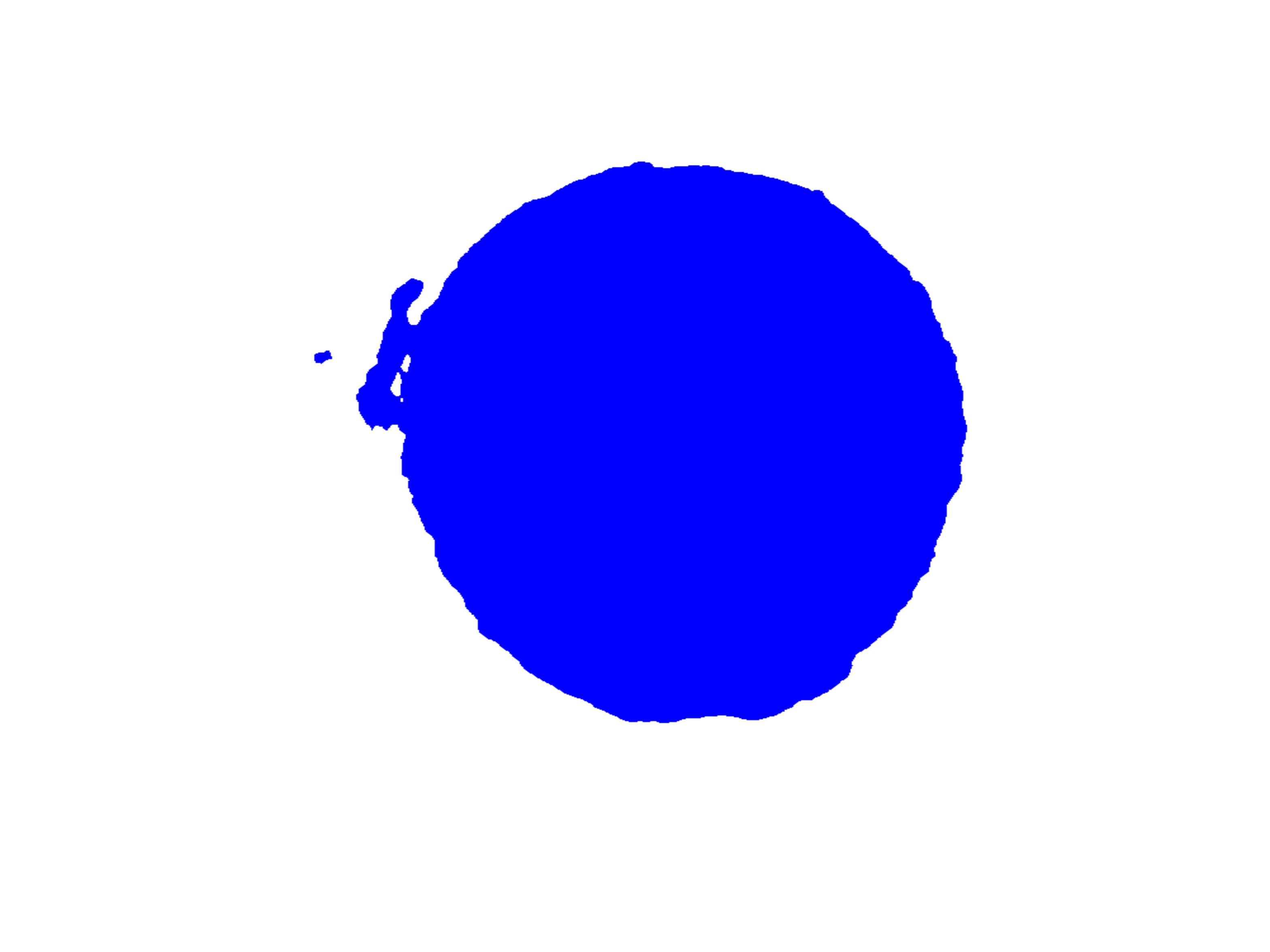} & \includegraphics[width=0.15\textwidth]{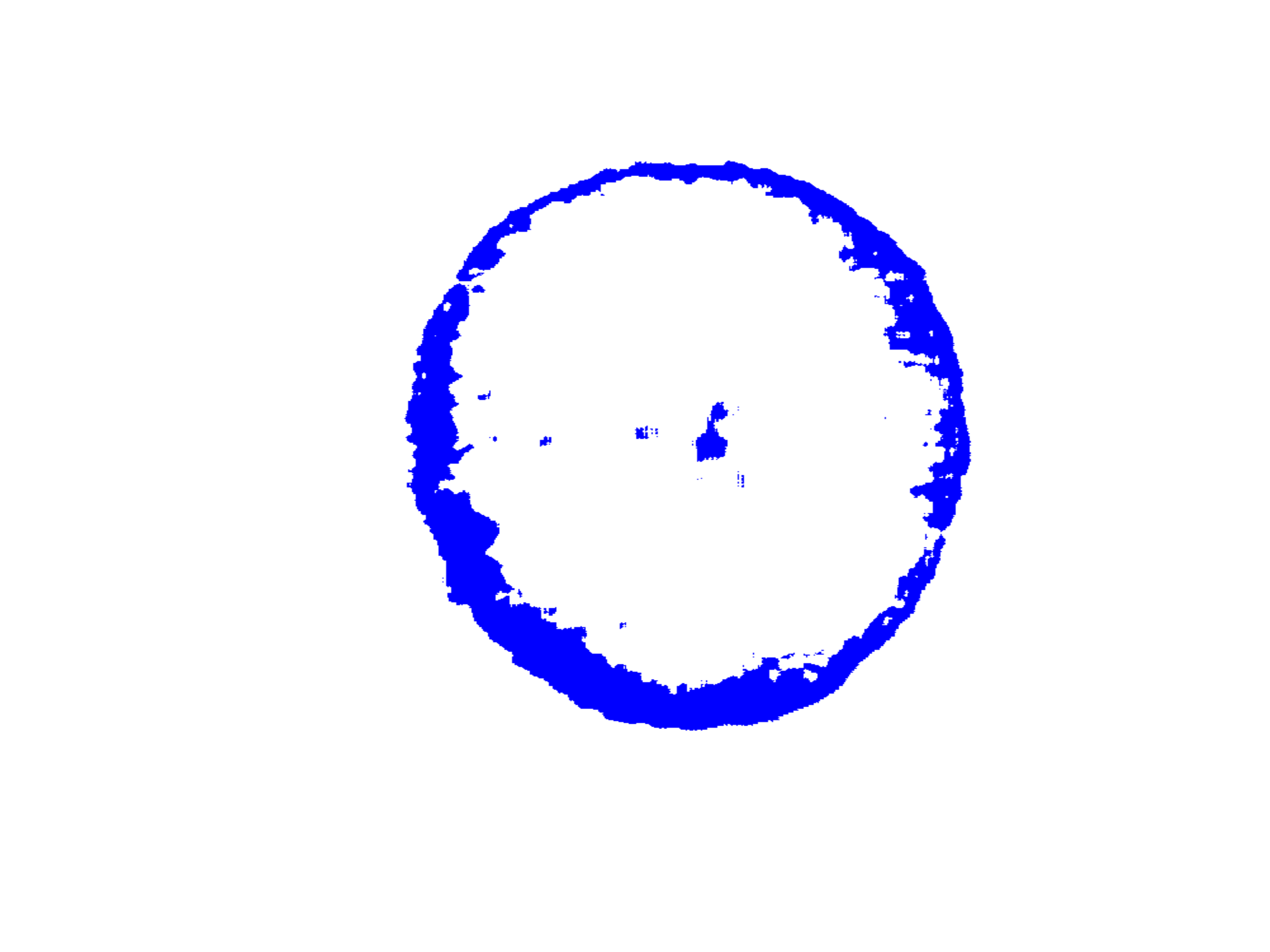} & \includegraphics[width=0.15\textwidth]{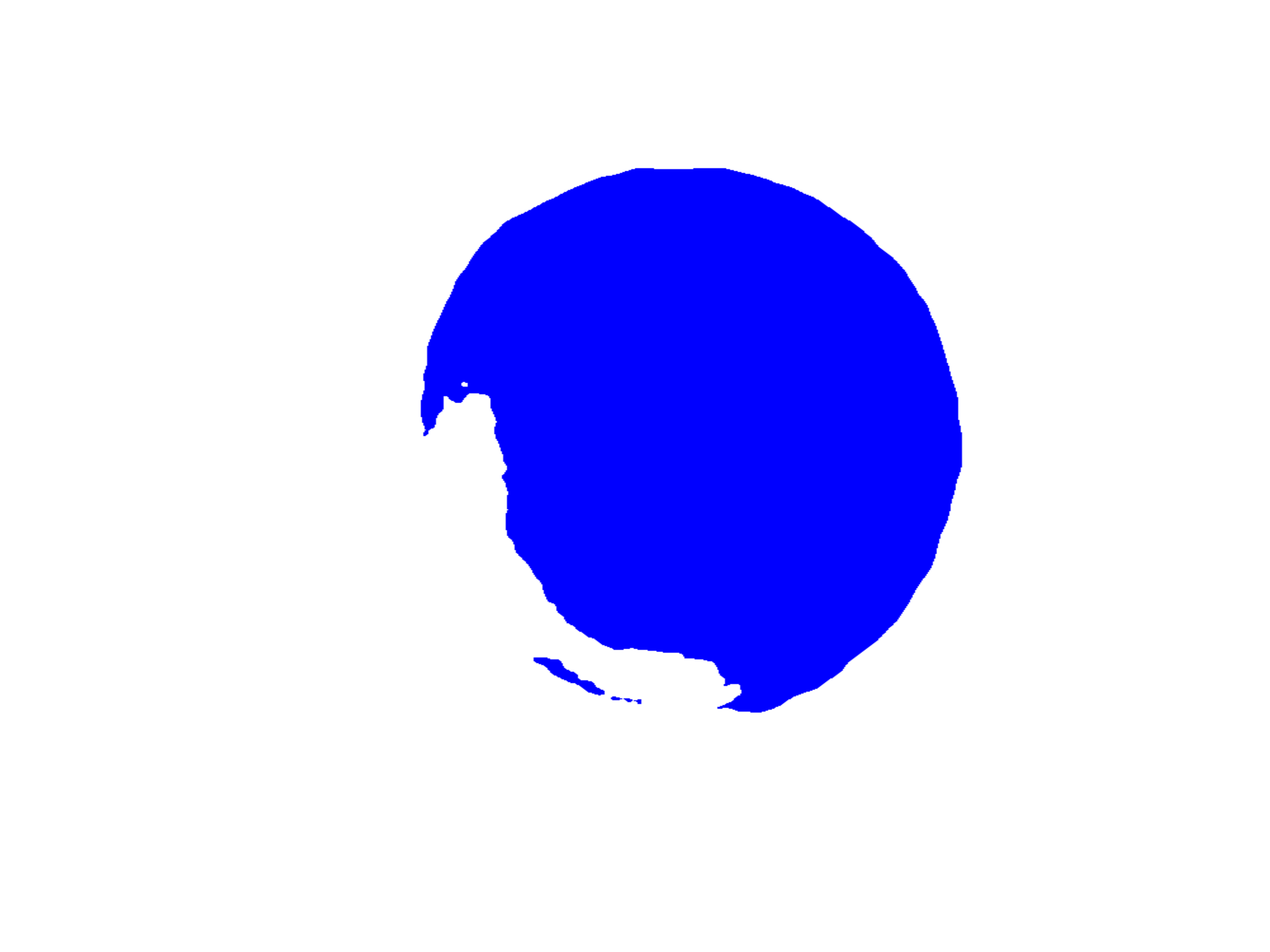} \\ 
\centering NST                & \includegraphics[width=0.15\textwidth]{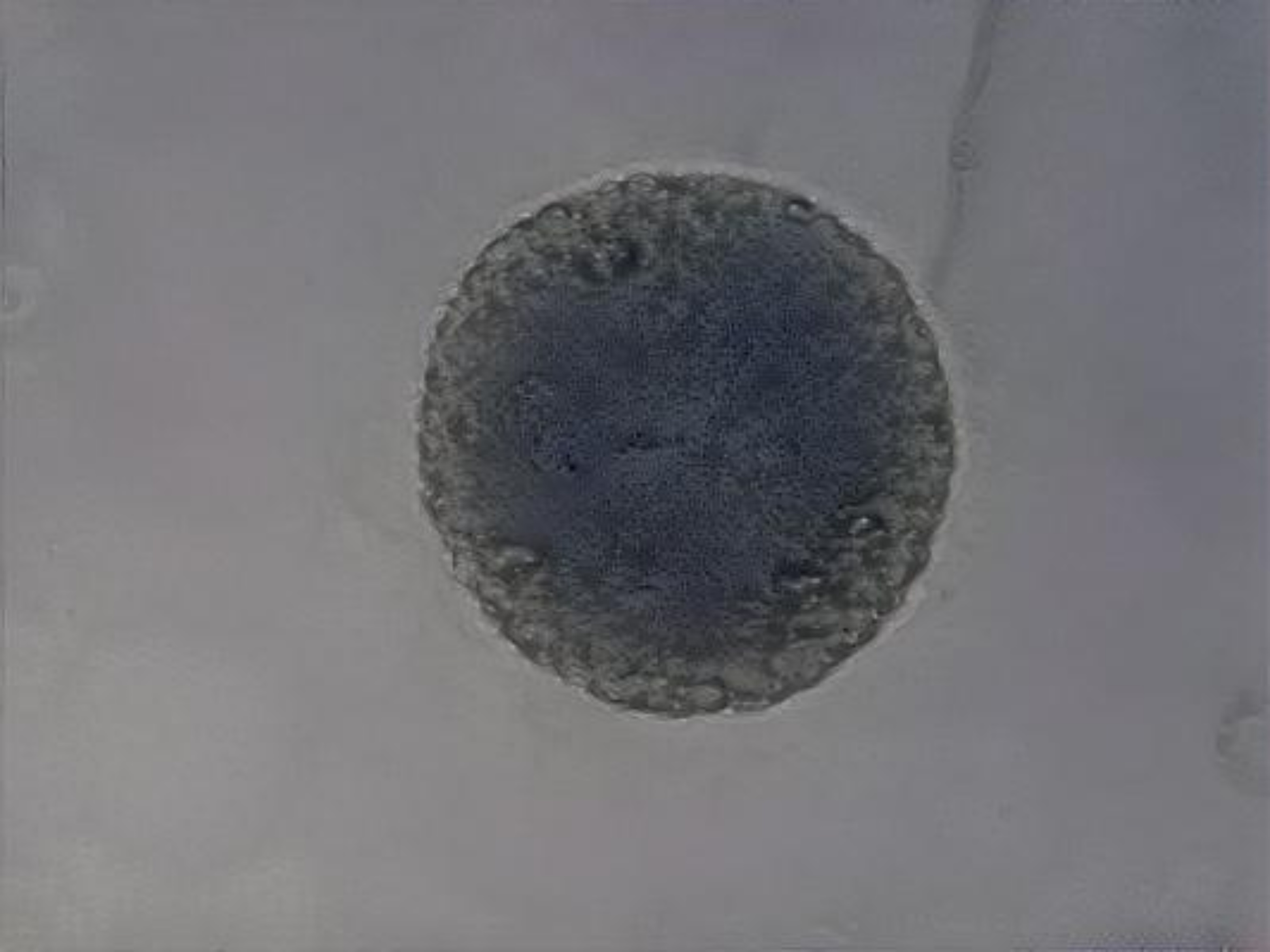}   &
\includegraphics[width=0.15\textwidth]{results/truth.jpg}&
\includegraphics[width=0.15\textwidth]{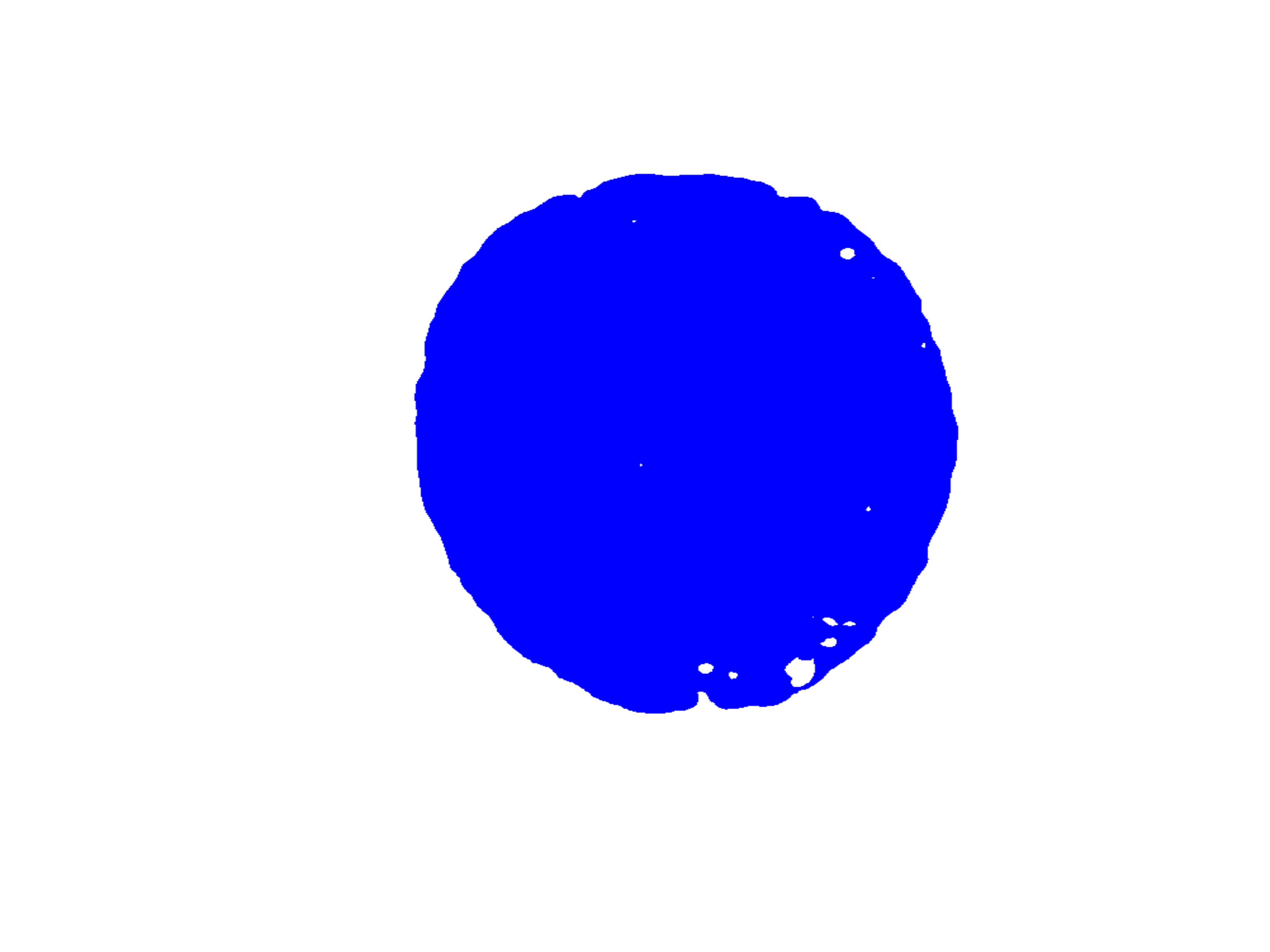}   & \includegraphics[width=0.15\textwidth]{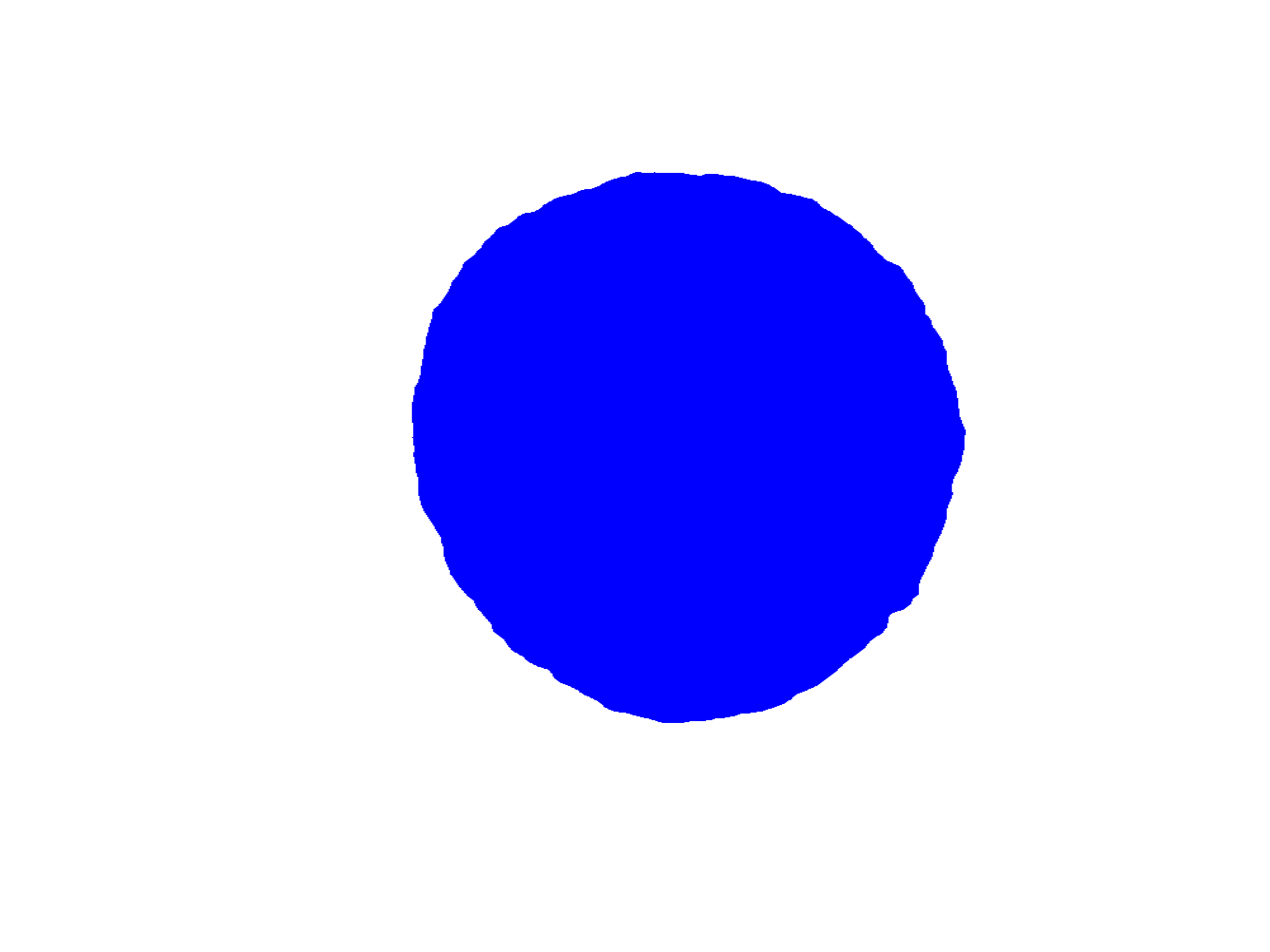}   & \includegraphics[width=0.15\textwidth]{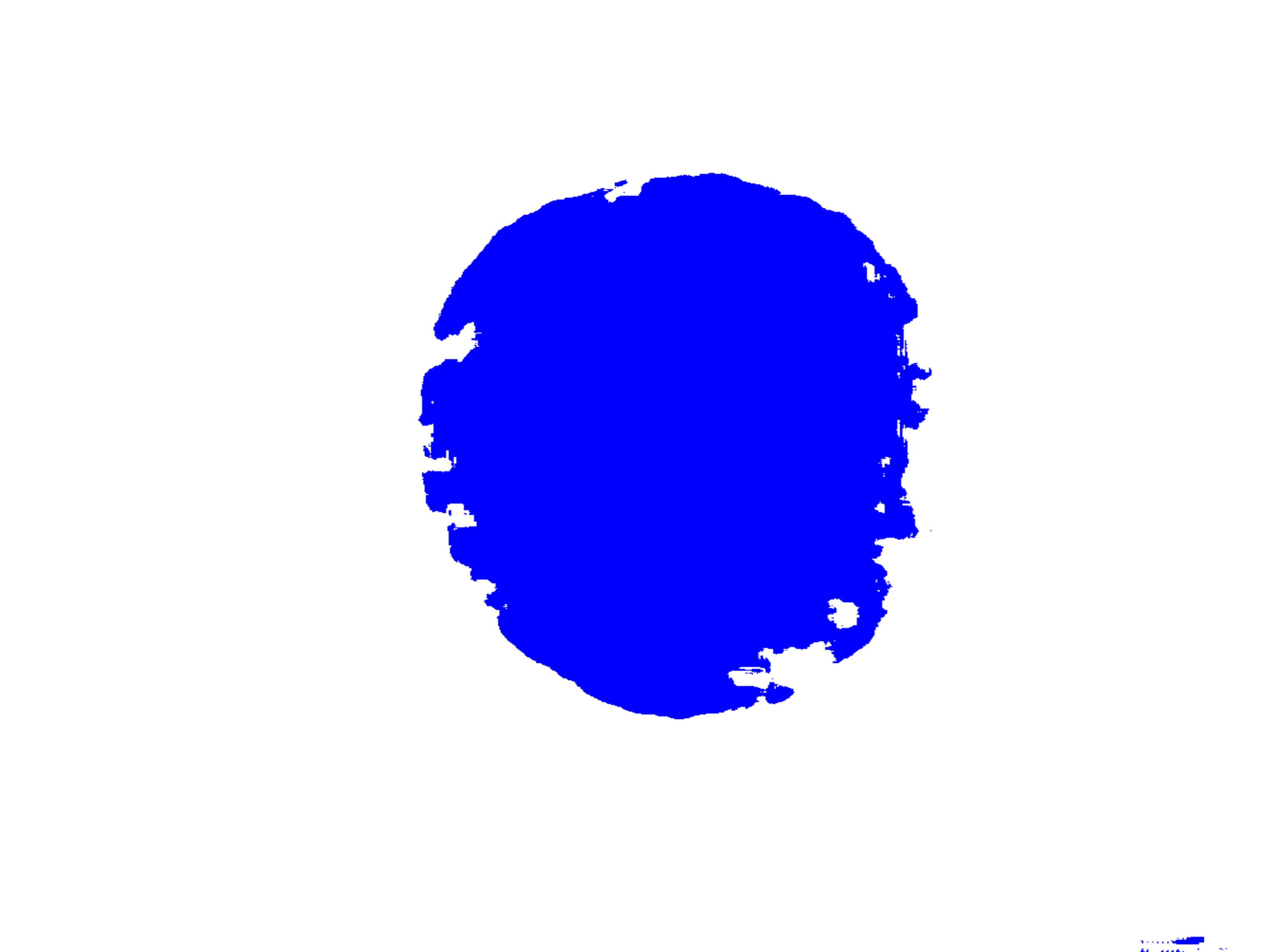}   & \includegraphics[width=0.15\textwidth]{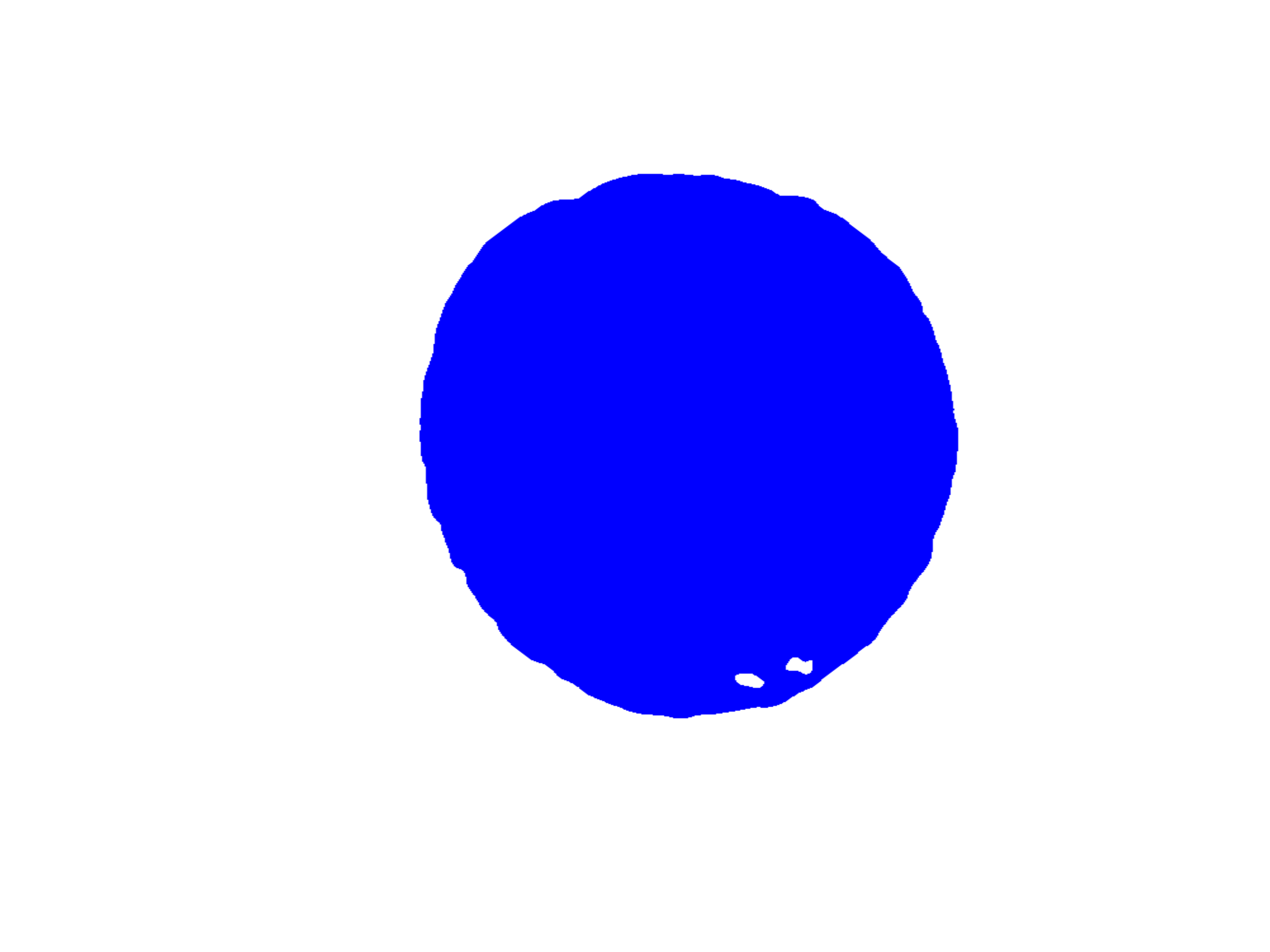}   \\
\centering Deep Image Analogy &  \includegraphics[width=0.15\textwidth]{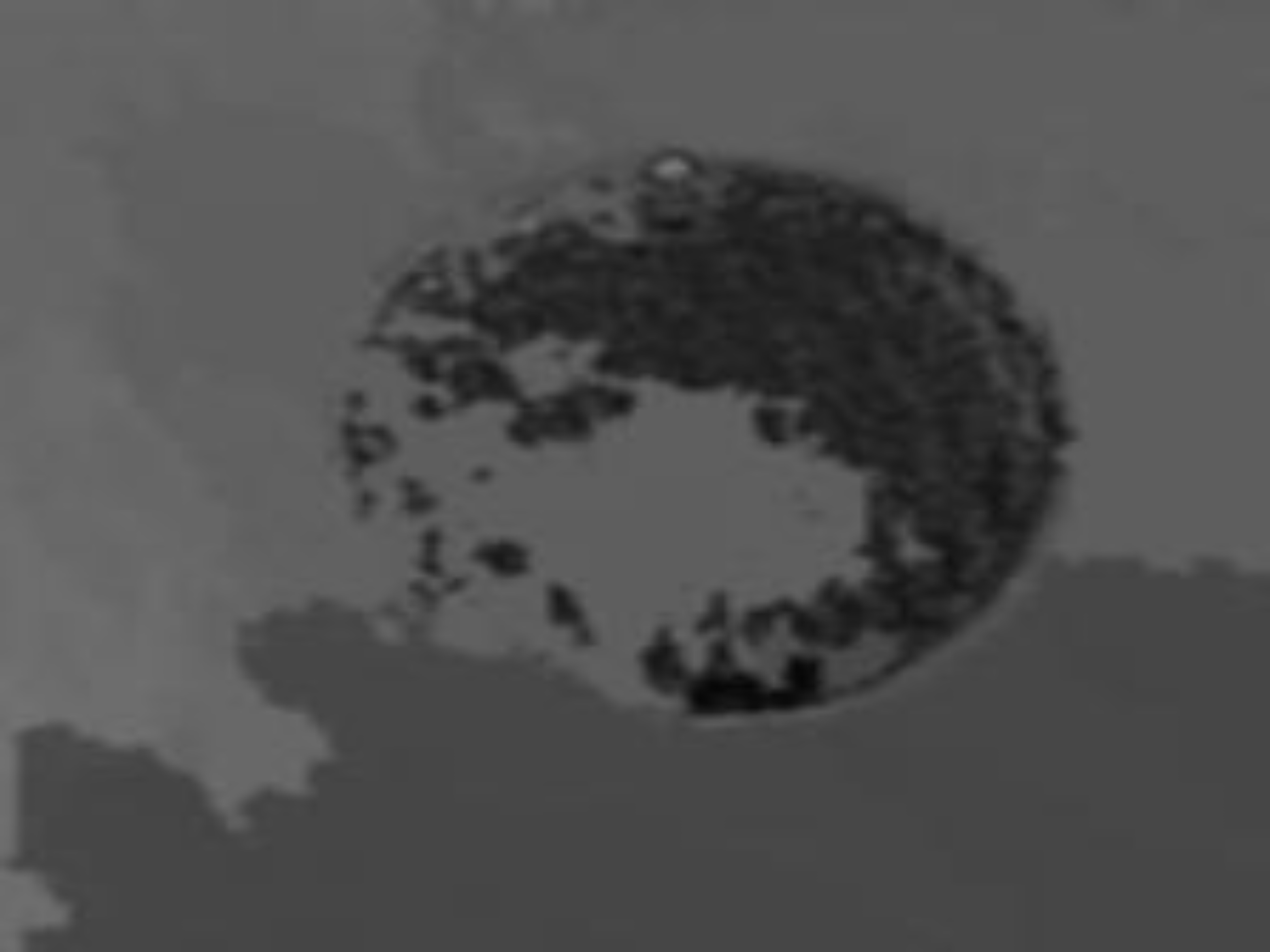} &
\includegraphics[width=0.15\textwidth]{results/truth.jpg}&
\includegraphics[width=0.15\textwidth]{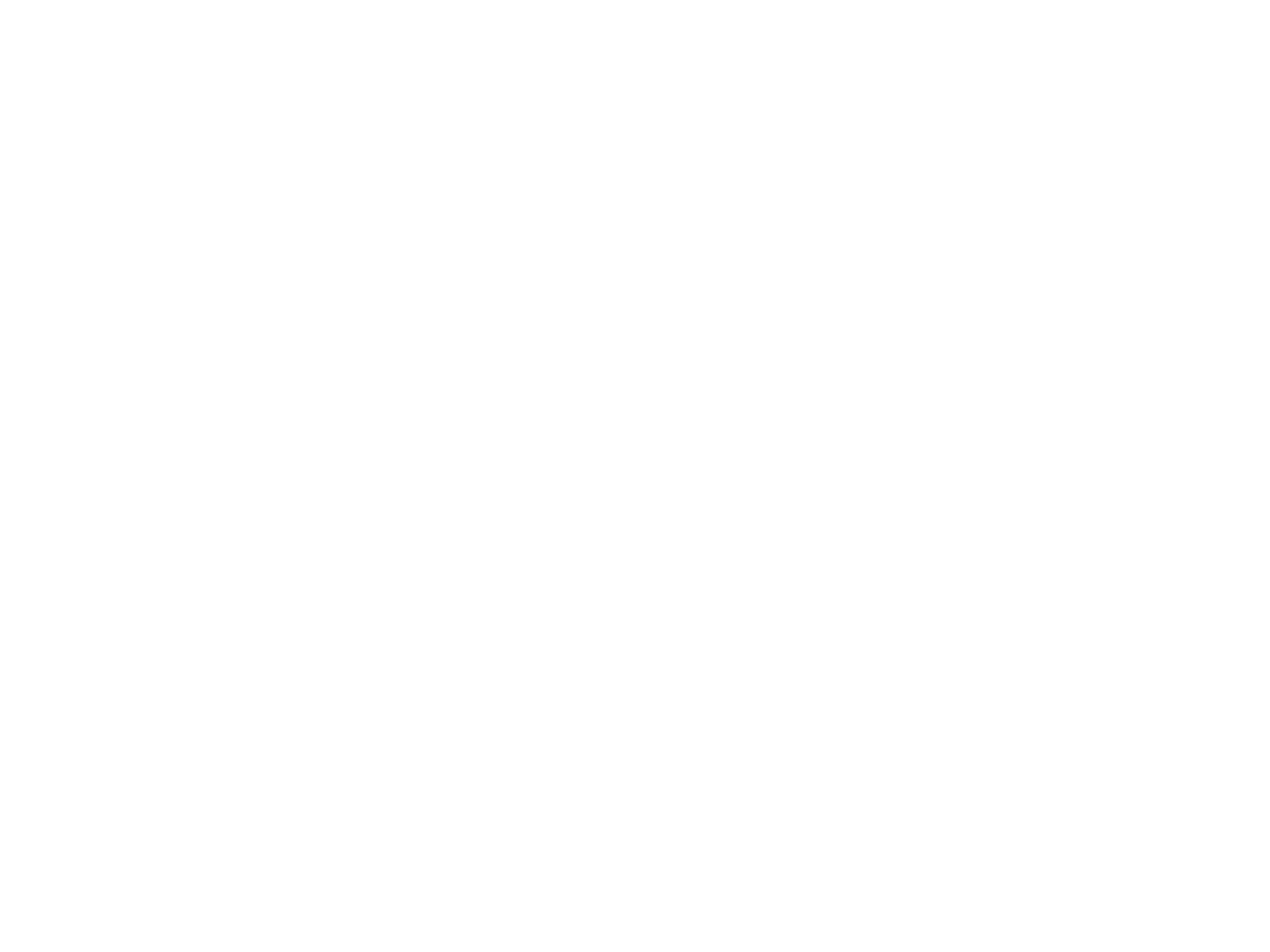}& \includegraphics[width=0.15\textwidth]{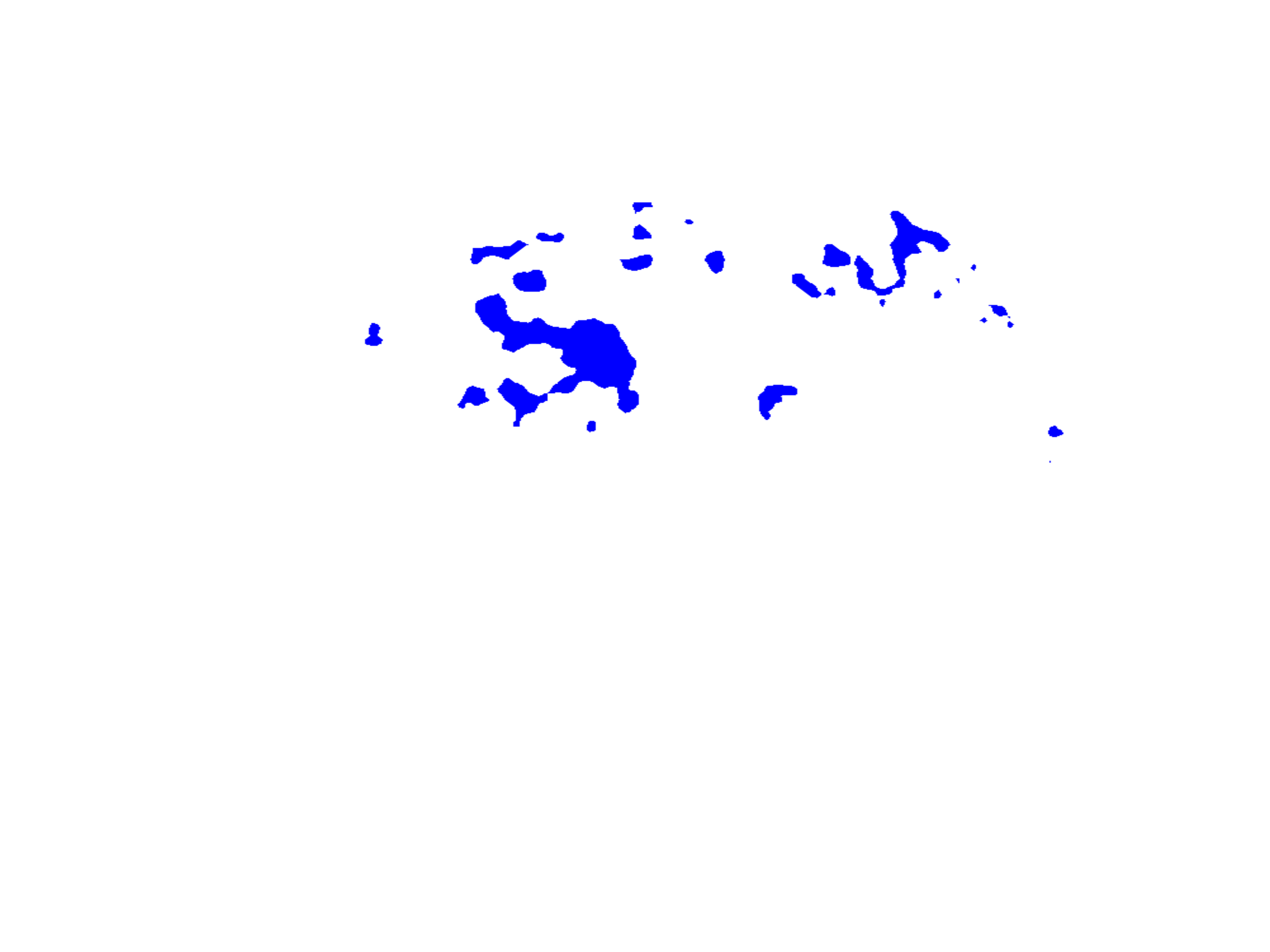}& \includegraphics[width=0.15\textwidth]{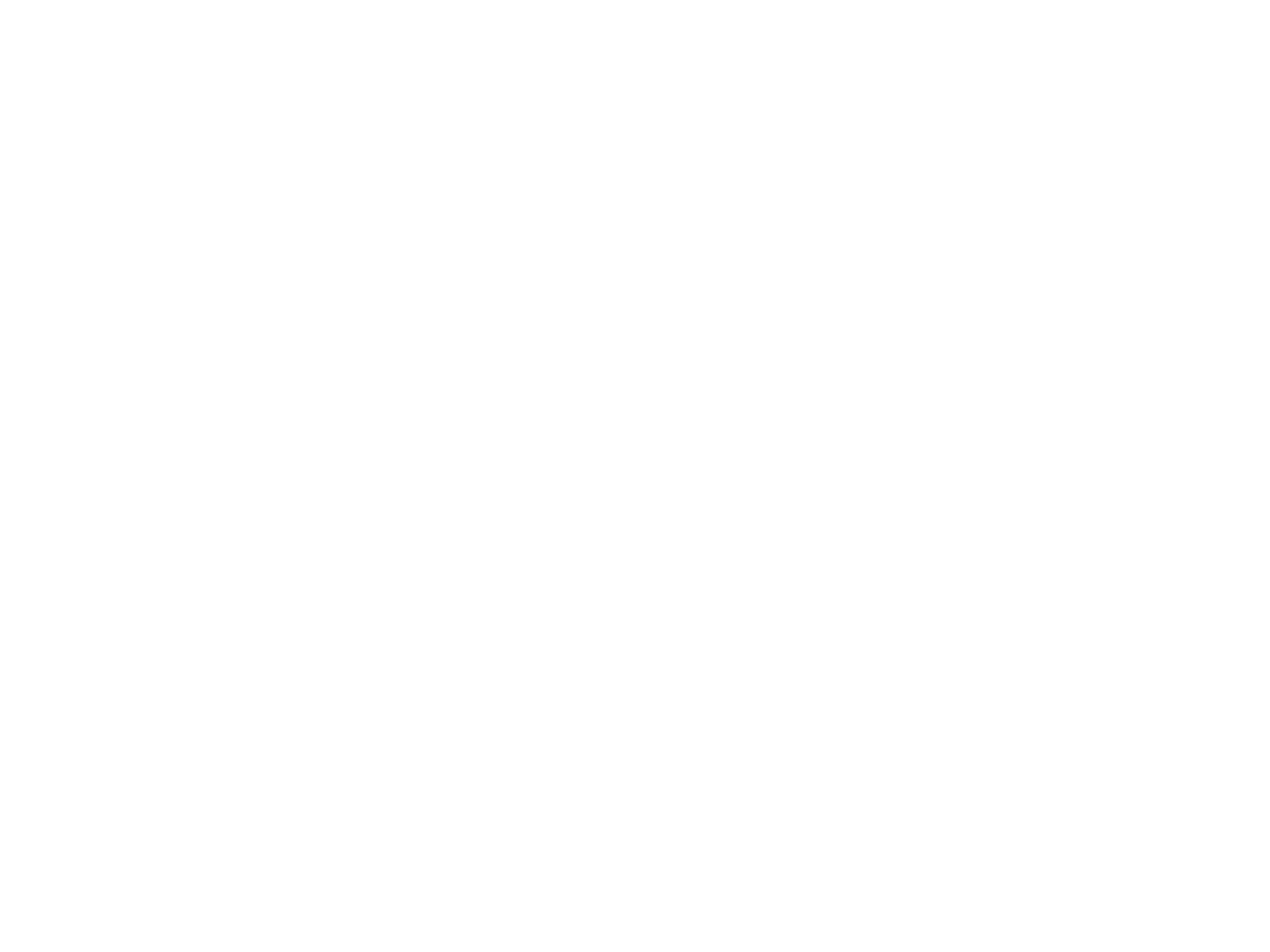} & \includegraphics[width=0.15\textwidth]{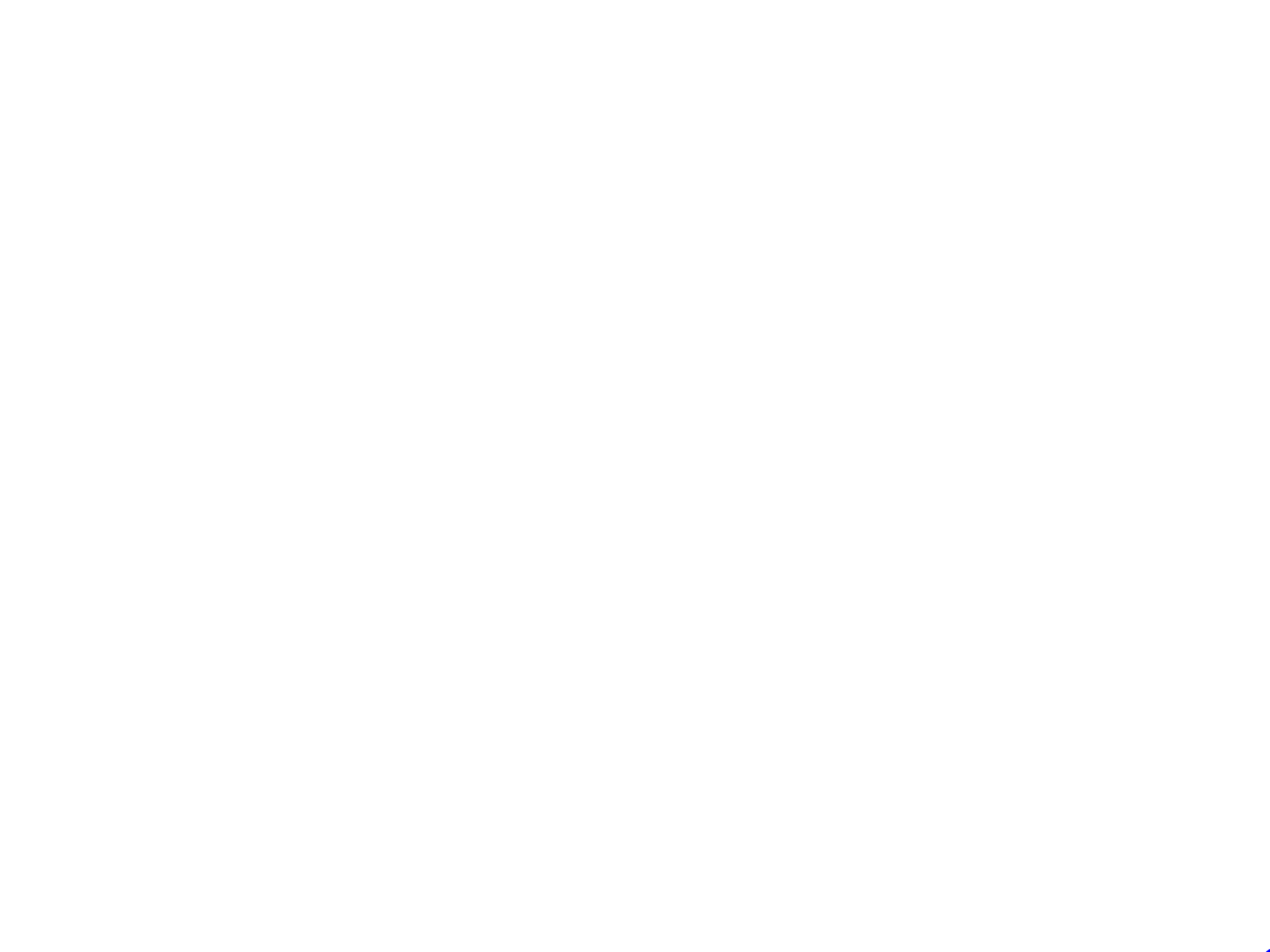} \\ 
\centering STROTSS            & \includegraphics[width=0.15\textwidth]{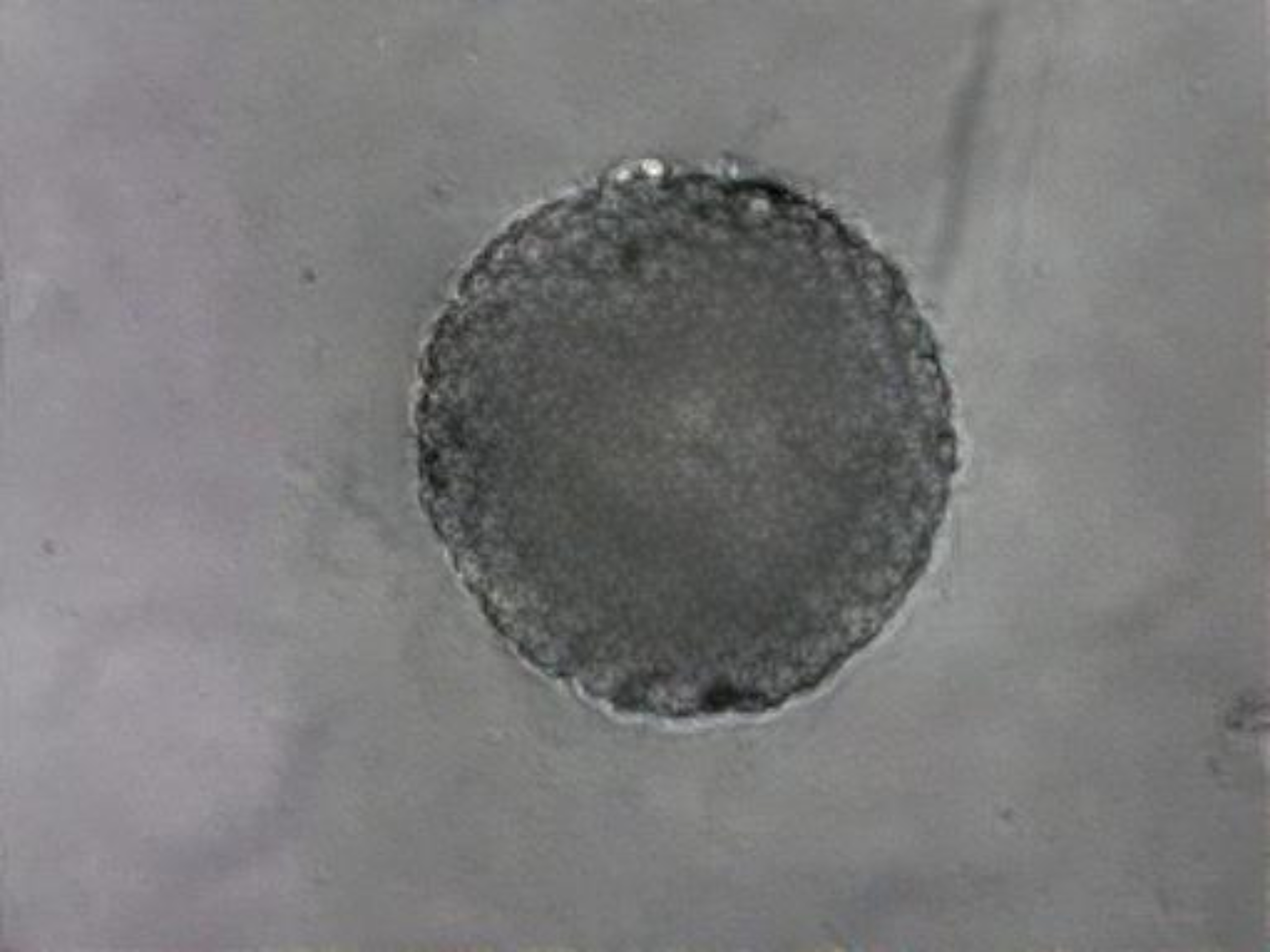}  &
\includegraphics[width=0.15\textwidth]{results/truth.jpg}&
\includegraphics[width=0.15\textwidth]{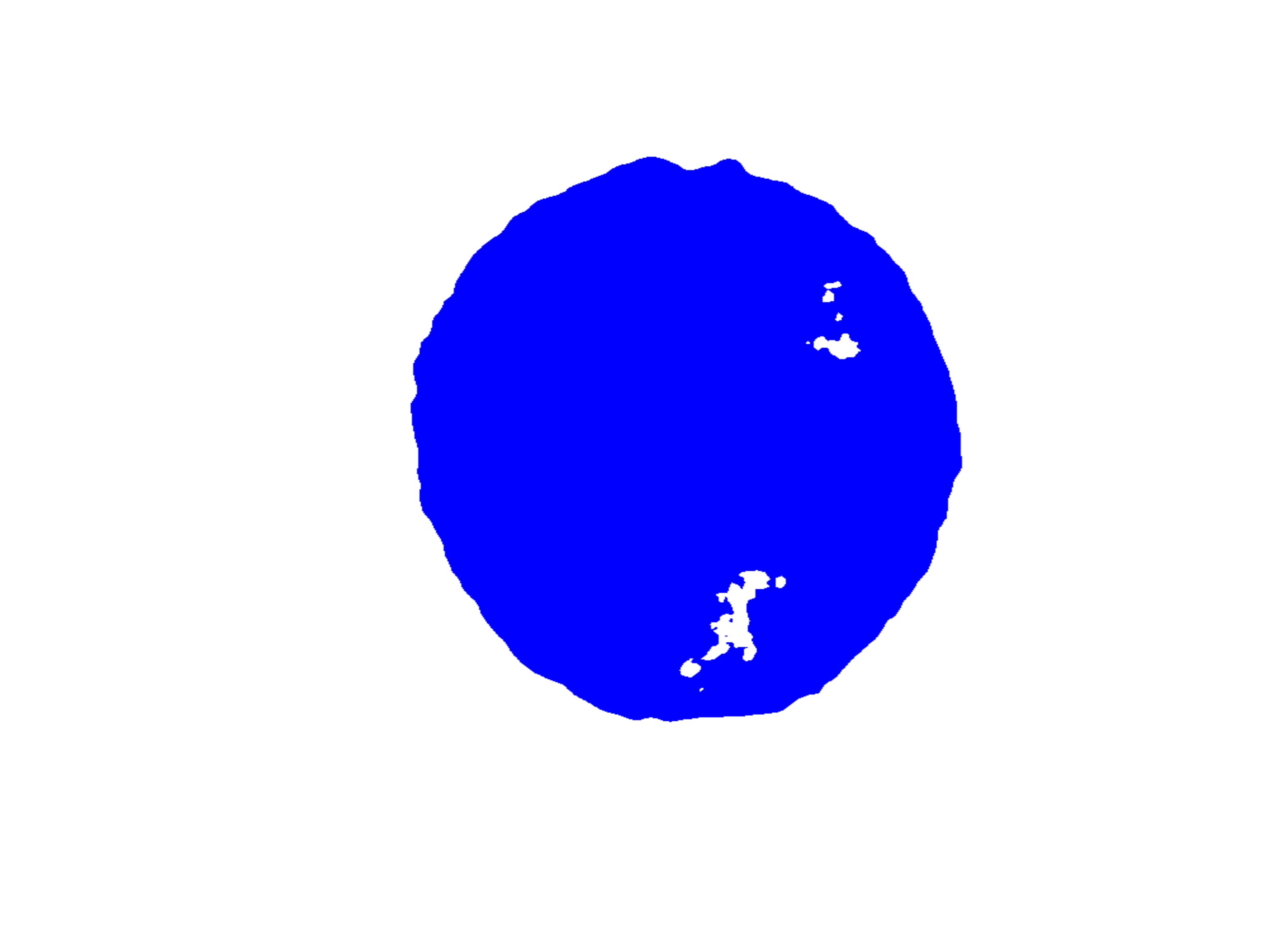}  & \includegraphics[width=0.15\textwidth]{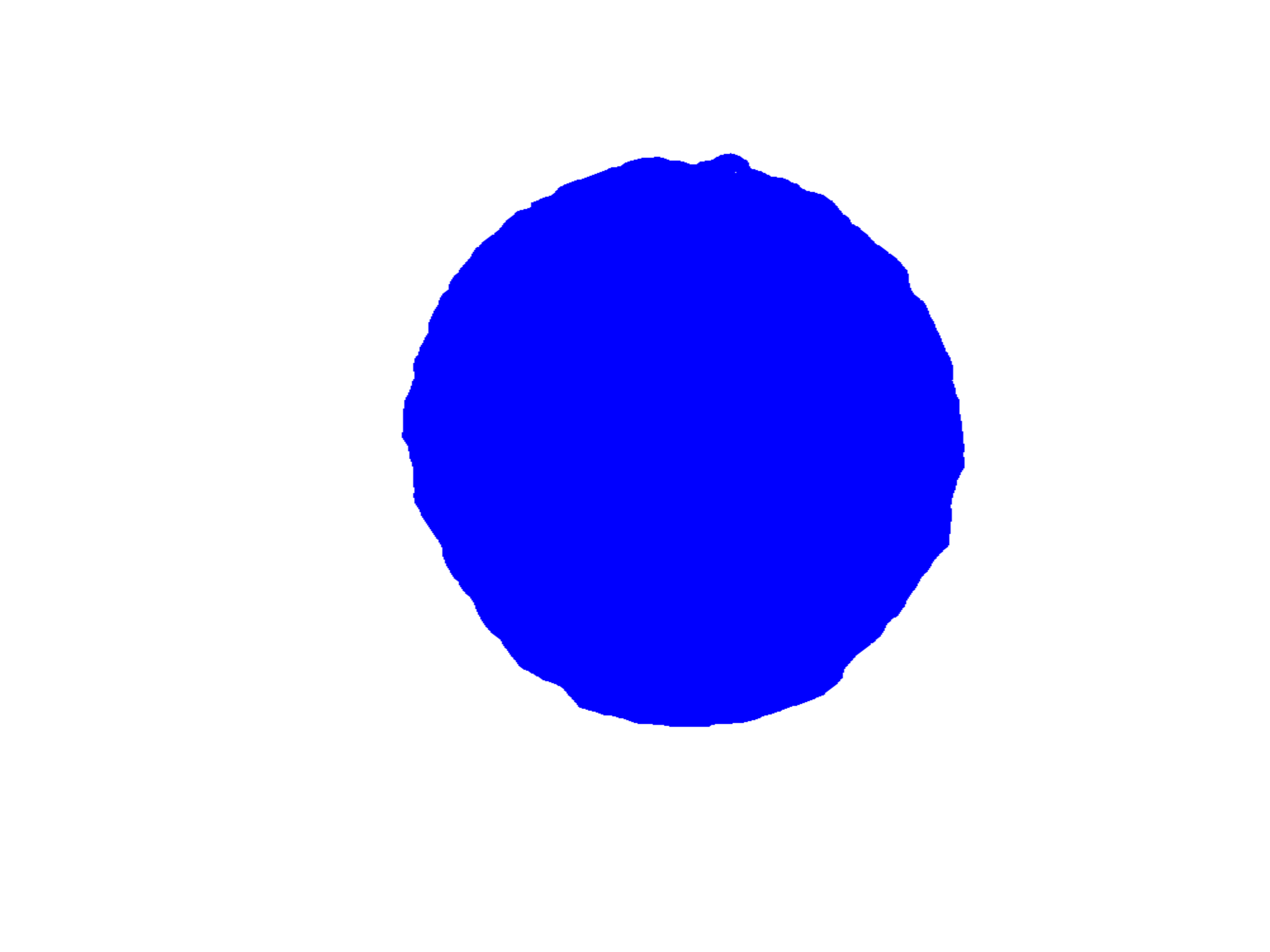}  & \includegraphics[width=0.15\textwidth]{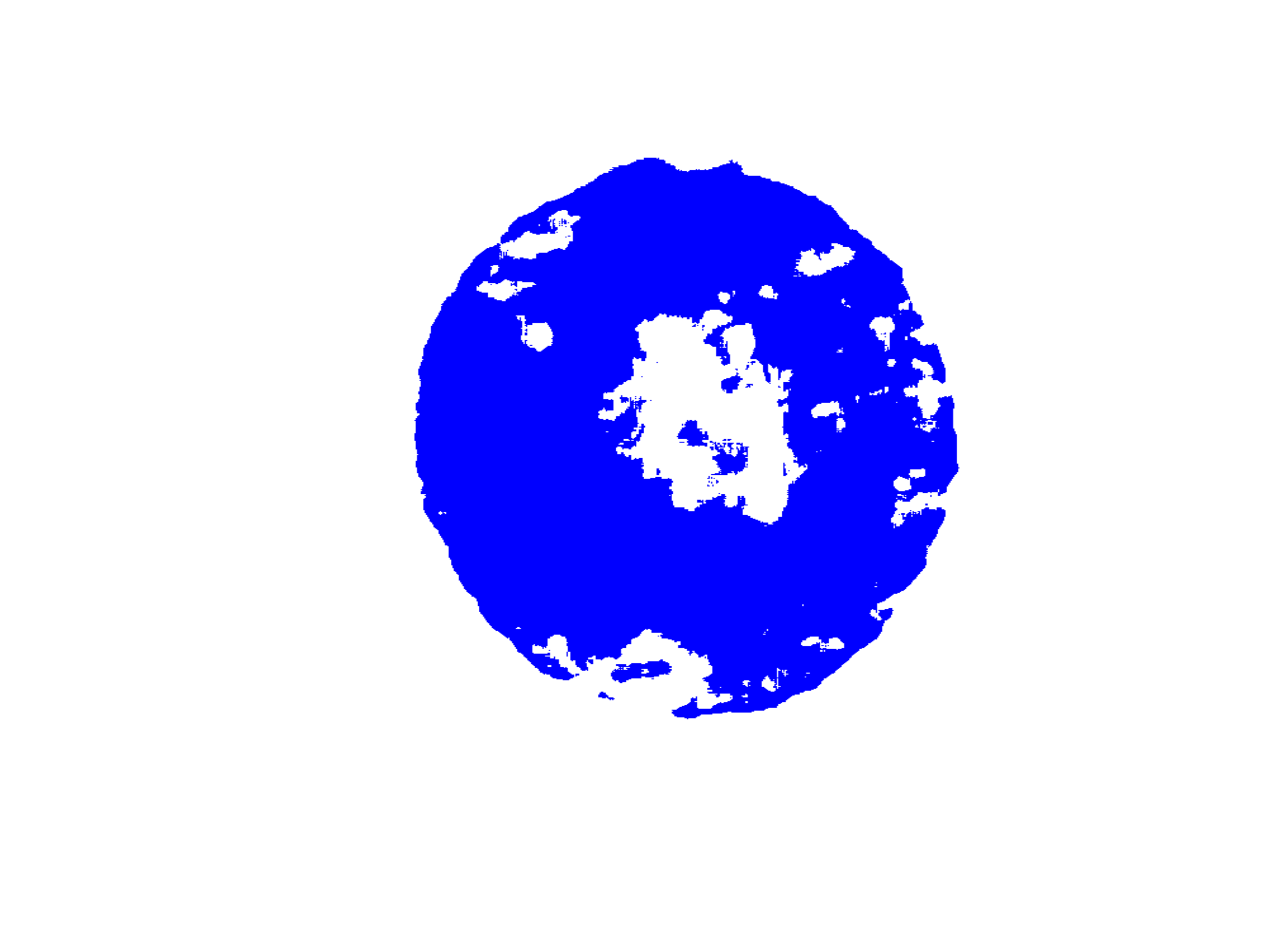}  & \includegraphics[width=0.15\textwidth]{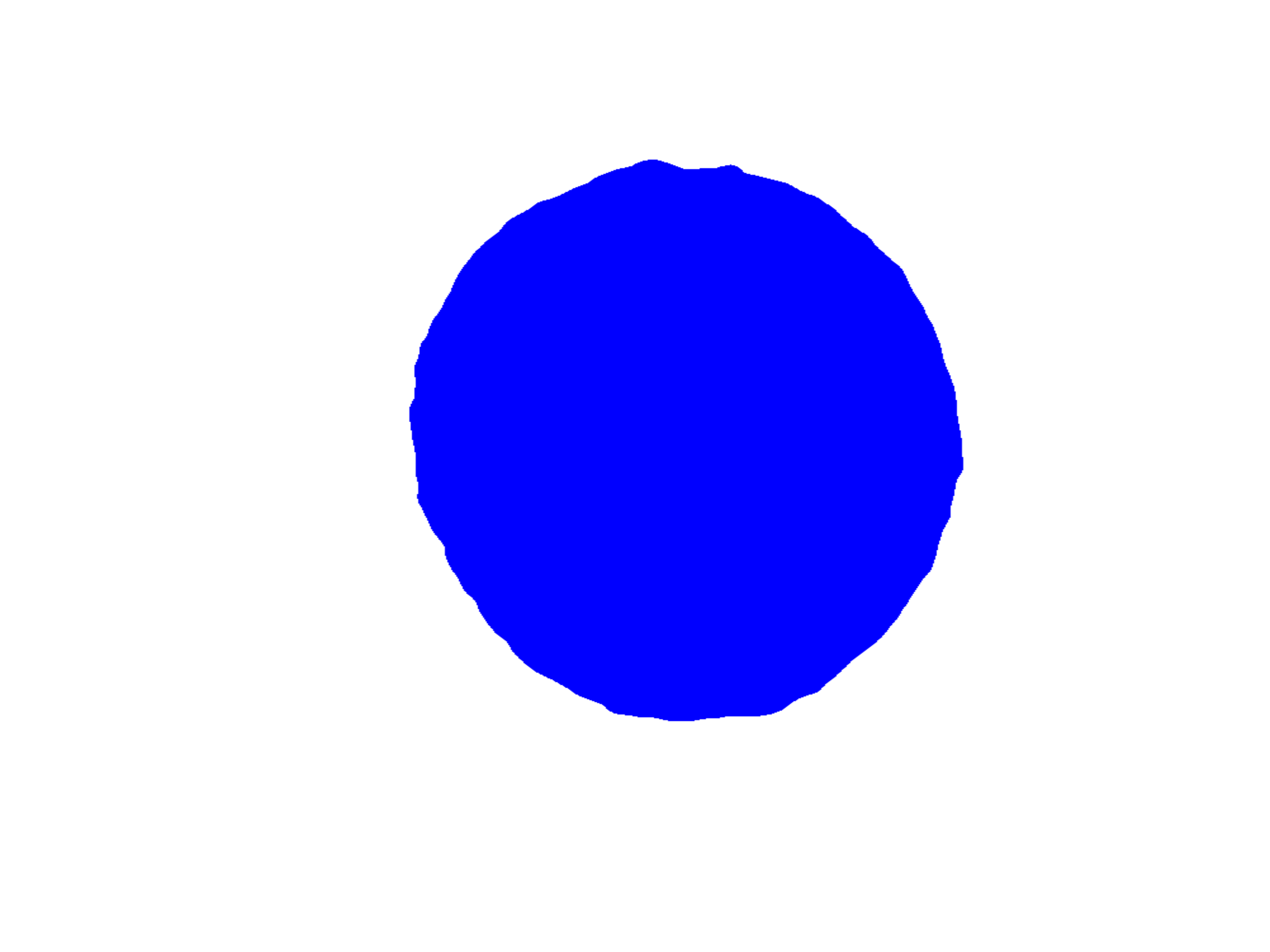}  \\ 
\centering CycleGAN           & \includegraphics[width=0.15\textwidth]{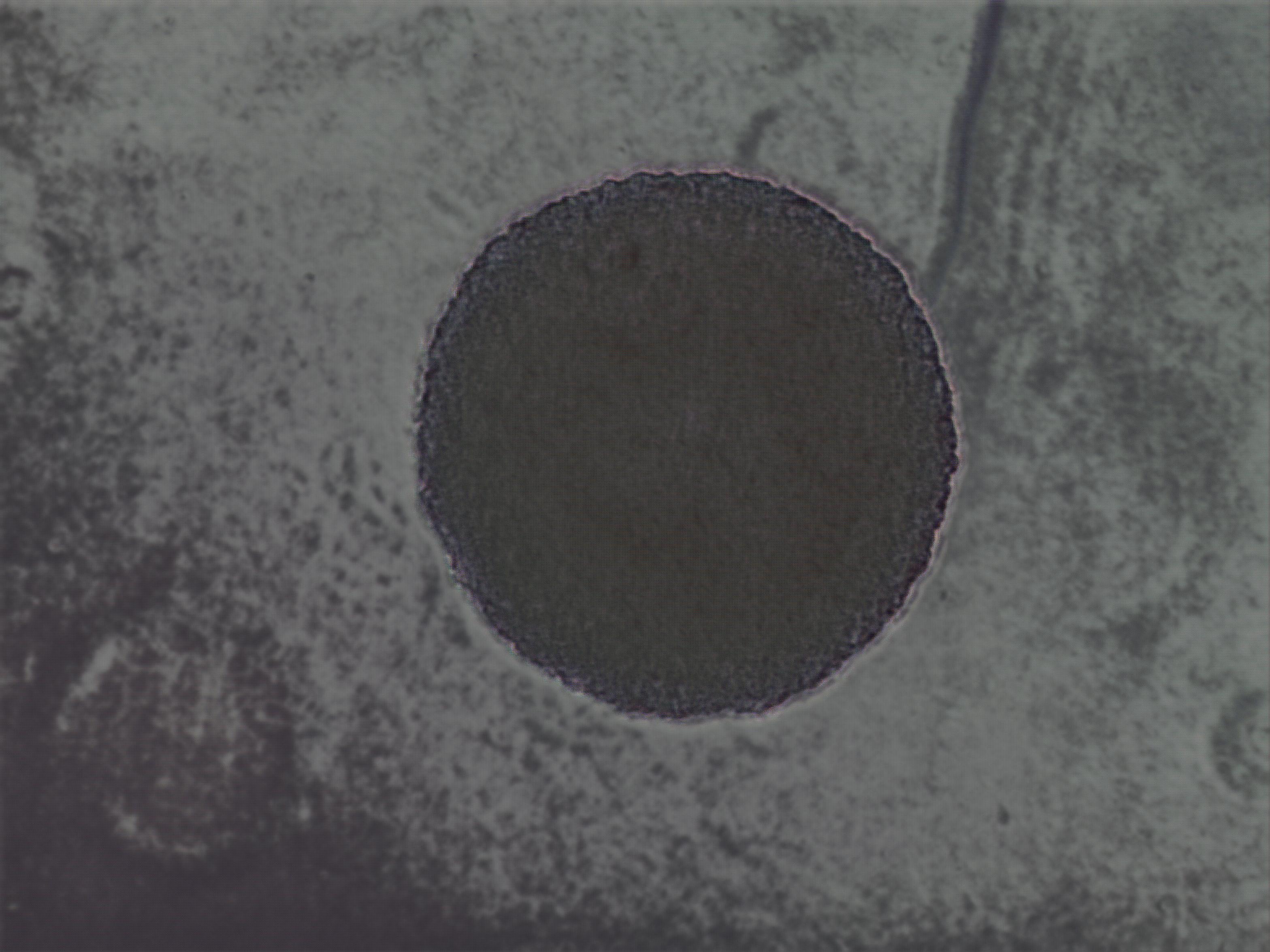} &
\includegraphics[width=0.15\textwidth]{results/truth.jpg}&
\includegraphics[width=0.15\textwidth]{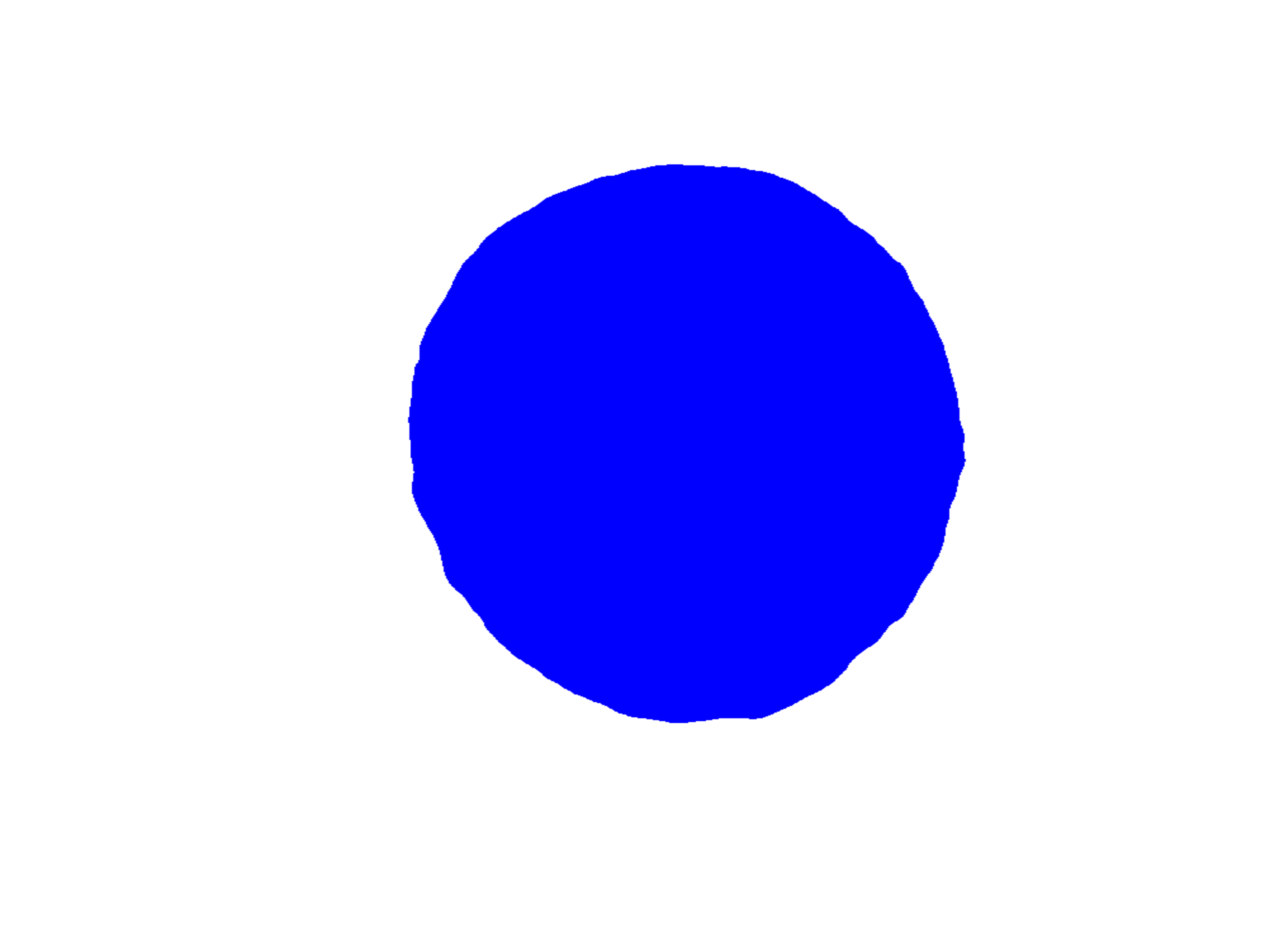} & \includegraphics[width=0.15\textwidth]{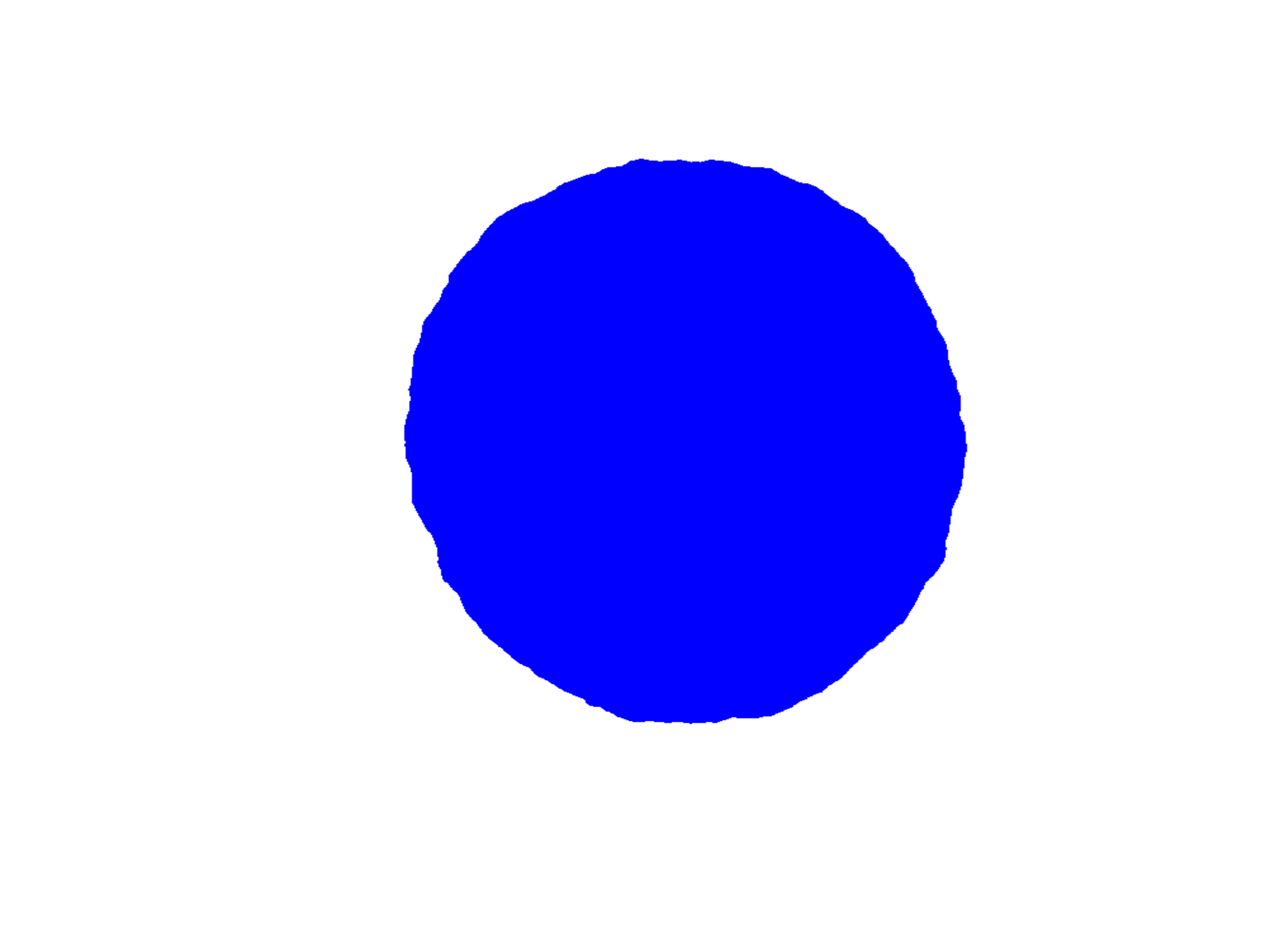} & \includegraphics[width=0.15\textwidth]{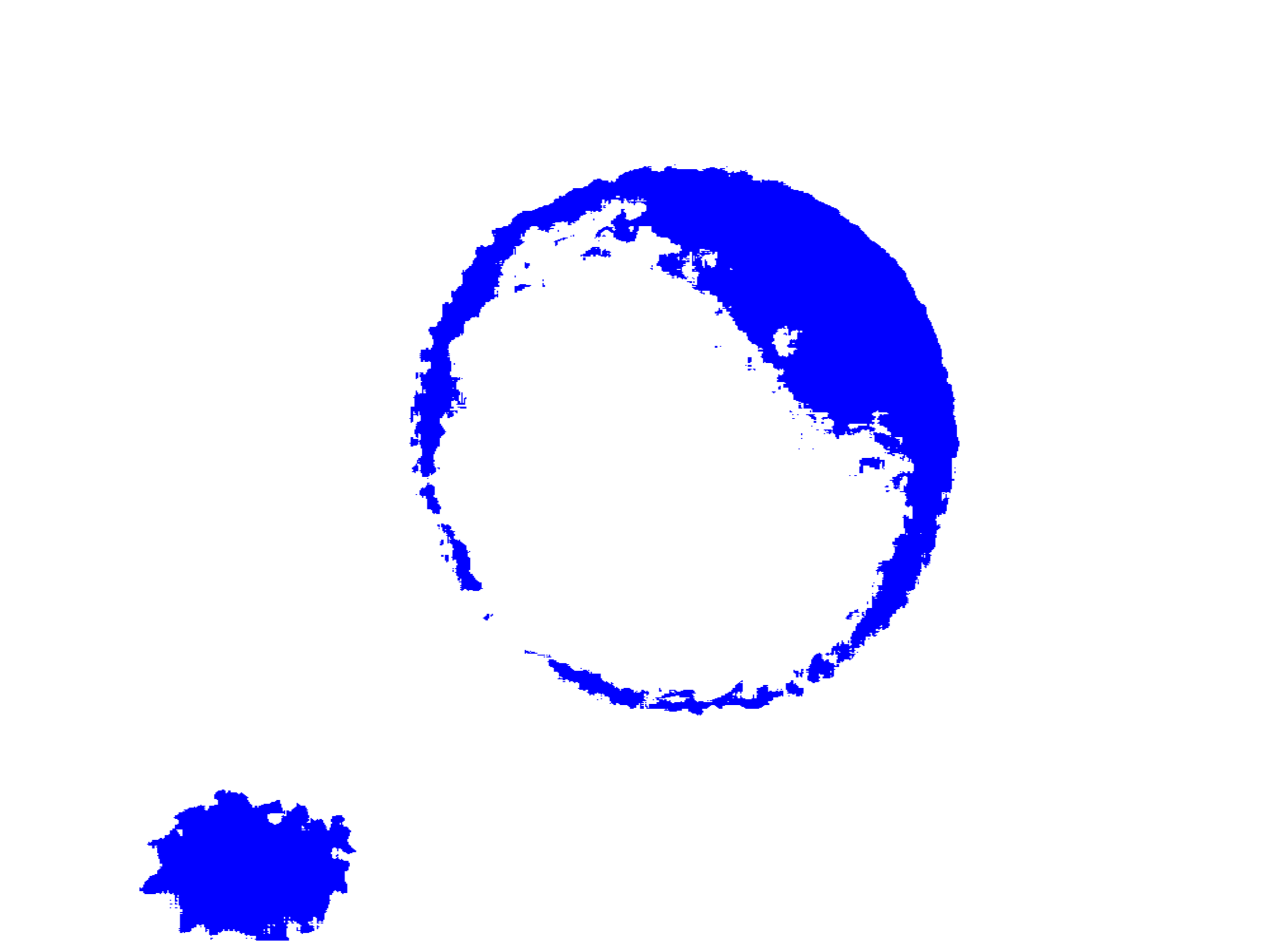} & \includegraphics[width=0.15\textwidth]{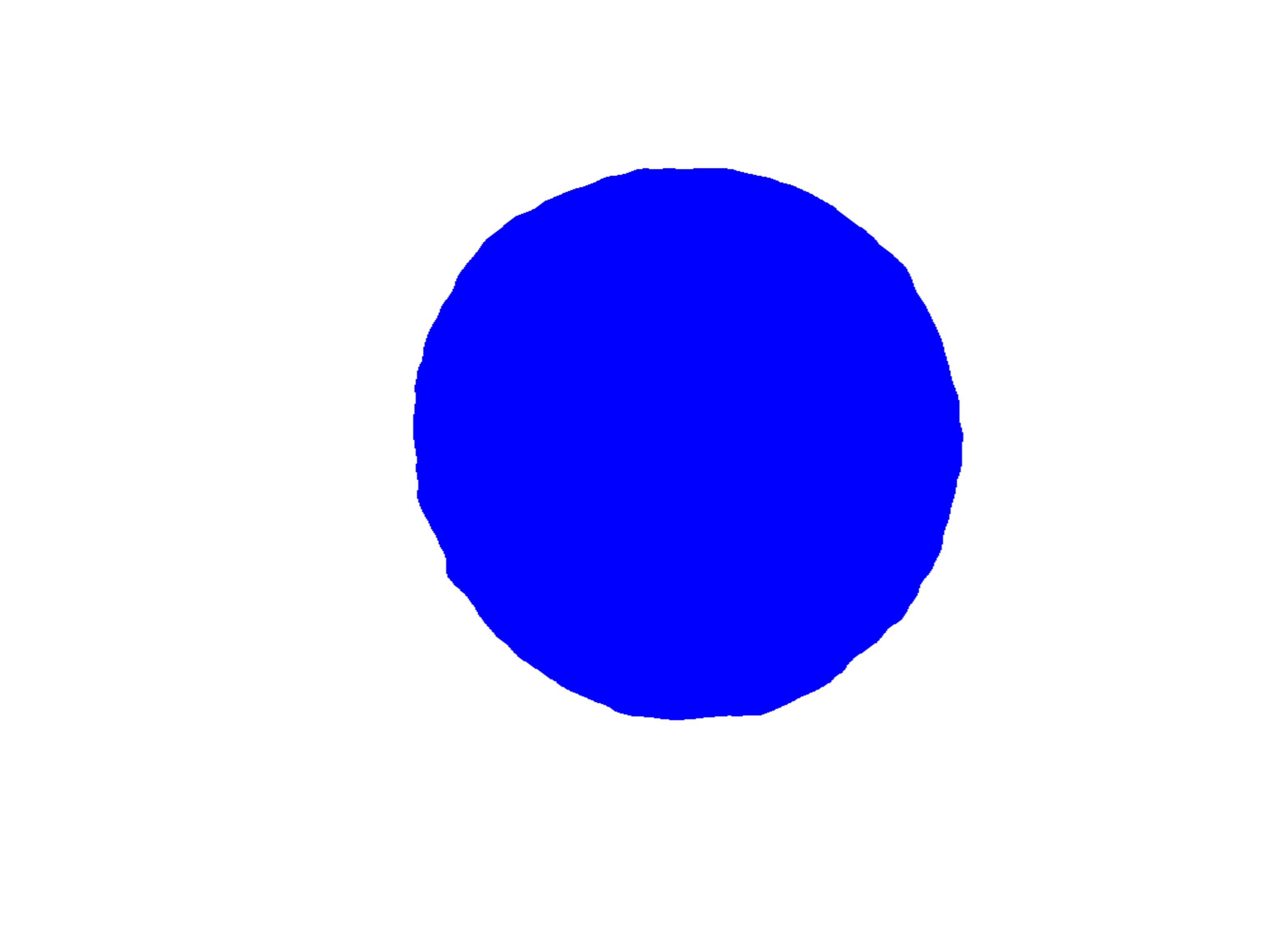} \\ 
\centering DualGAN            & \includegraphics[width=0.15\textwidth]{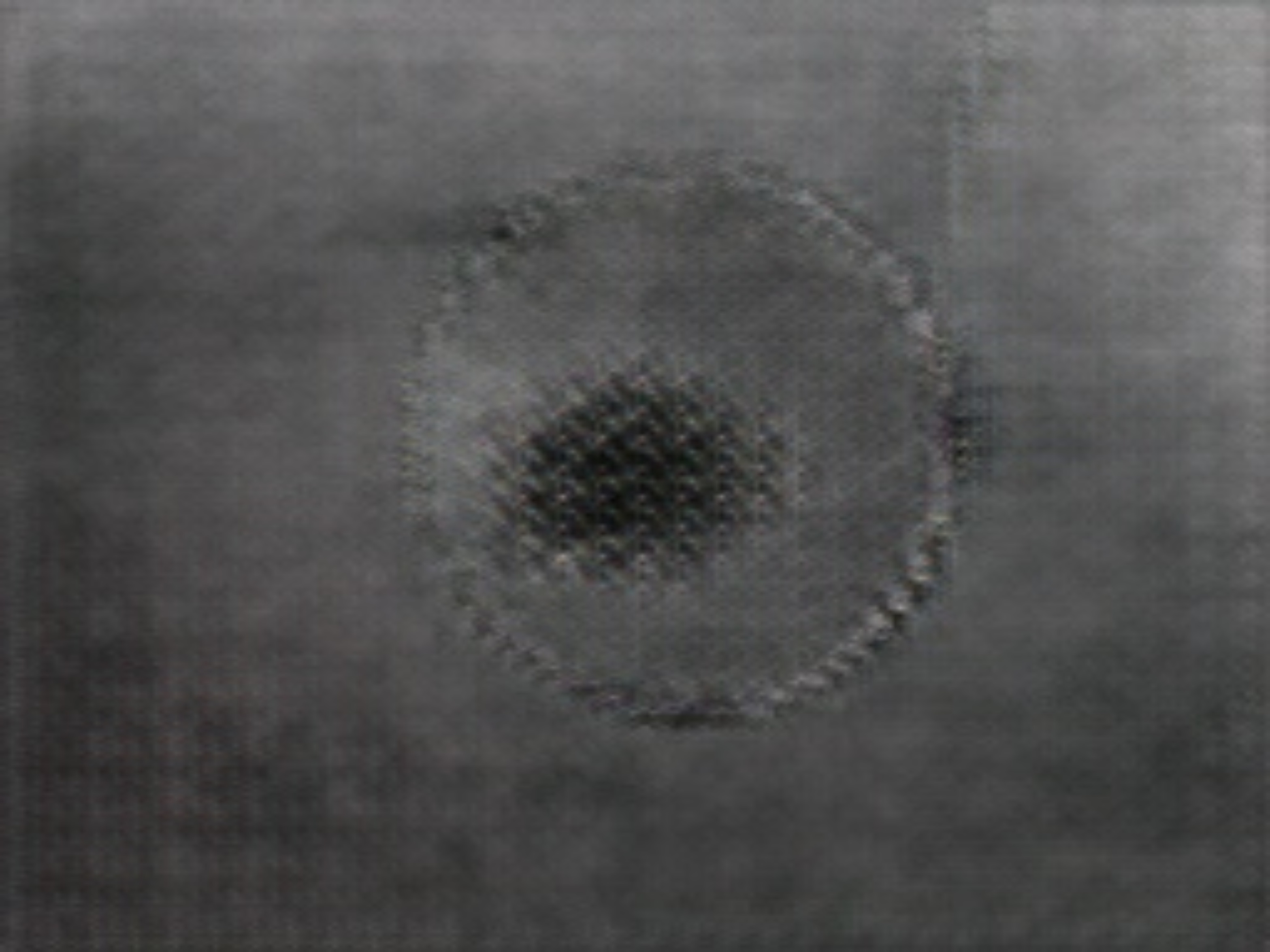}  &
\includegraphics[width=0.15\textwidth]{results/truth.jpg}&
\includegraphics[width=0.15\textwidth]{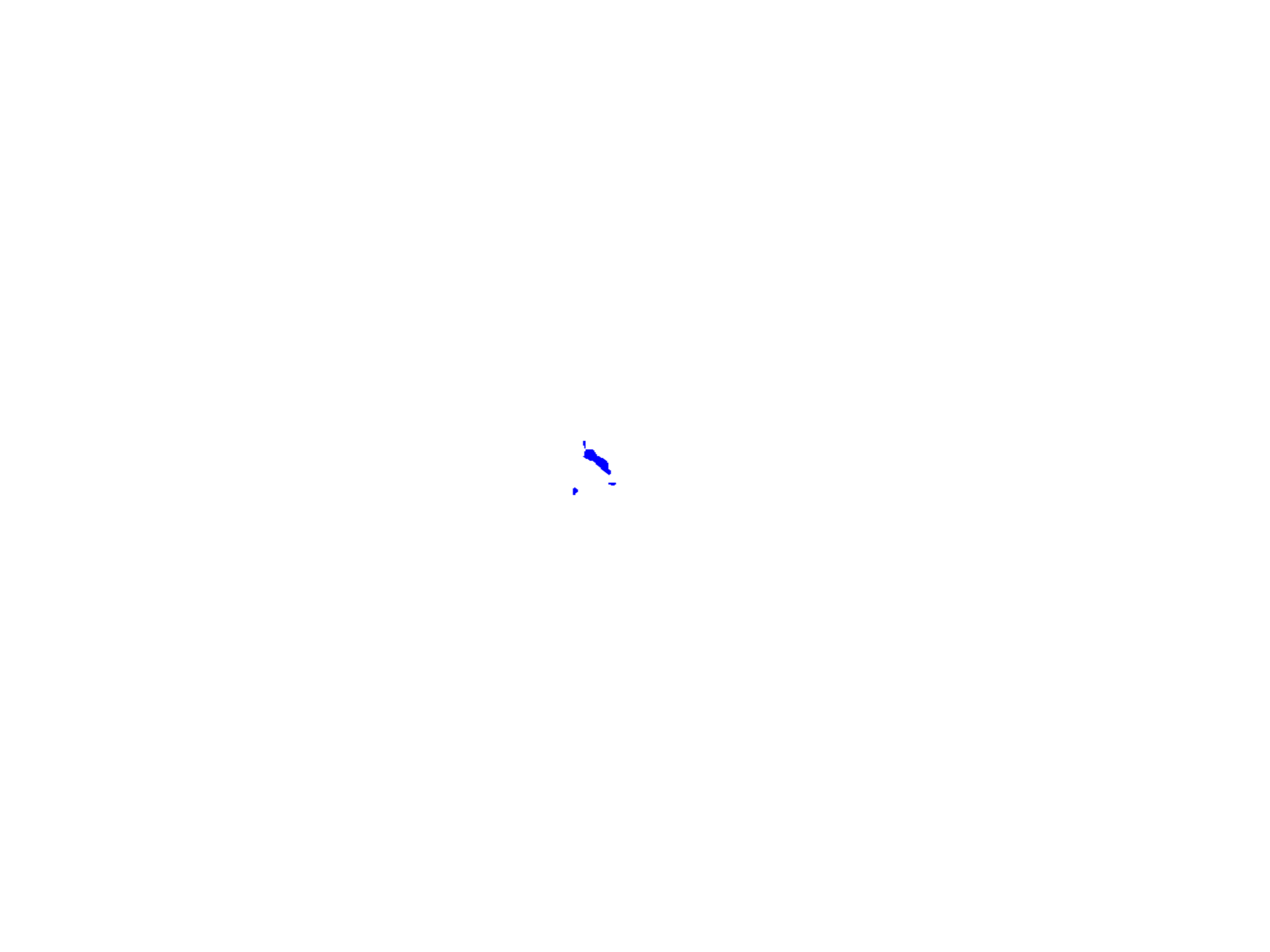}  & \includegraphics[width=0.15\textwidth]{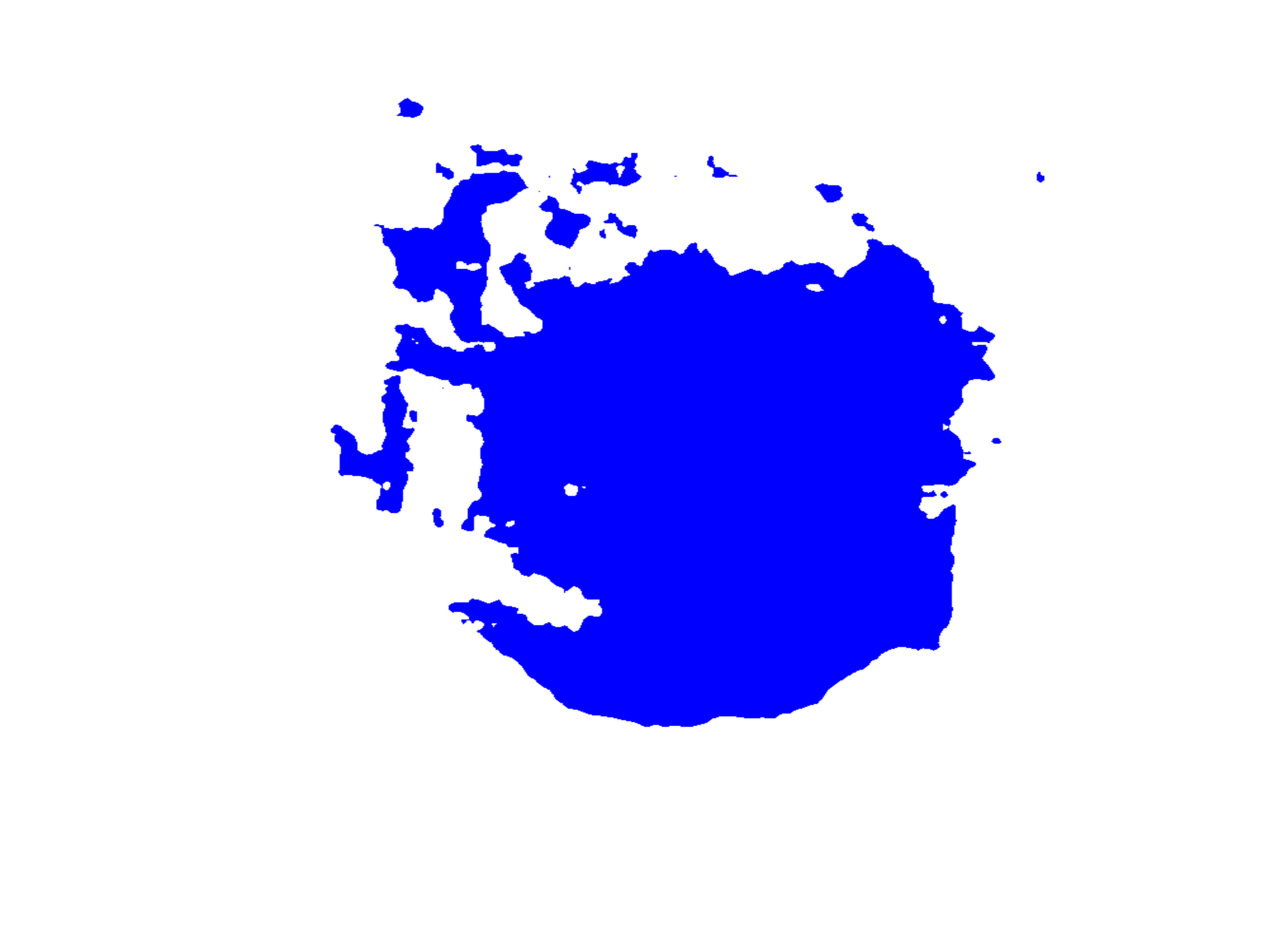}  & \includegraphics[width=0.15\textwidth]{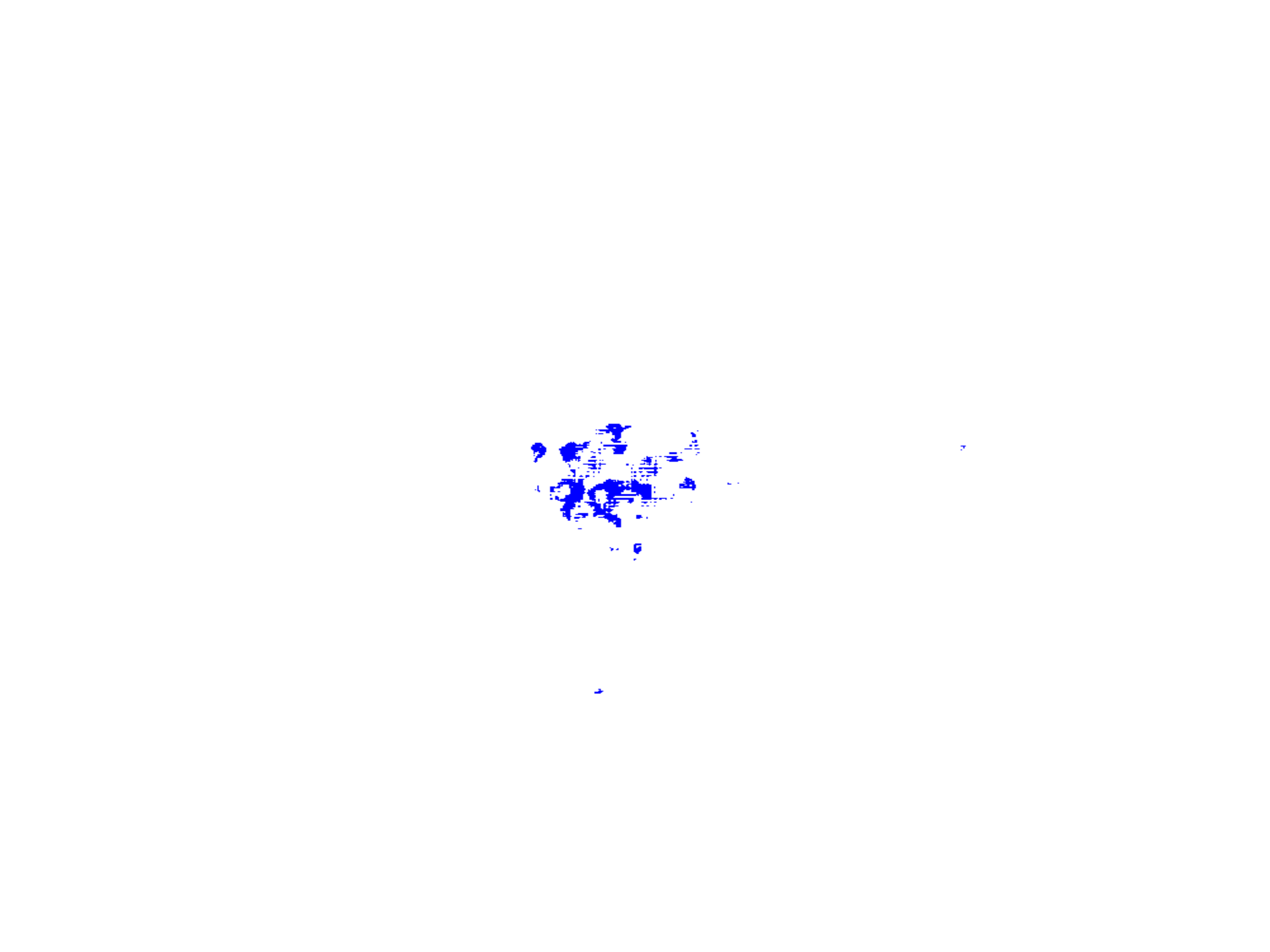}  & \includegraphics[width=0.15\textwidth]{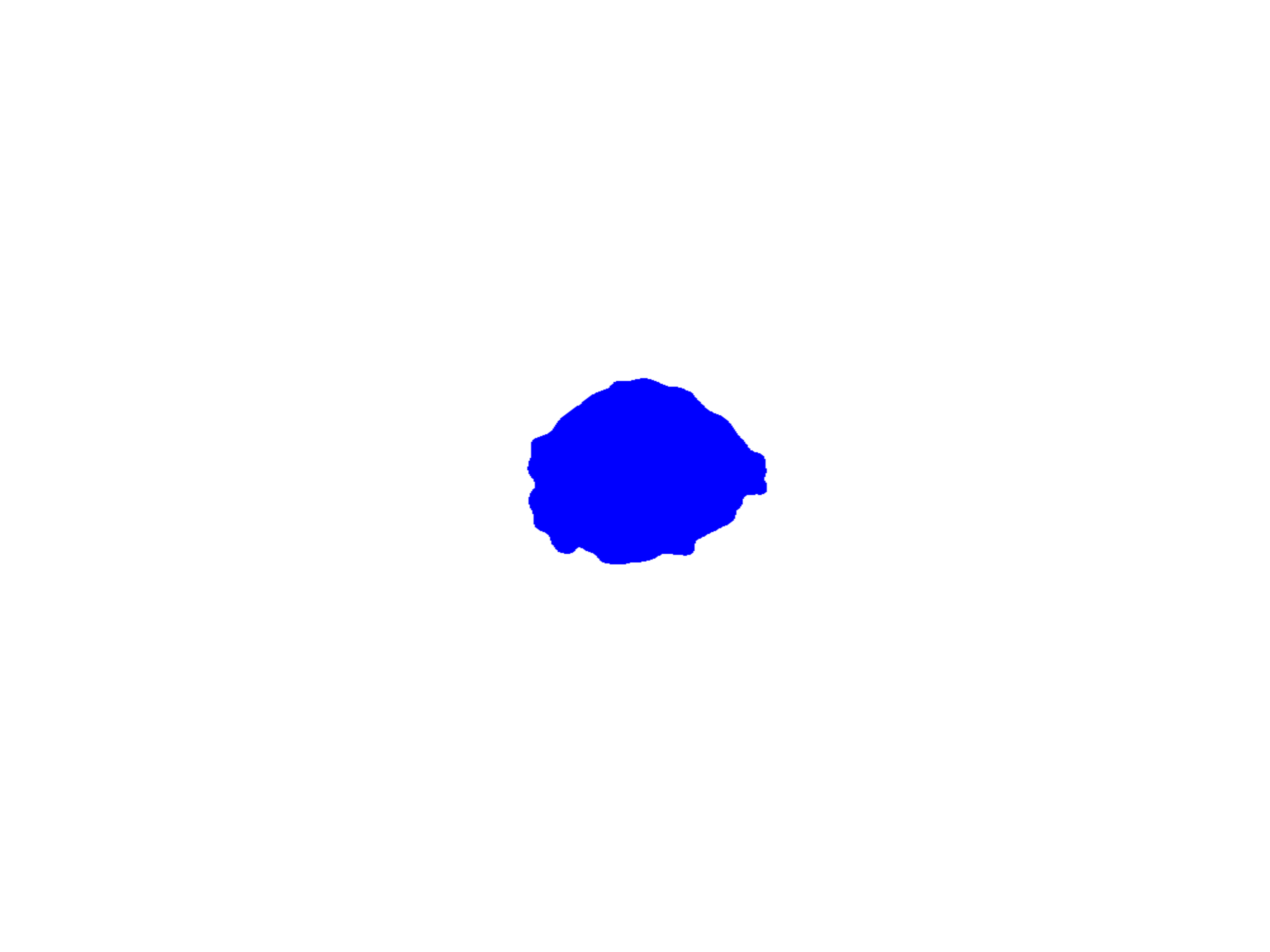}  \\ 
\centering ForkGAN            & \includegraphics[width=0.15\textwidth]{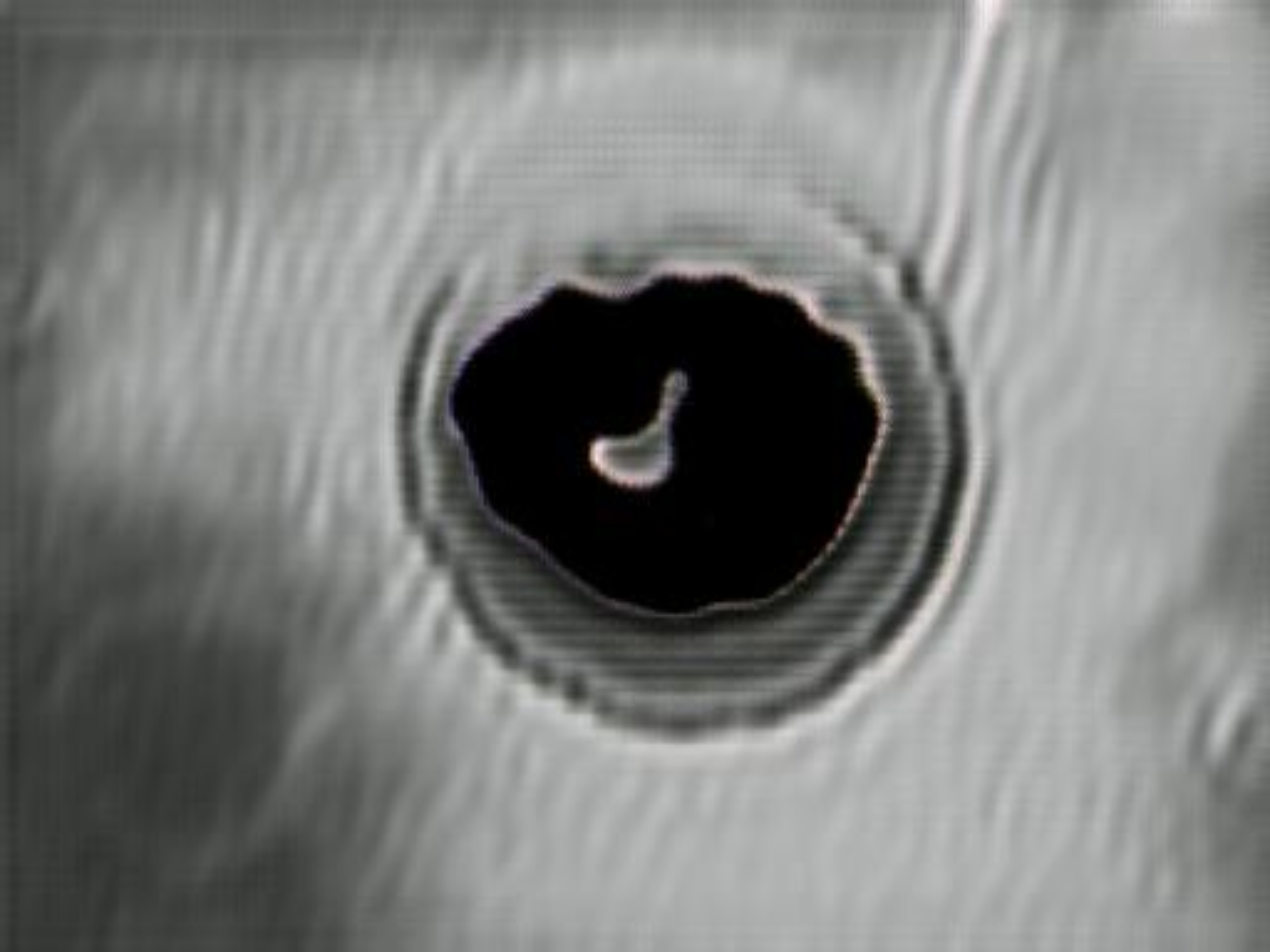}  &
\includegraphics[width=0.15\textwidth]{results/truth.jpg}&
\includegraphics[width=0.15\textwidth]{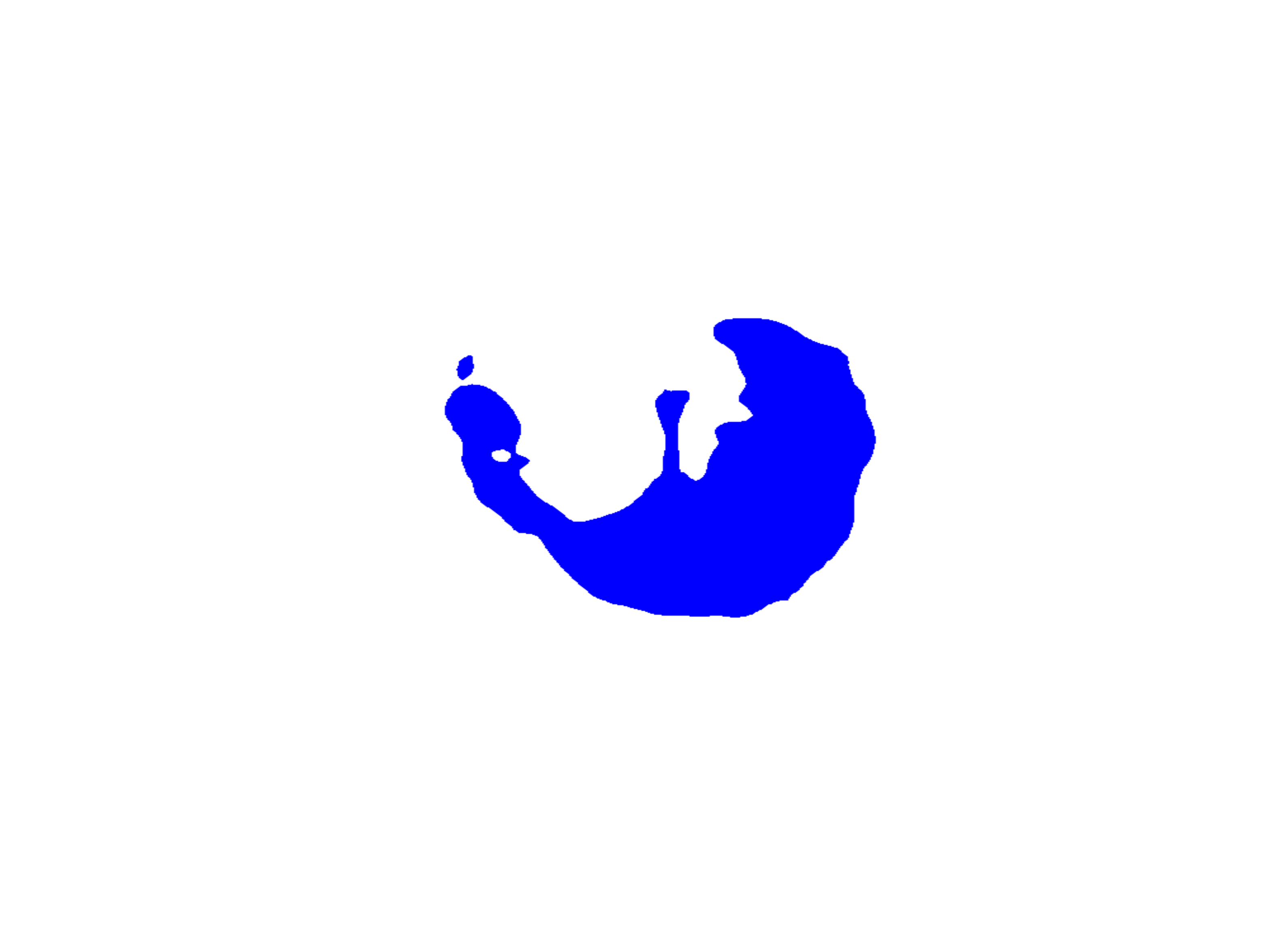}  & \includegraphics[width=0.15\textwidth]{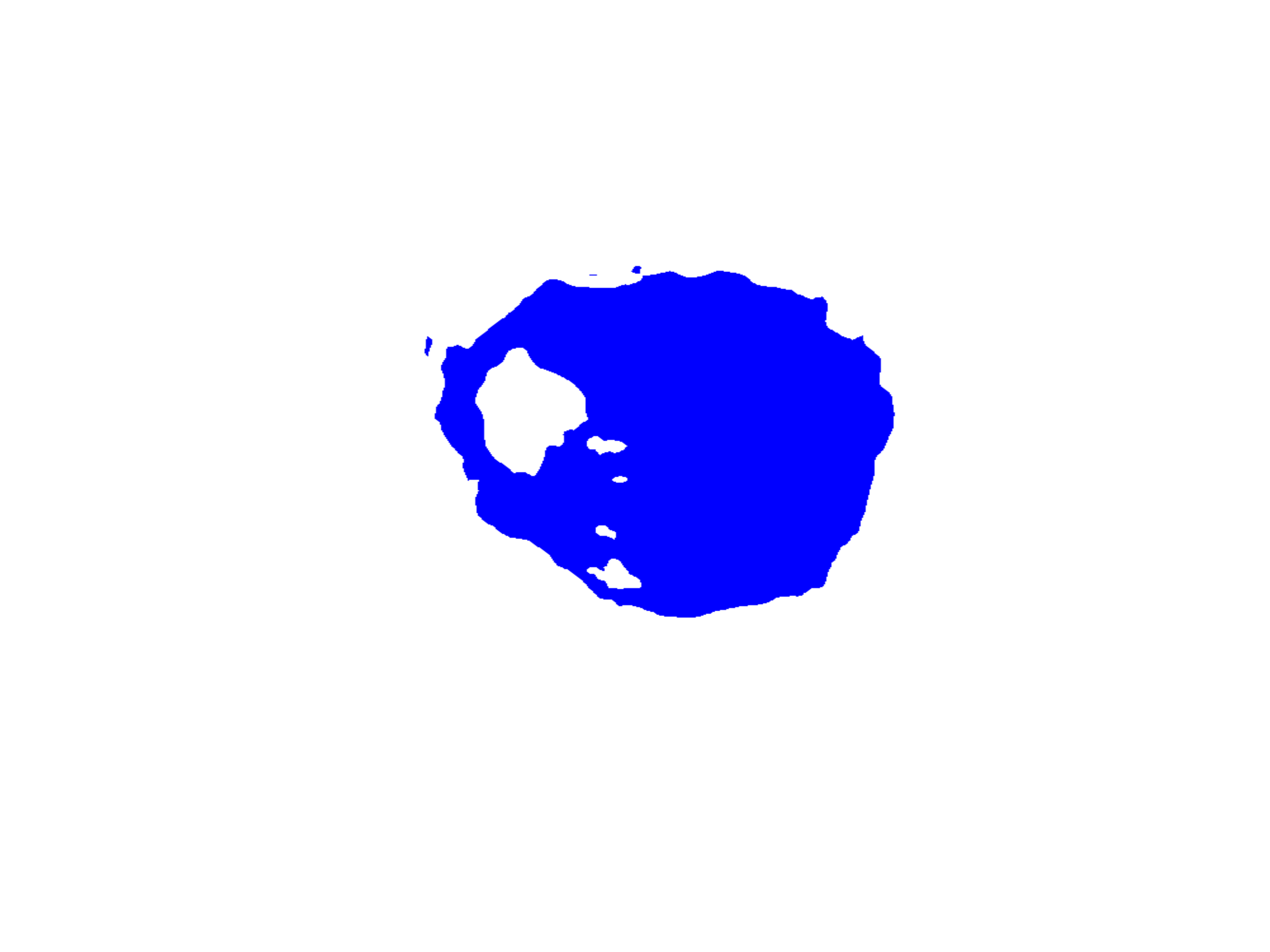}  & \includegraphics[width=0.15\textwidth]{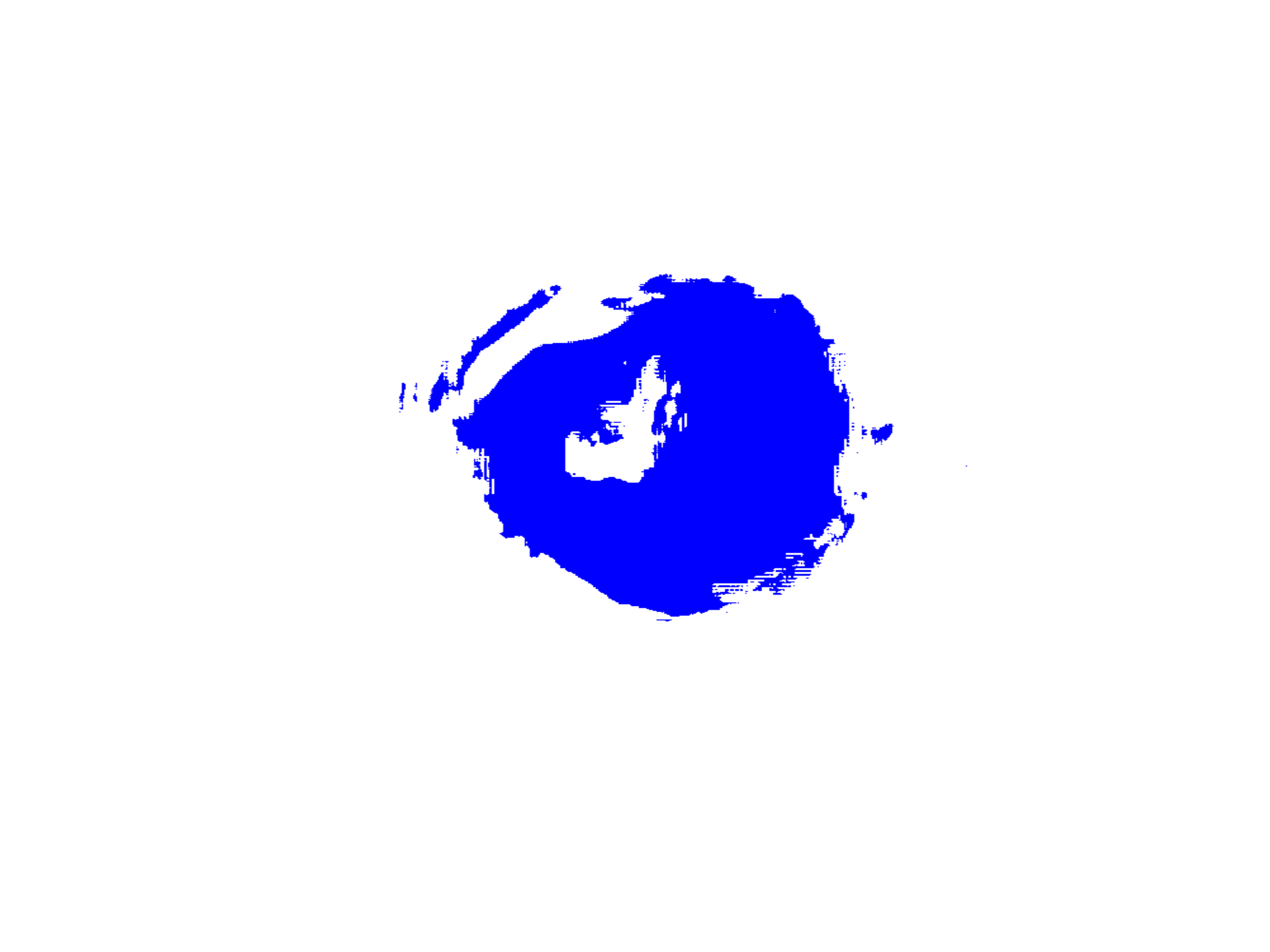}  & \includegraphics[width=0.15\textwidth]{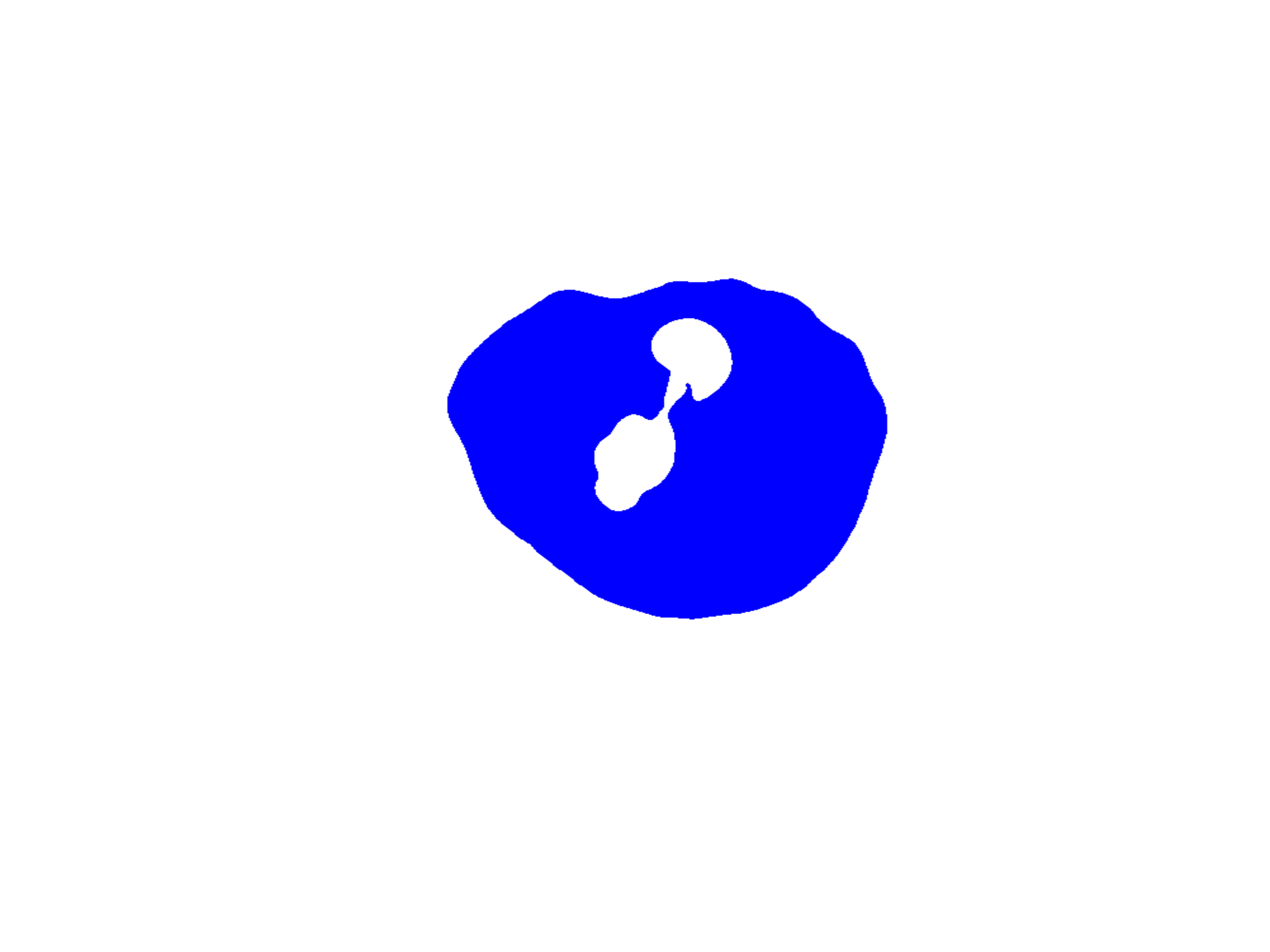}  \\
\centering GANILLA            & \includegraphics[width=0.15\textwidth]{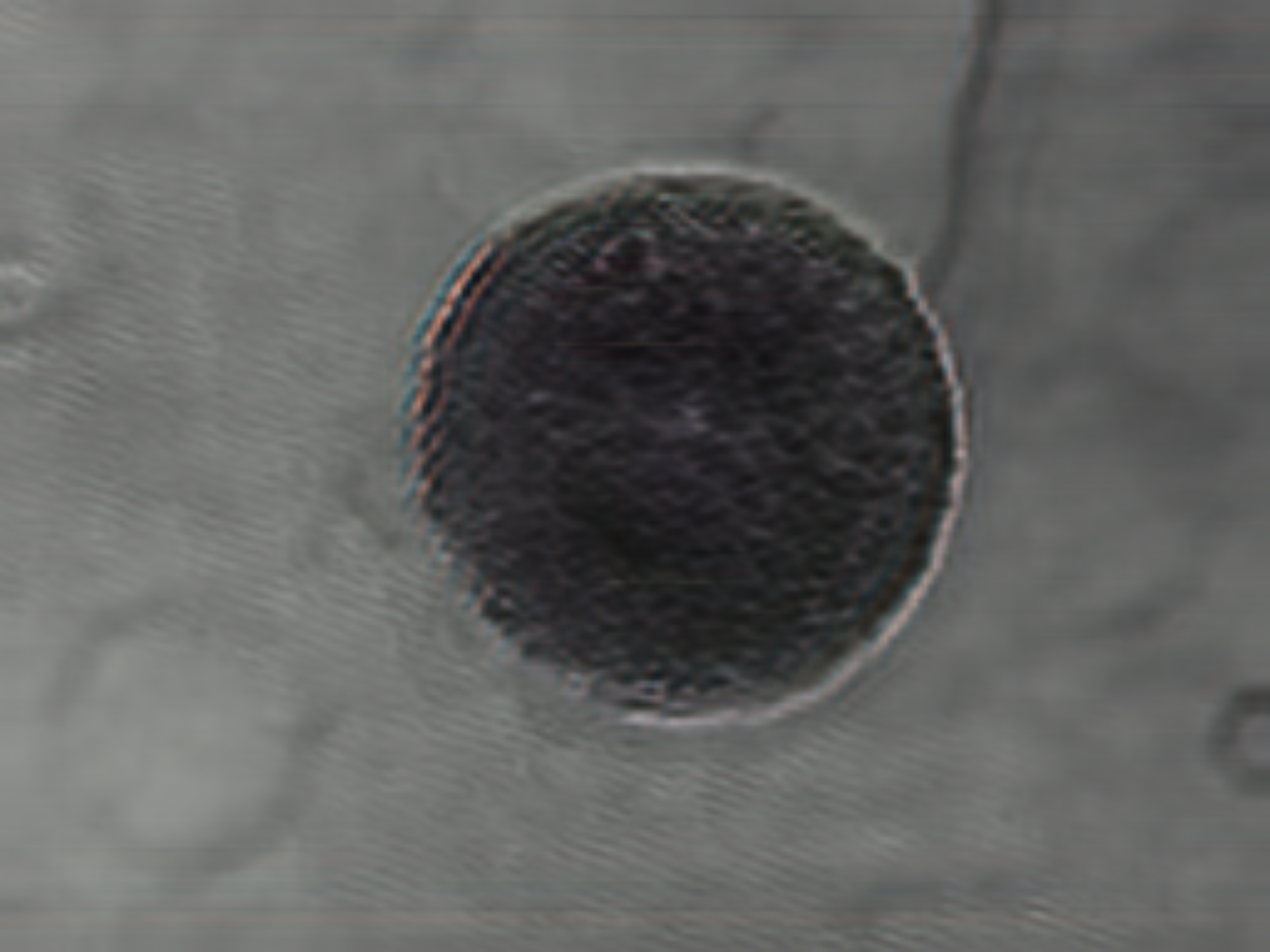}  &
\includegraphics[width=0.15\textwidth]{results/truth.jpg}&
\includegraphics[width=0.15\textwidth]{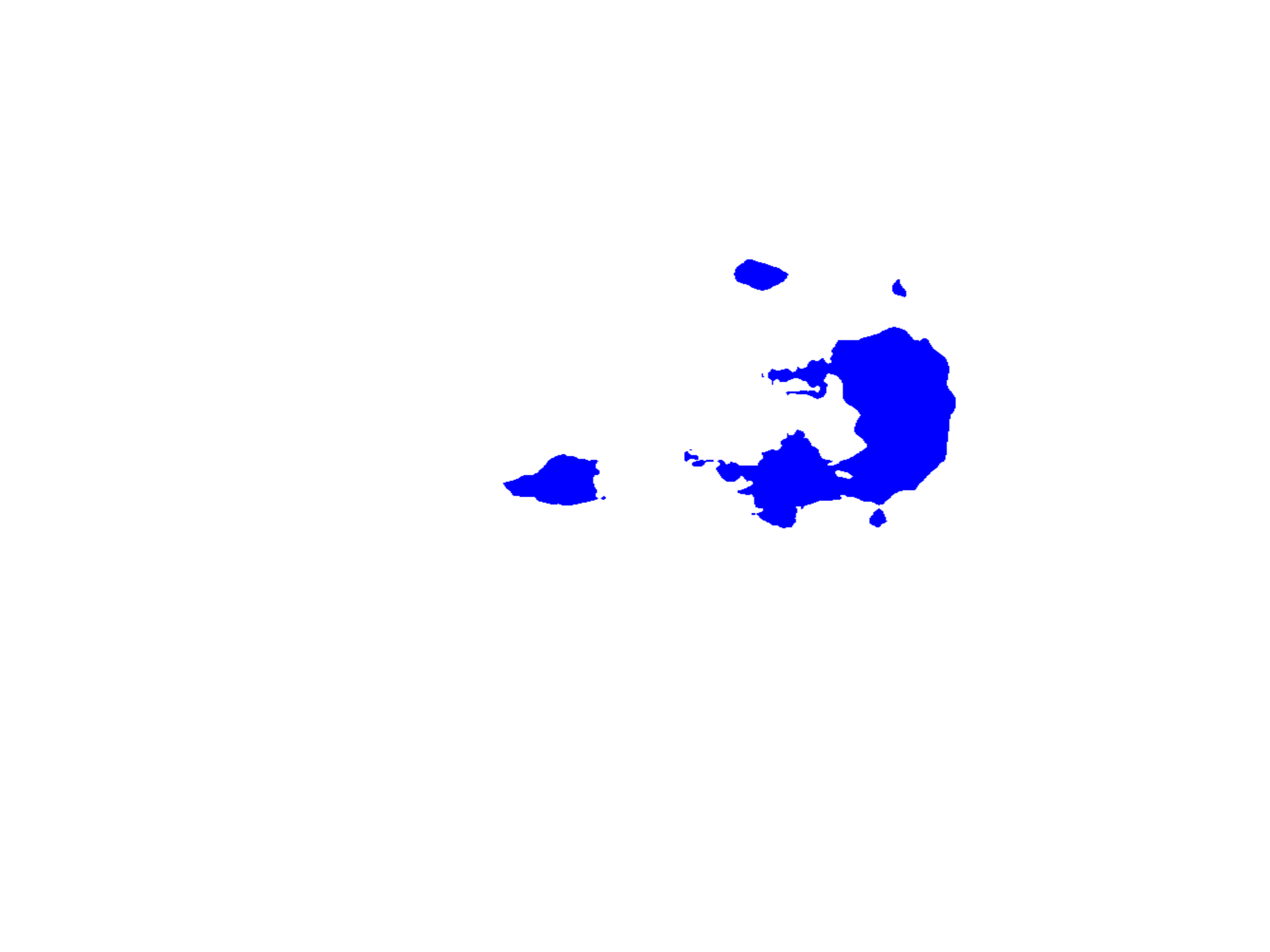}  & \includegraphics[width=0.15\textwidth]{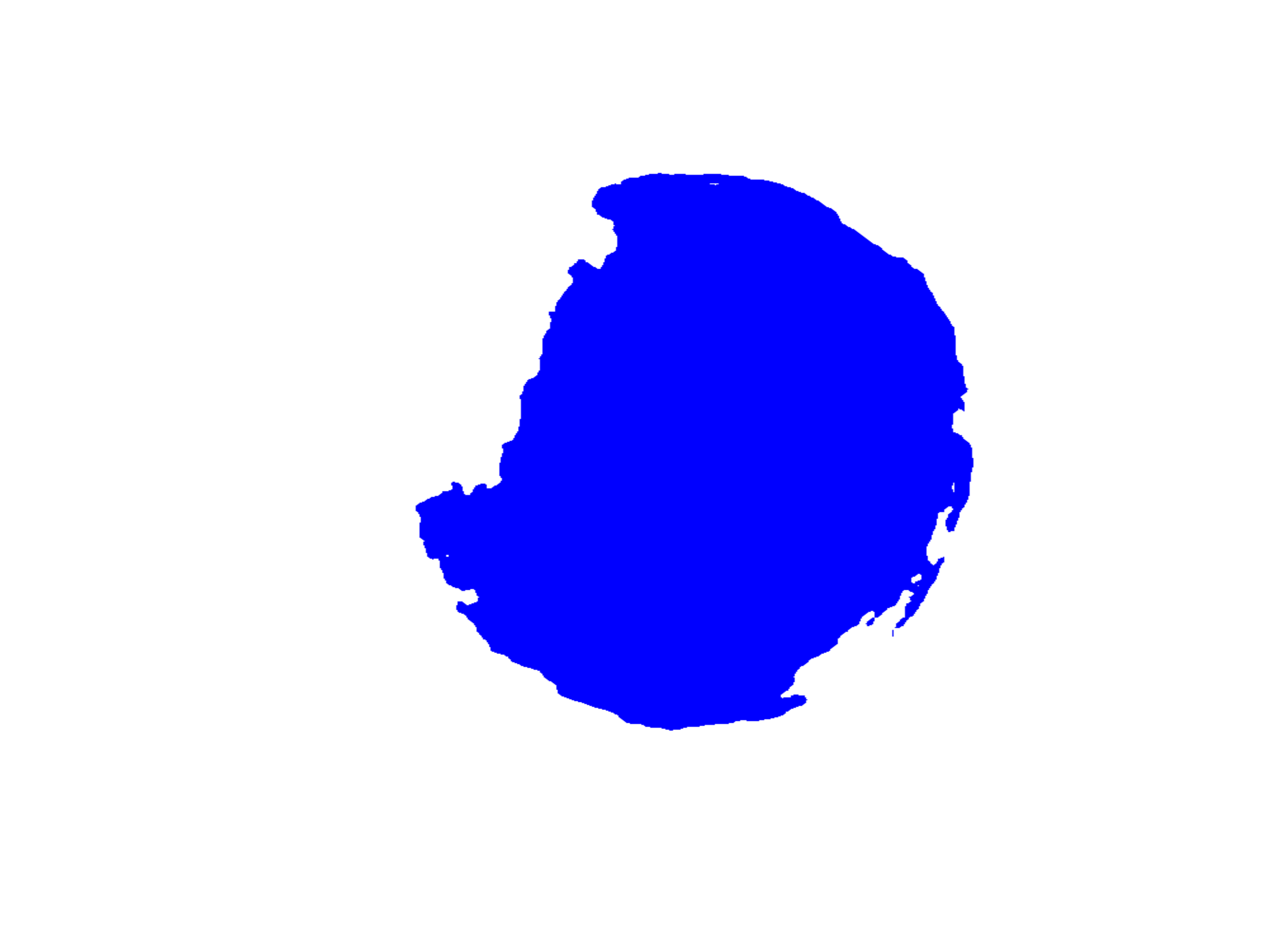}  & \includegraphics[width=0.15\textwidth]{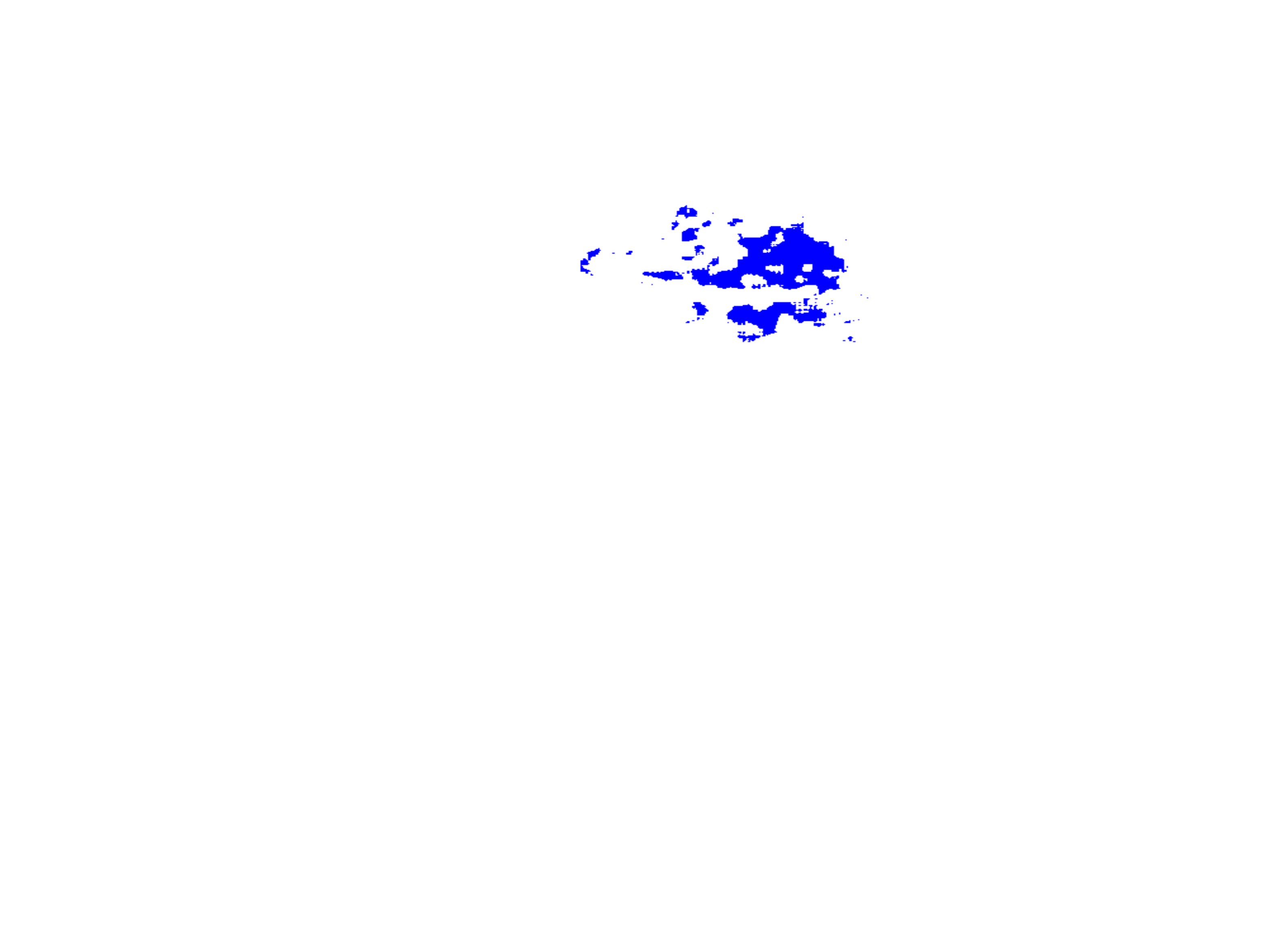}  & \includegraphics[width=0.15\textwidth]{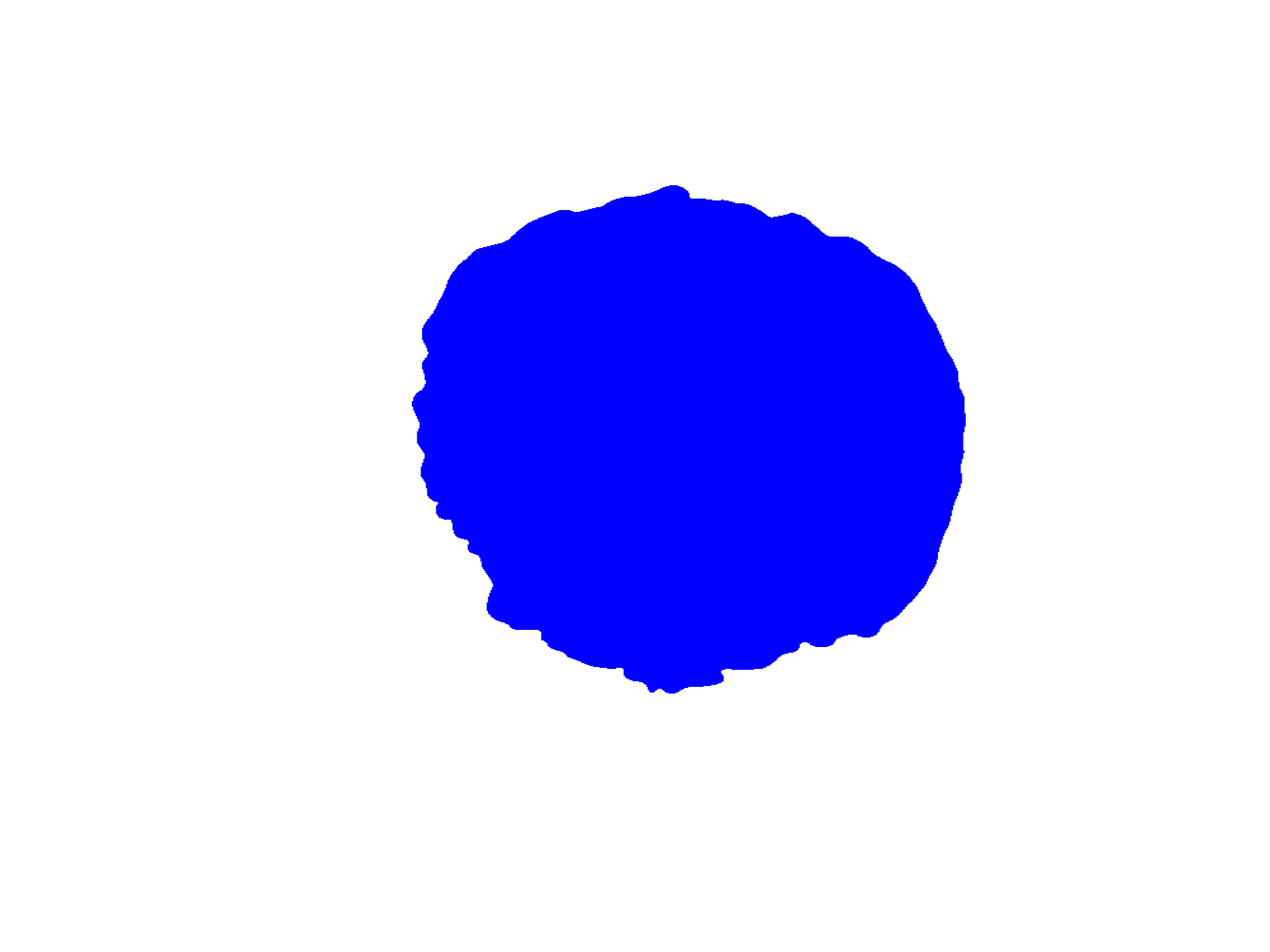}  \\ 
\centering CUT                & \includegraphics[width=0.15\textwidth]{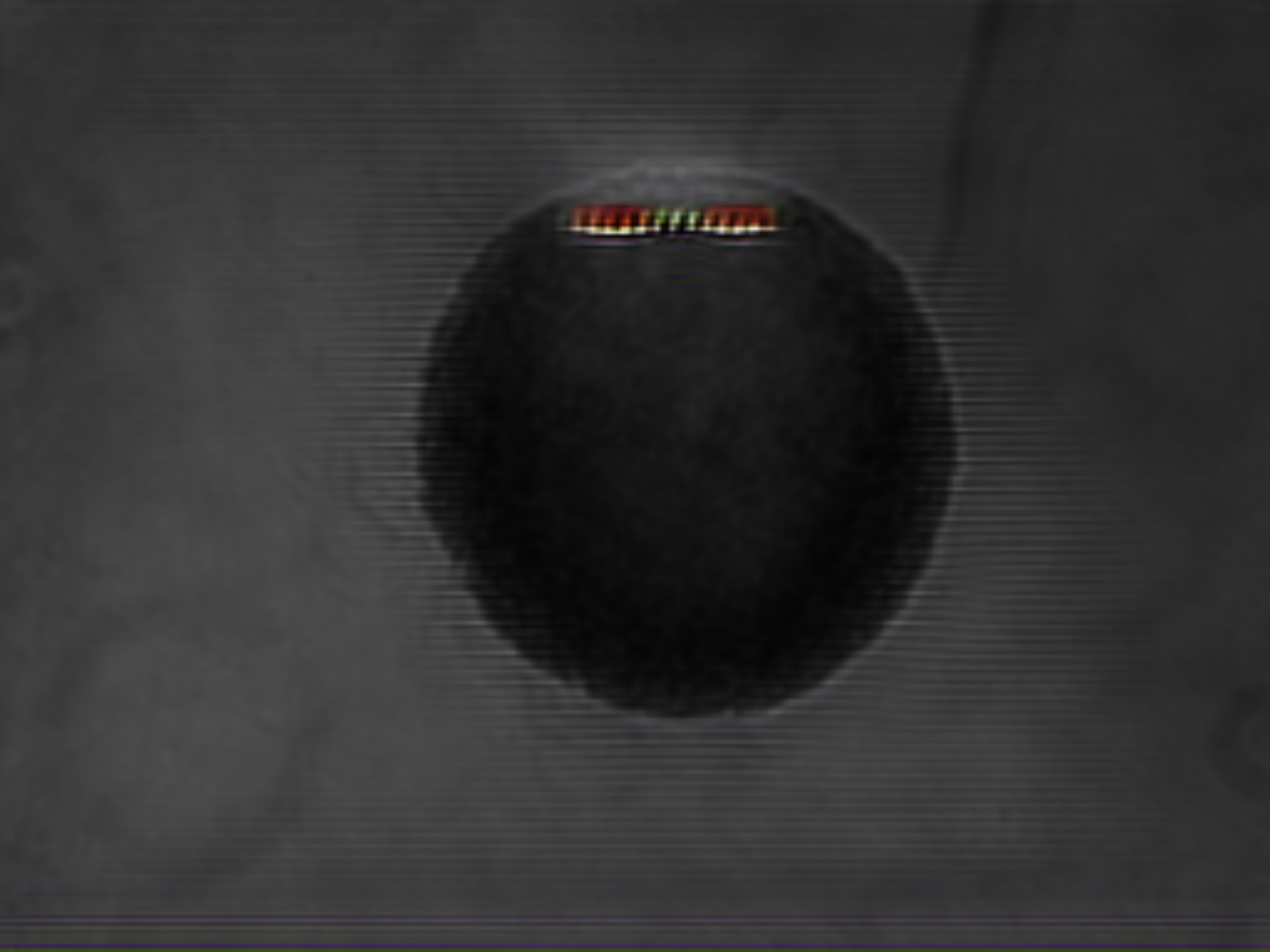}      &
\includegraphics[width=0.15\textwidth]{results/truth.jpg}&
\includegraphics[width=0.15\textwidth]{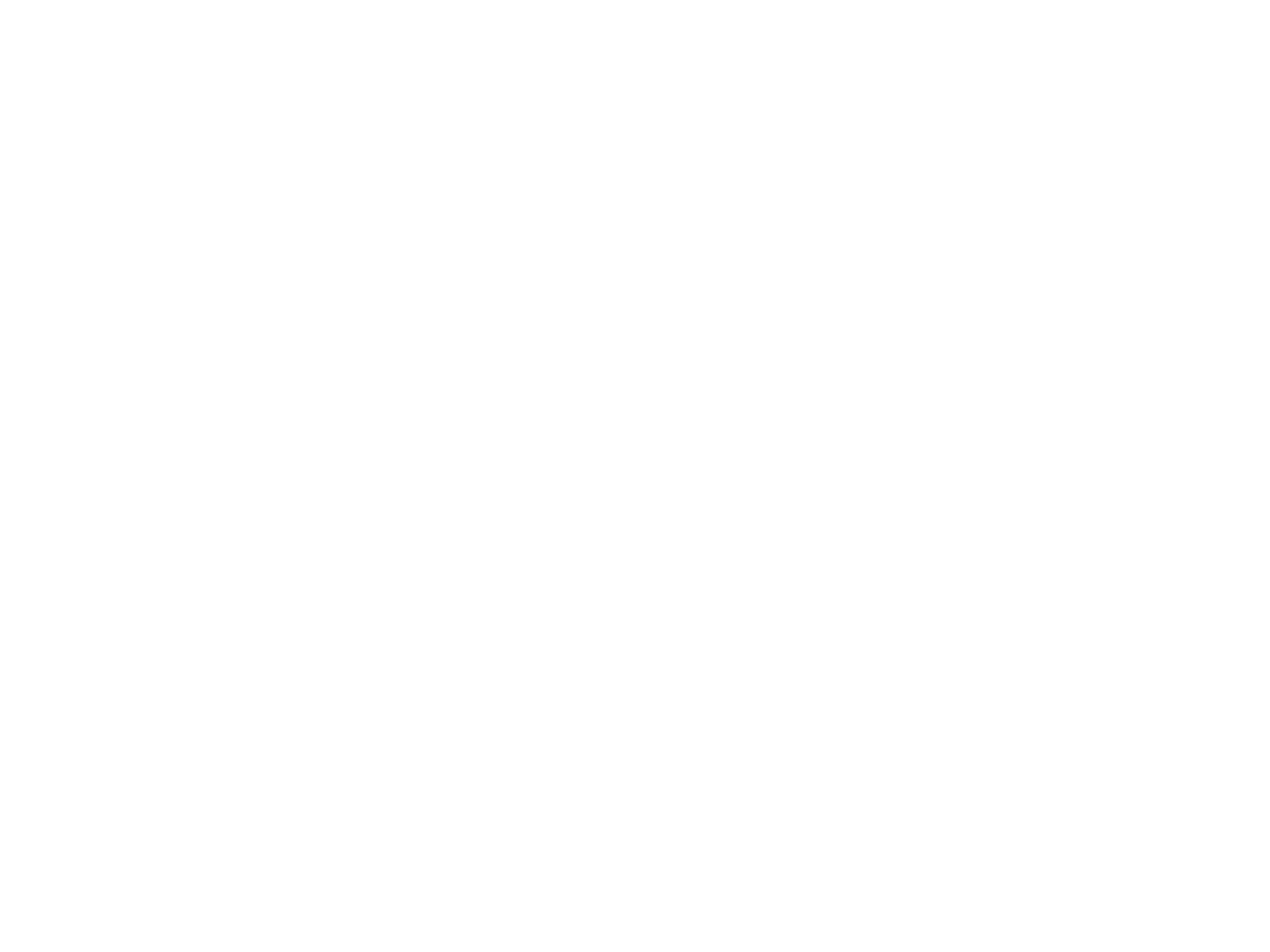}      & \includegraphics[width=0.15\textwidth]{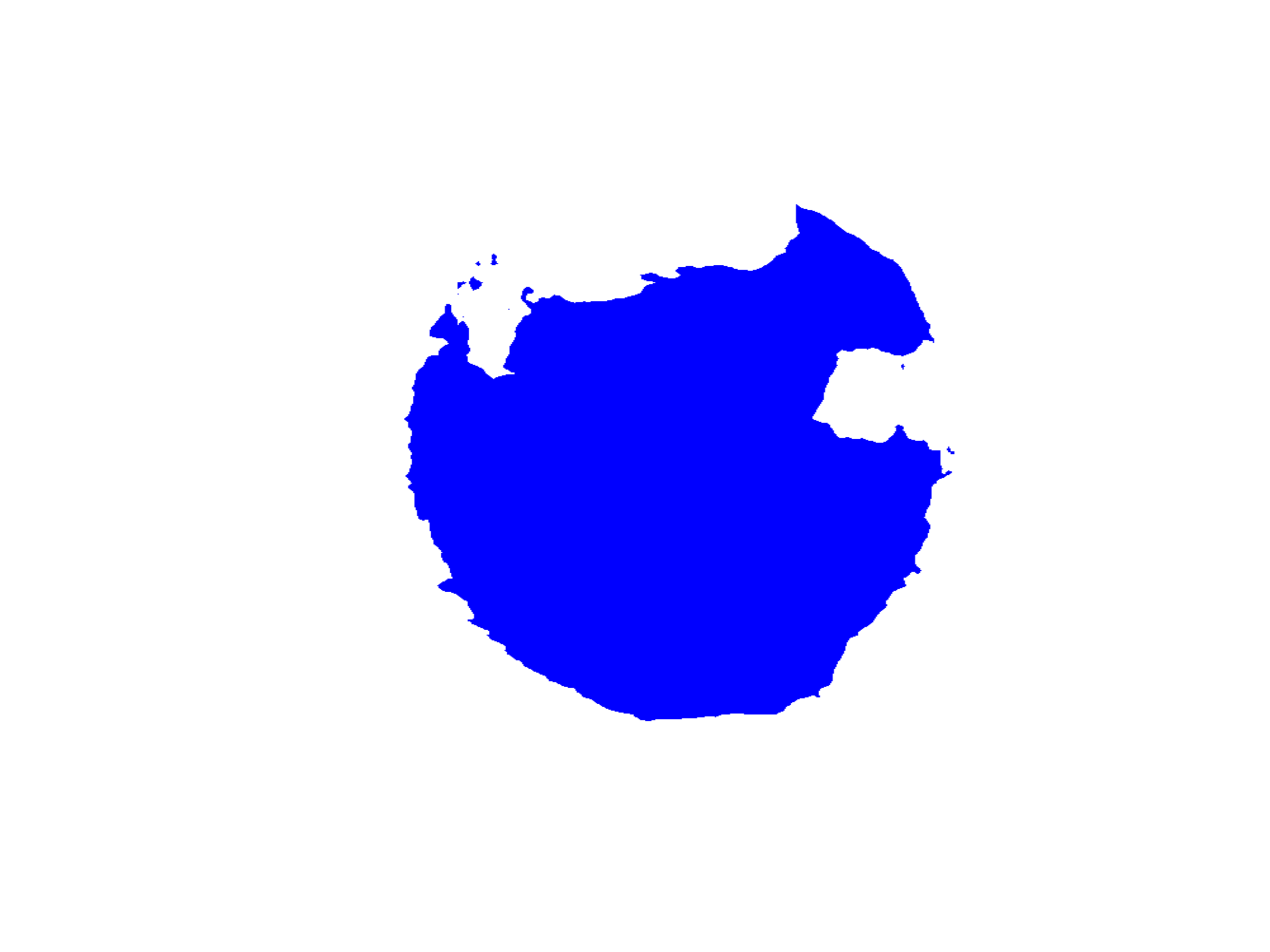}      & \includegraphics[width=0.15\textwidth]{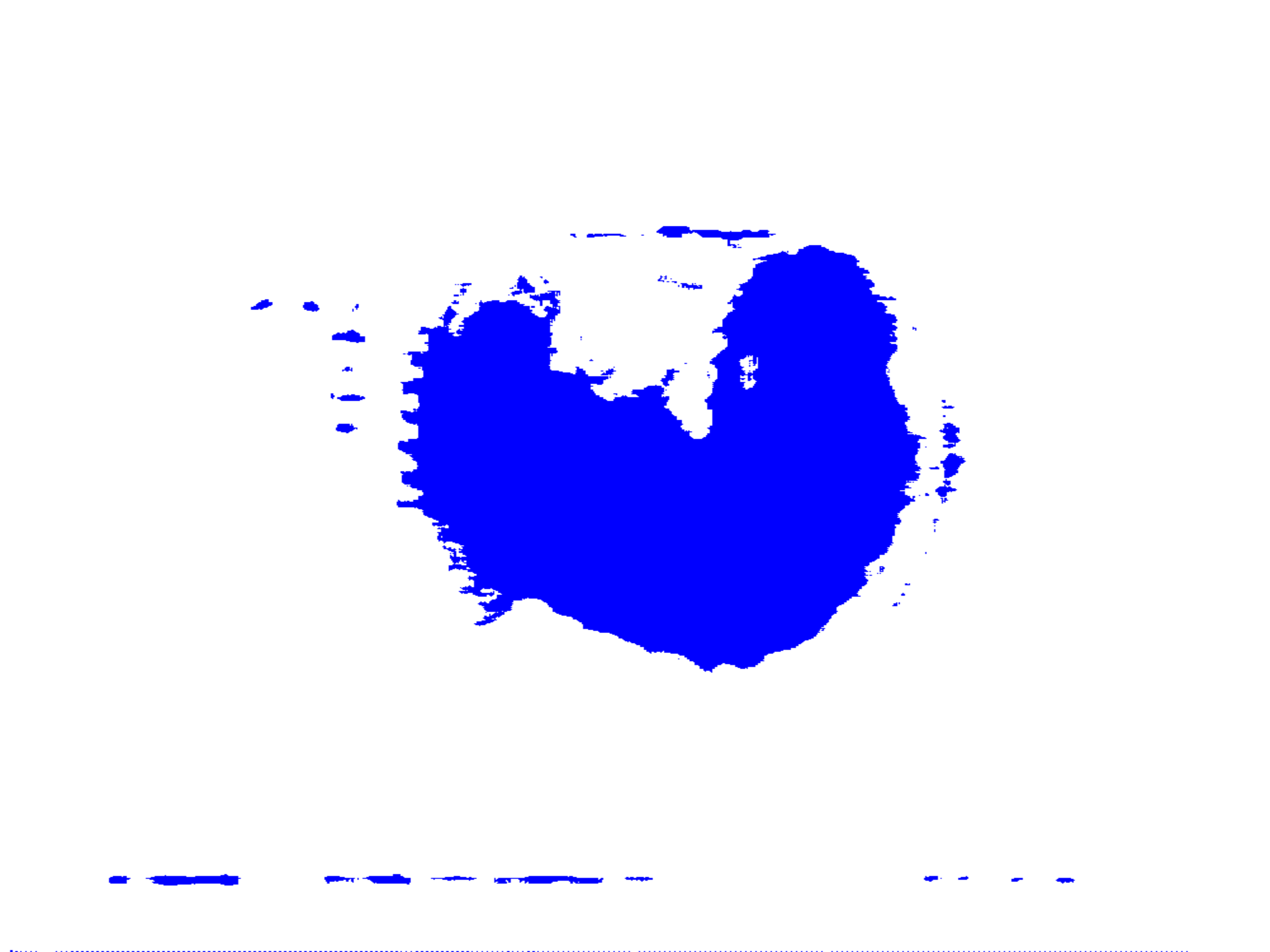}      & \includegraphics[width=0.15\textwidth]{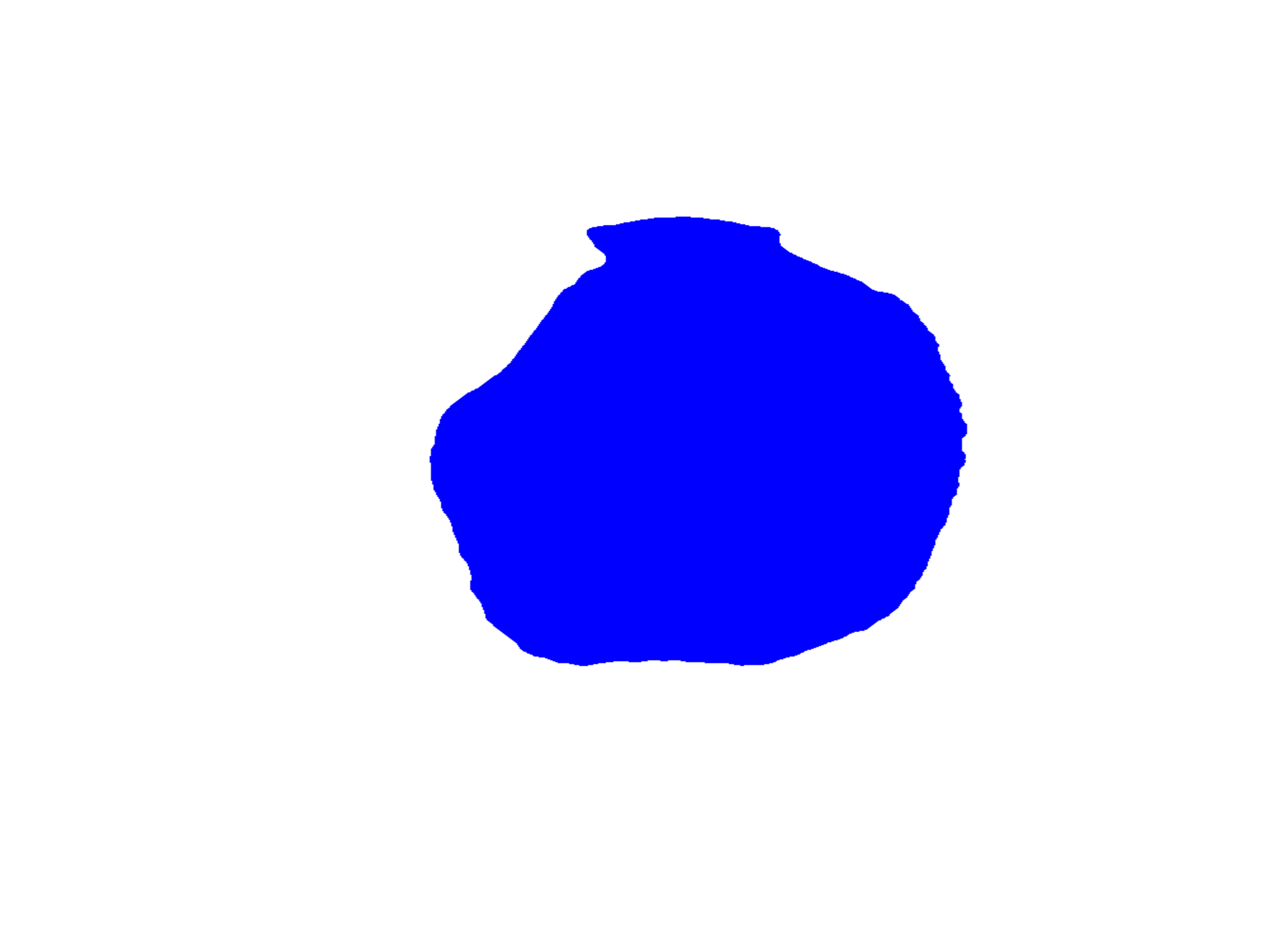}      \\ 
\centering FastCUT            & \includegraphics[width=0.15\textwidth]{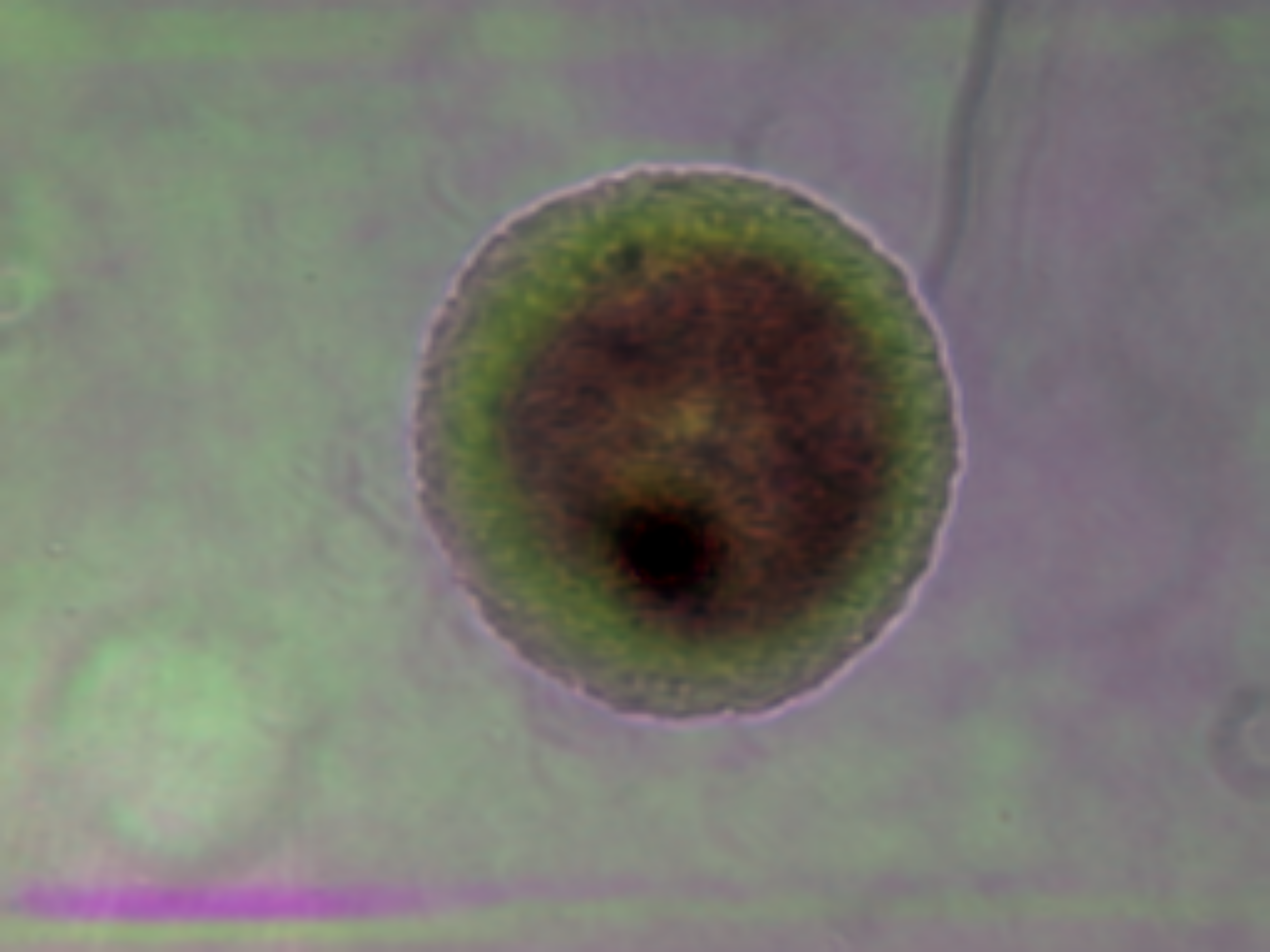}&
\includegraphics[width=0.15\textwidth]{results/truth.jpg}&
\includegraphics[width=0.15\textwidth]{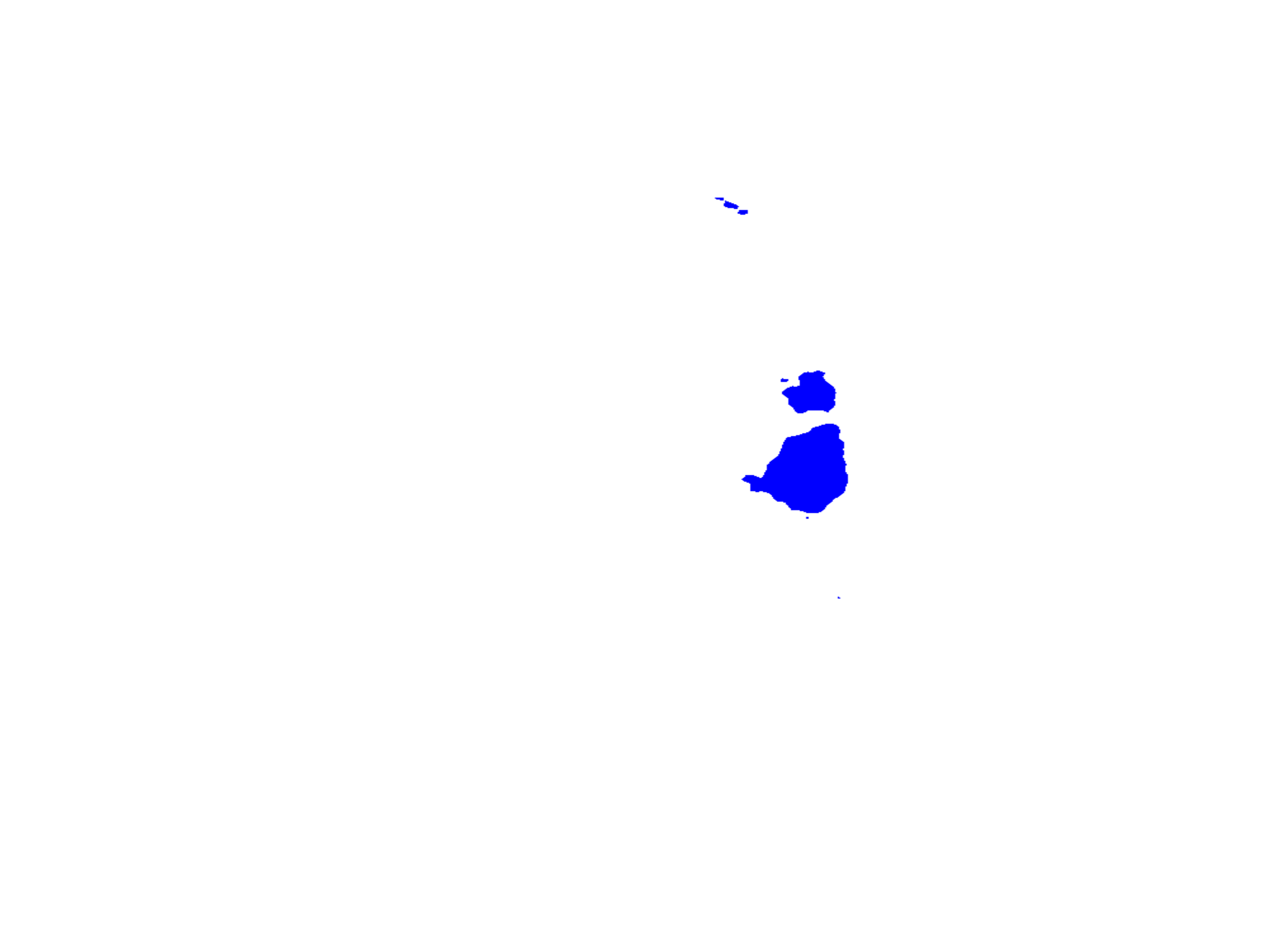} & \includegraphics[width=0.15\textwidth]{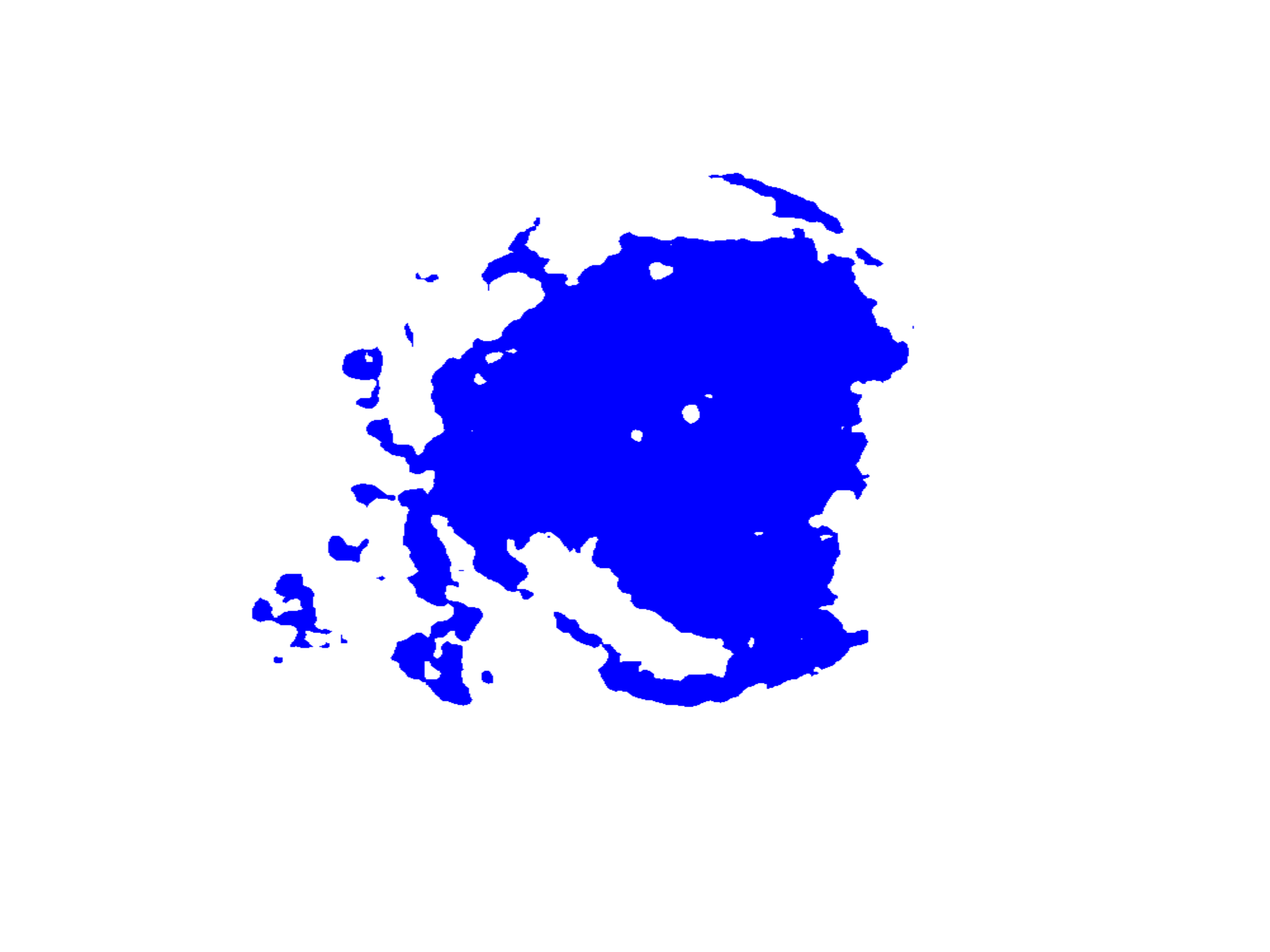}& \includegraphics[width=0.15\textwidth]{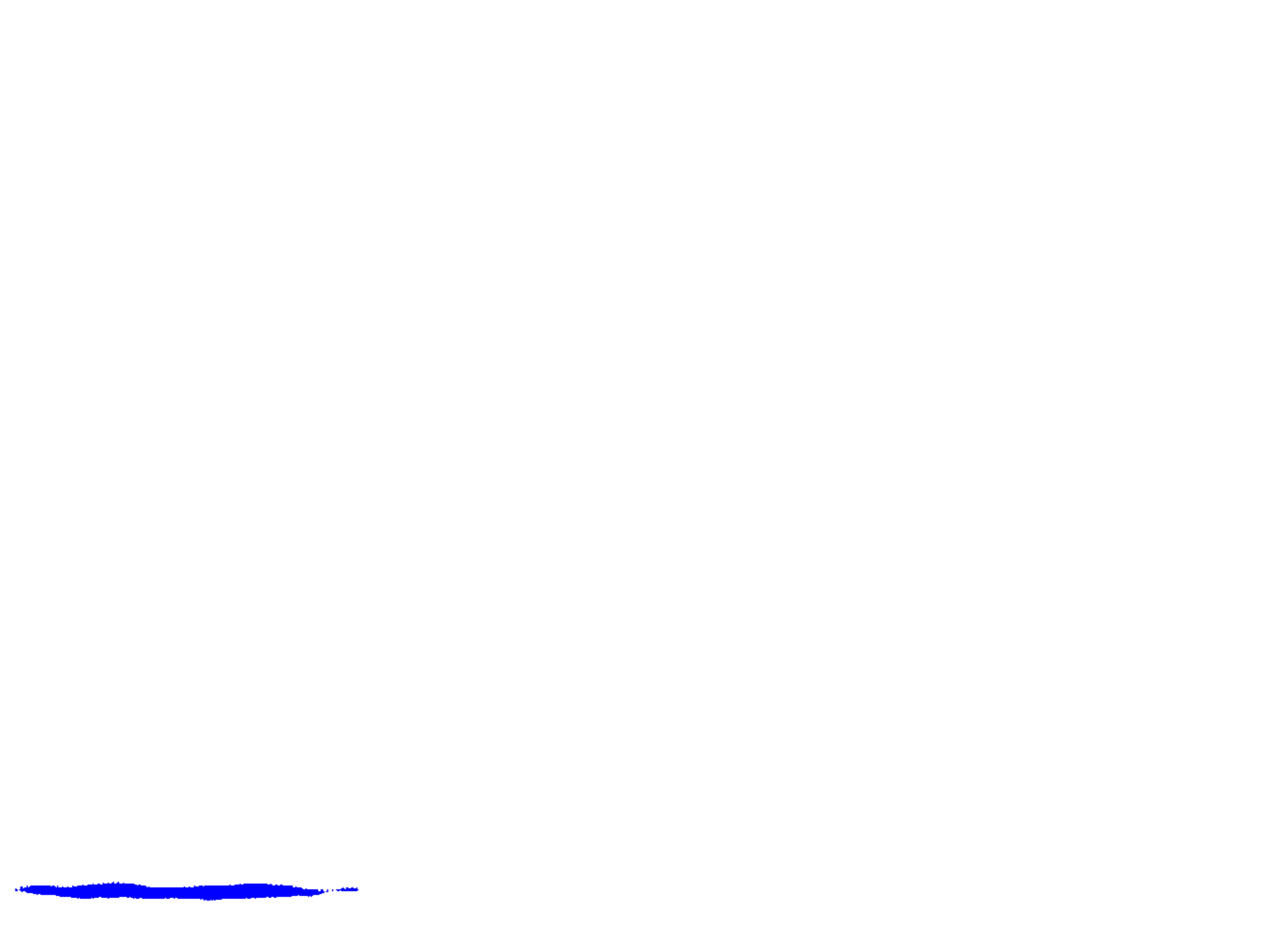} & \includegraphics[width=0.15\textwidth]{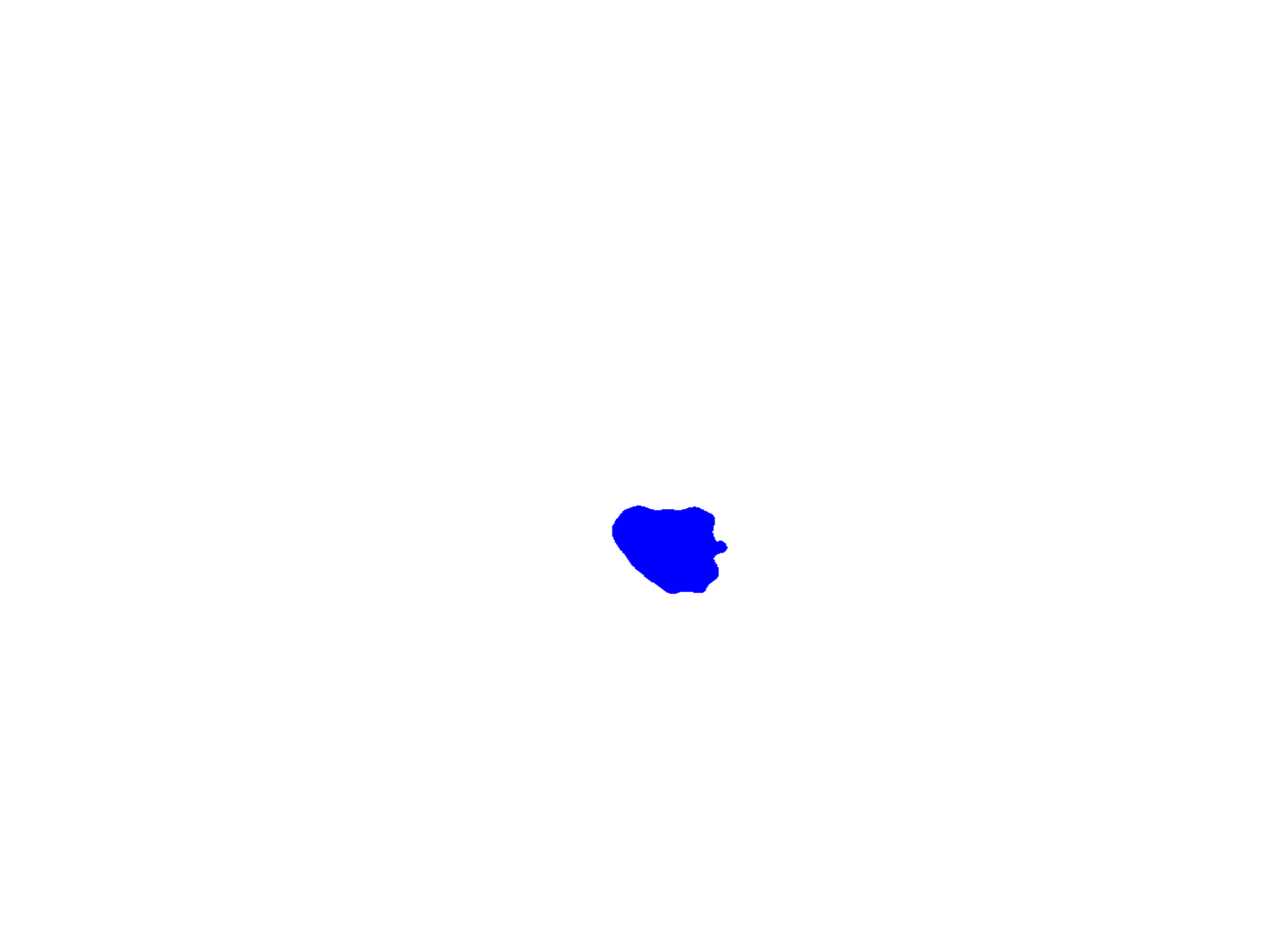}\\ 
\bottomrule
\end{tabular}}
    \caption{An example showing the segmentation produced by the DeepLab, HRNet, U-Net and U$^2$-Net models after applying a style transfer algorithm or an image-to-image translation model to a given image}
    \label{fig:my_label}
    \end{center}
\end{figure*}

We can visually inspect the images produced by the different transformation algorithms to discover the difficulties faced by the segmentation models, see Figure~\ref{fig:my_label}. We can notice that the three successful models (NST, STROTSS, and CycleGAN) produce images that preserve the content of the image but with a style that is similar to the style of those used for training the segmentation models. On the contrary, the deep image analogy method, and the DualGAN and ForkGAN models do not keep the content of the image; and, thus the segmentation models are not able to properly segment the images. For the rest of the image-to-image  transformation models (GANILLA, CUT and FastCUT), the content of the image is kept, but the style is not properly transferred (colour artefacts are added to the transformed image); hence, the images are not segmented properly by the models. 

From Figure~\ref{fig:my_label}, we can also appreciate the sensibility of the segmentation models to variations in the input image. The HRNet Seg and U$^2$-net models are more robust than the DeepLab and U-net models --- recall that all the models achieved an IoU over 97\% when evaluating in data from the distribution of the training set. Hence, the style transfer and image-to-image translation methods can be employed not only to deal with the domain shift problem of computer vision models, but also to evaluate the robustness of such models.

\section{Conclusions}

In this paper, we have studied the benefits of applying unpaired style transfer techniques and image-to-image translation methods to deal with the domain shift problem in the context of tumour segmentation. The results show us that, using those translation methods, it is possible to recover the performance of a model that suffers from the domain shift problem. In addition, we have shown that style transfer methods achieve similar results to those obtained by image-to-image translation methods with the advantage of not requiring a training step, and can be deployed by providing a single image from the source dataset. We have also noticed that style transfer techniques and image-to-image translation methods have a different impact on the performance of the segmentation models; hence, it is important to have a simple approach to test different algorithms. This has been solved in this work with the development of a high-level API that facilitates the process of testing different alternatives for style transfer and unpaired image-to-image translation.

\section*{Competing interests}
  The authors declare that they have no competing interests.
 
\section*{Funding}
This work was partially supported by Ministerio de Ciencia e Innovación [PID2020-115225RB-I00 / AEI / 10.13039/501100011033].  Manuel García-Domínguez has a FPI grant from Community of La Rioja 2018.

\bibliographystyle{plain}
\bibliography{biblio}


\begin{figure*}
    \begin{tabular}[t]{m{2cm}m{8cm}}
\includegraphics[width=1in,height=1.25in,clip,keepaspectratio]{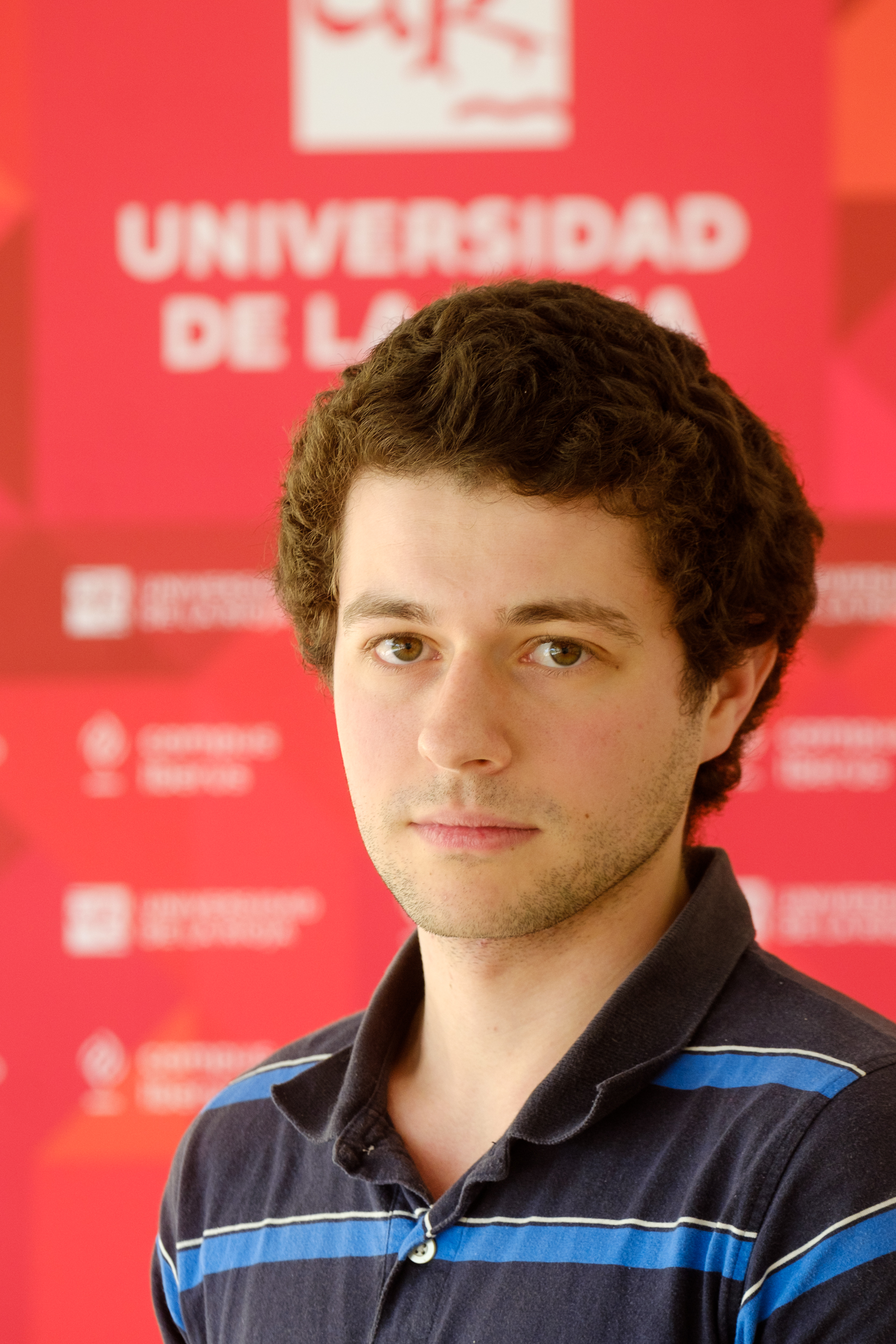}&
Manuel García-Domínguez received the B.S. degree in computer science in 2016, and the M.S. degree in computer technologies, in 2017 from University of La Rioja, Logroño, Spain. He is currently pursuing the Ph.D. degree in computer science from  University of La Rioja, Logroño, Spain. His research interests include  machine learning and image analysis. \\ 
\includegraphics[width=1in,height=1.25in,clip,keepaspectratio]{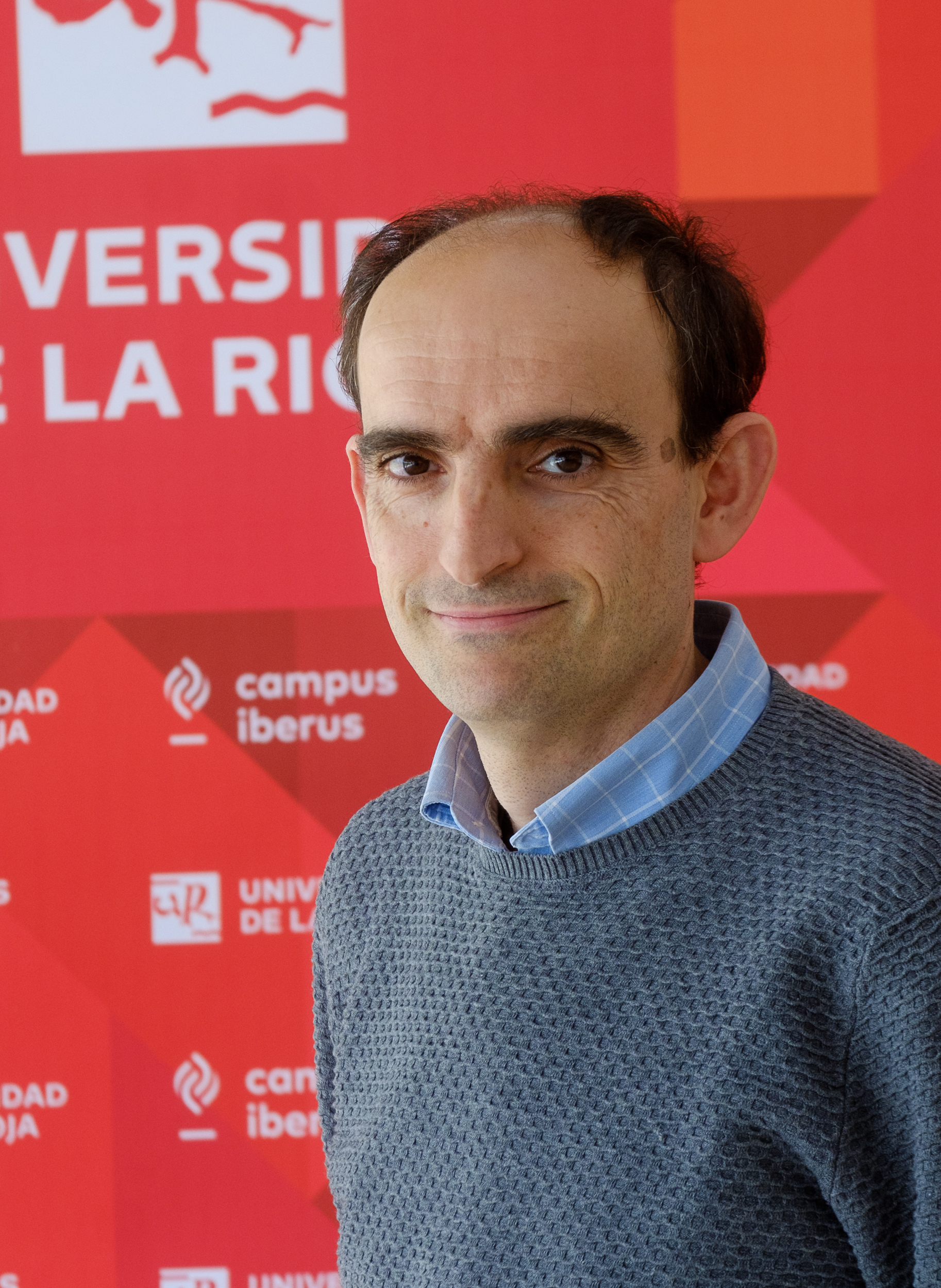}&
César Domínguez received the B.S. degree in mathematics from Zaragoza University, Zaragoza, Spain, in 1996, and the Ph.D. degree in mathematics from University of La Rioja, Logroño, Spain, in 2003. He is currently an Associate Professor at  University of La Rioja. His research interests include formal methods, e-learning, machine learning, deep learning and image analysis. \\
\includegraphics[width=1in,height=1.25in,clip,keepaspectratio]{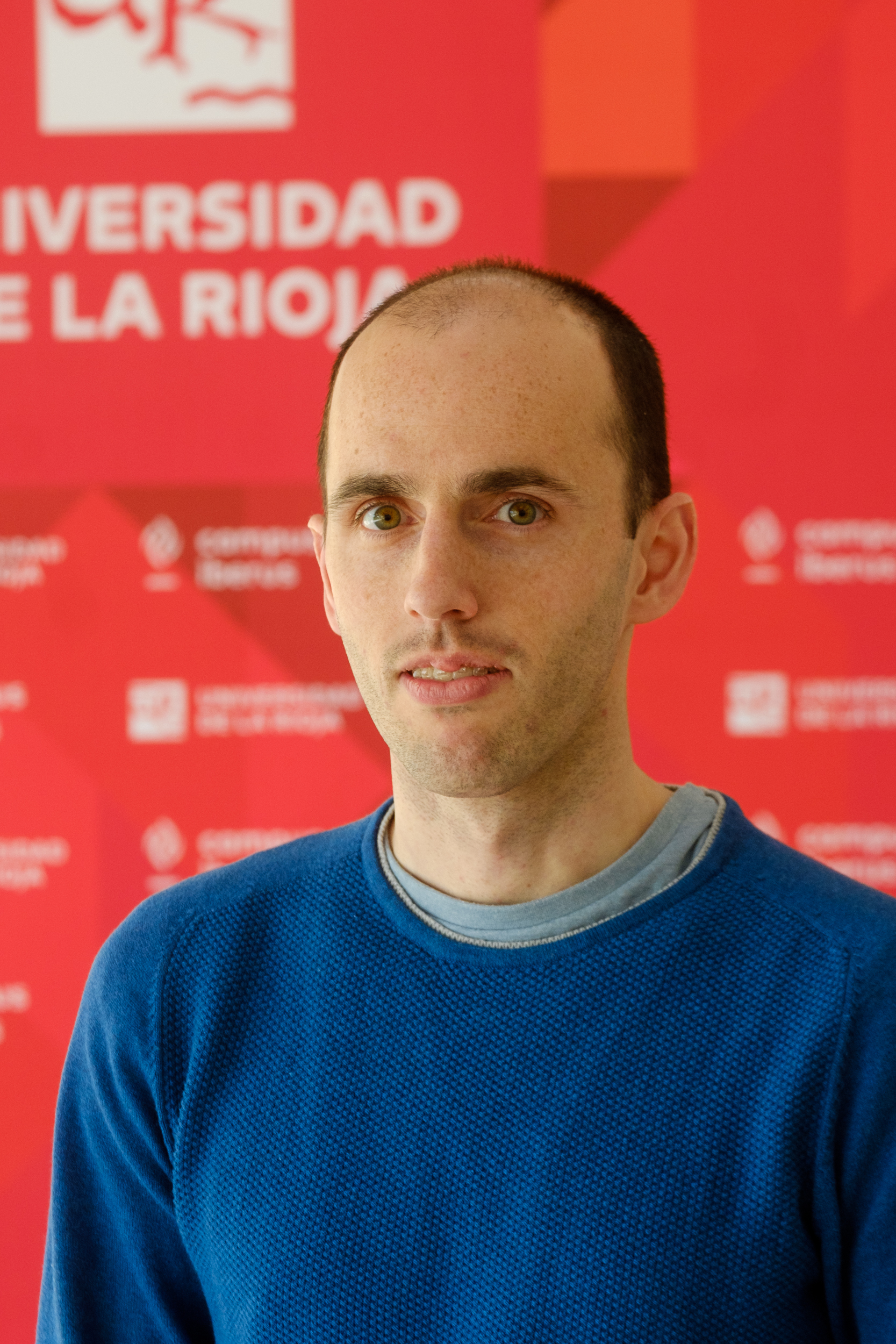} &
Jónathan Heras received the B.S. degree in mathematics from University of La Rioja, Logroño, Spain, in 2007, the B.S. degree in computer science from University of La Rioja, Logroño, Spain, in 2008 and the Ph.D. degree in computer science from the University of La Rioja, Logroño, Spain, in 2011. He is currently an Associate Professor at University of La Rioja. His research interests include machine learning, deep learning and computer vision.\\
\includegraphics[width=1in,height=1.25in,clip,keepaspectratio]{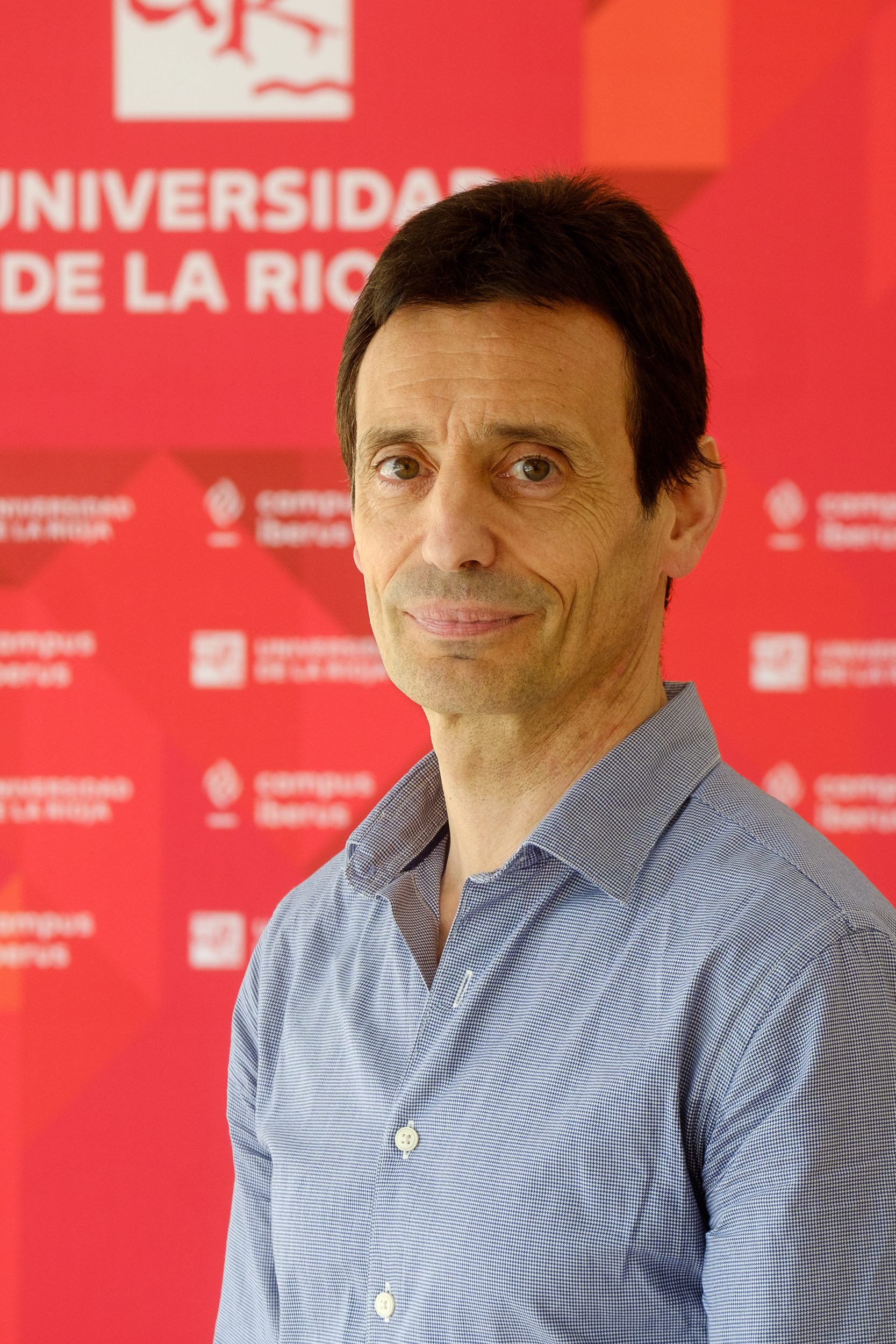} &
Eloy Mata received the B.S. degree in mathematics from Zaragoza University, Zaragoza, Spain, in 1987, and the Ph.D. degree in mathematics from University of La Rioja, Logroño, Spain, in 2009. He is currently an Associate Professor at University of La Rioja. His research interests include web services, machine learning and high performance computing.\\
\includegraphics[width=1in,height=1.25in,clip,keepaspectratio]{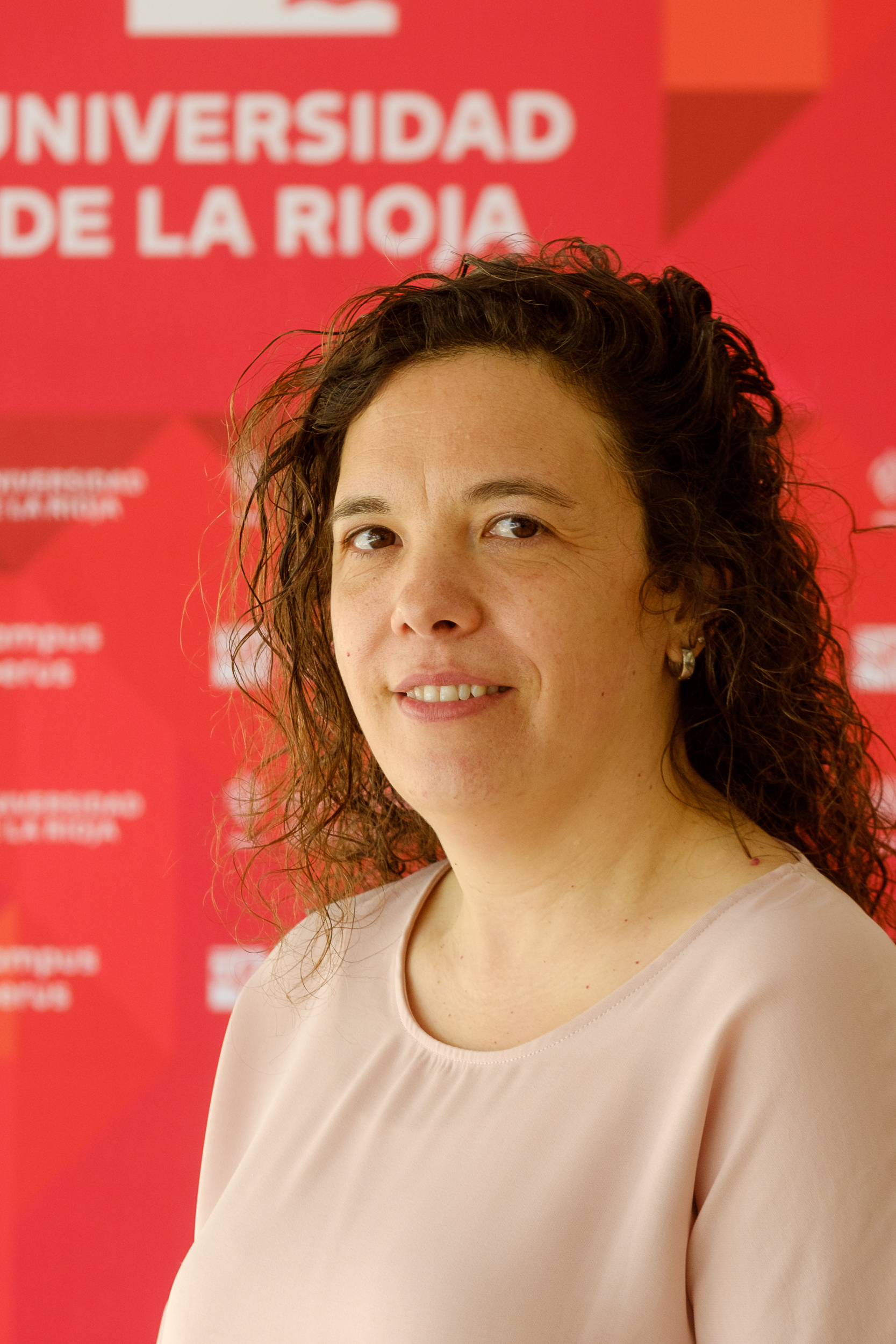}&
Vico Pascual received the B.S. degree in mathematics from Zaragoza University, Zaragoza, Spain, in 1994, and the Ph.D. degree in mathematics from University of La Rioja, Logroño, Spain, in 2002. She is currently an Associate Professor at  University of La Rioja. Her research interests include machine learning, deep learning and computer vision. \\ 

\end{tabular}
\end{figure*}

\end{document}